\documentclass{article}

\usepackage[nonatbib, final]{neurips_2020}

\usepackage[utf8]{inputenc} 
\usepackage[T1]{fontenc}    
\usepackage[hypertexnames=false]{hyperref} 
\usepackage{url}            
\usepackage{booktabs}       
\usepackage{amsfonts}       
\usepackage{nicefrac}       
\usepackage{microtype}      

\usepackage{amsmath}
\usepackage{amssymb}
\usepackage{xcolor}
\usepackage{graphicx}
\usepackage{bm}
\usepackage{tikz}

\graphicspath{{figures_arxiv_reduced/}}

\newcounter{commentH}
\newcommand{\randomcolor}{%
  \definecolor{randomcolor}{HSB}
   {
    \the\value{commentH},
    256,
    160
   }%
  \color{randomcolor}%
 }
\newcommand{\newrandomcolor}{%
  \addtocounter{commentH}{75}%
  \ifnum\value{commentH}>255%
      \addtocounter{commentH}{-256}%
  \fi%
}

\newcounter{commentcounter}
\newif\ifinlinecomments

\inlinecommentstrue

\ifinlinecomments

\else

\fi

\DeclareMathOperator*{\argmin}{\arg \min}

\newcommand{\editu}[2]{E($\bu_{#1}$, #2)}
\newcommand{\editv}[2]{E($\bv_{#1}$, #2)}
\newcommand{\bx}{\mathbf{x}}
\newcommand{\by}{\mathbf{y}}
\newcommand{\bz}{\mathbf{z}}

\newcommand{\bw}{\mathbf{w}}
\newcommand{\bv}{\mathbf{v}}

\newcommand{\bu}{\mathbf{u}}
\newcommand{\bU}{\mathbf{U}}
\newcommand{\bV}{\mathbf{V}}

\newcommand{\SMref}[1]{SM~\S\ref{sm:sec#1}} 

\newcommand{\h}{0pt}
\newcommand{\hh}{0pt}
\newcommand{\hhh}{0pt}
\newcommand{\sidelabel}[1]{this will be redefined}

\newcommand{\imagewitheditframed}[4]{
\begin{tikzpicture}[inner sep=0,outer sep=0]\node (img){\includegraphics[width=#1]{#2}};\draw[#4, line width=1pt, overlay] (current bounding box.north east) rectangle (current bounding box.south west);\node[above right,color=white,scale=0.75,overlay] at (img.south west) {#3};\end{tikzpicture}}

\newcommand{\imagewithedit}[3]{
\begin{tikzpicture}[inner sep=0,outer sep=0]\node (img){\includegraphics[width=#1]{#2}};\node[above right,color=white,scale=0.75] at (img.south west) {#3};\end{tikzpicture}}

\newcommand{\imageBlockSixA}[2]{
\includegraphics[width=#1]{#2_0}{}\hfill%
\includegraphics[width=#1]{#2_1}{}\hfill%
\includegraphics[width=#1]{#2_2}{}}

\newcommand{\imageBlockSixB}[2]{
\includegraphics[width=#1]{#2_3}{}\hfill%
\includegraphics[width=#1]{#2_4}{}\hfill%
\includegraphics[width=#1]{#2_5}{}}

\newcommand{\hsp}{\hfill}
\newcommand{\imageRowSevenFramed}[9]{
\imagewitheditframed{#1}{#2_0}{$-2\sigma$}{#3}\hsp%
\imagewitheditframed{#1}{#2_1}{$-1.33\sigma$}{#4}\hsp%
\imagewitheditframed{#1}{#2_2}{$-0.67\sigma$}{#5}\hsp%
\imagewitheditframed{#1}{#2_3}{$0\sigma$}{#6}\hsp%
\imagewitheditframed{#1}{#2_4}{$0.67\sigma$}{#7}\hsp%
\imagewitheditframed{#1}{#2_5}{$1.33\sigma$}{#8}\hsp%
\imagewitheditframed{#1}{#2_6}{$2\sigma$}{#9}}

\newcommand{\imageRowFive}[2]{
\includegraphics[width=#1]{#2_0}\hfill
\includegraphics[width=#1]{#2_1}\hfill
\includegraphics[width=#1]{#2_2}\hfill
\includegraphics[width=#1]{#2_3}\hfill
\includegraphics[width=#1]{#2_4}
}

\newcommand{\imageRowFour}[2]{
\includegraphics[width=#1]{#2_0}\hfill
\includegraphics[width=#1]{#2_1}\hfill
\includegraphics[width=#1]{#2_2}\hfill
\includegraphics[width=#1]{#2_3}
}

\newcommand{\figStyleResampling}{
\renewcommand{\h}{0.1145\linewidth}
\renewcommand{\hh}{1ex}
\renewcommand{\hhh}{4ex}
\begin{figure*}[t]
\hspace{5mm}
\hfill
\makebox[\h]{$\bz_1$}\hfill\makebox[\h]{$\bz_2$}\hfill\makebox[\h]{$\bz_3$}\hfill\makebox[\h]{$\bz_4$}%
\hspace{2.25mm}
\makebox[\h]{$\bz_1$}\hfill\makebox[\h]{$\bz_2$}\hfill\makebox[\h]{$\bz_3$}\hfill\makebox[\h]{$\bz_4$}\\[0.5mm]
\resizebox{!}{\h}{\rotatebox{90}{\parbox[c][\hhh][c]{1.5cm}{\centering \scriptsize Layer 1}}}\hfill
\imageRowFour{\h}{biggan_style_resampling/style_resample_husky_layer0}\hspace{\hh}\imageRowFour{\h}{biggan_style_resampling/style_resample_church_layer0}
\resizebox{!}{\h}{\rotatebox{90}{\parbox[c][\hhh][c]{1.5cm}{\centering \tiny Applied from\\\scriptsize Layer 4\\\tiny onward}}}\hfill
\imageRowFour{\h}{biggan_style_resampling/style_resample_husky_layer3}\hspace{\hh}\imageRowFour{\h}{biggan_style_resampling/style_resample_church_layer3}
\resizebox{!}{\h}{\rotatebox{90}{\parbox[c][\hhh][c]{1.5cm}{\centering \scriptsize Layer 8}}}\hfill
\imageRowFour{\h}{biggan_style_resampling/style_resample_husky_layer7}\hspace{\hh}\imageRowFour{\h}{biggan_style_resampling/style_resample_church_layer7}\\ \vspace{-\baselineskip}  
\caption{\label{fig:BigGANStyles}Style variation in BigGAN. Changing the latent vector in BigGAN in the middle of the network alters the style of the generated image. The images on the top row are generated from a base latent (not shown) by substituting $\bz_1\hdots\bz_4$ in its place from layer 1 onwards (resp. from layer 4 and 8 onwards on the following rows). Early changes affect the entire image, while later changes produce more local and subtle variations. 
Notably, comparing the dog and church images on the last row reveals the latents have class-agnostic effects on color and texture.}
\end{figure*}
}

\newcommand{\sidelabeltwo}[2]{\rotatebox{90}{\parbox[t][1.5em][c]{#1}{\tiny \centering {#2}}}}
\newcommand{\figTeaser}{
\renewcommand{\h}{0.1825\linewidth}
\renewcommand{\hh}{1.5cm}
\newcommand{\teaserlabelsize}{\footnotesize}
\begin{figure*}[t]
\newcommand{\baseimg}{teaser/stylegan2_cars/teaser_440749230}
\sidelabeltwo{1.5cm}{StyleGAN2 Cars}\hfill%
\imagewithedit{\h}{\baseimg_0}{}\hfill
\imagewithedit{\h}{\baseimg_1}{\editv{22}{9-10}}\hfill
\imagewithedit{\h}{\baseimg_2}{\editv{41}{9-10}}\hfill
\imagewithedit{\h}{\baseimg_3}{\editv{0}{0-4}}\hfill
\imagewithedit{\h}{\baseimg_4}{\editv{16}{3-5}}\vspace{-1mm}\\
\makebox[1.5em]{}\hfill%
\makebox[\h]{\teaserlabelsize Initial image}\hfill
\makebox[\h]{\teaserlabelsize change color}\hfill
\makebox[\h]{\teaserlabelsize add grass}\hfill
\makebox[\h]{\teaserlabelsize rotate}\hfill
\makebox[\h]{\teaserlabelsize change type}\vspace{1mm}\\
\renewcommand{\baseimg}{teaser/stylegan2_ffhq/teaser_6293435}
\sidelabeltwo{\h}{StyleGAN2\\FFHQ}\hfill%
\imagewithedit{\h}{\baseimg_0}{}\hfill 
\imagewithedit{\h}{\baseimg_1}{\editv{20}{6}}\hfill 
\imagewithedit{\h}{\baseimg_2}{\editv{57}{7-8}}\hfill 
\imagewithedit{\h}{\baseimg_3}{\editv{23}{3-5}}\hfill 
\imagewithedit{\h}{\baseimg_4}{\editv{27}{8-17}}\vspace{-1mm}\\
\makebox[1.5em]{}\hfill%
\makebox[\h]{\teaserlabelsize Initial image}\hfill
\makebox[\h]{\teaserlabelsize add wrinkles}\hfill
\makebox[\h]{\teaserlabelsize hair color}\hfill
\makebox[\h]{\teaserlabelsize expression}\hfill
\makebox[\h]{\teaserlabelsize overexpose}\vspace{1mm}\\
\renewcommand{\baseimg}{teaser/biggan512_irish_setter/teaser_489408325}
\sidelabeltwo{\h}{BigGAN512-deep\\Irish setter}\hfill%
\imagewithedit{\h}{\baseimg_0}{}\hfill 
\imagewithedit{\h}{\baseimg_1}{\editu{3}{all}}\hfill 
\imagewithedit{\h}{\baseimg_2}{\editu{12}{all}}\hfill 
\imagewithedit{\h}{\baseimg_3}{\editu{15}{1-5}}\hfill 
\imagewithedit{\h}{\baseimg_4}{\editu{61}{4-7}}\vspace{-1mm}\\
\makebox[1.5em]{}\hfill%
\makebox[\h]{\teaserlabelsize Initial image}\hfill
\makebox[\h]{\teaserlabelsize rotate}\hfill
\makebox[\h]{\teaserlabelsize zoom out}\hfill
\makebox[\h]{\teaserlabelsize show horizon}\hfill
\makebox[\h]{\teaserlabelsize change scenery}

\caption{\label{fig:Teaser}
Sequences of image edits performed using control discovered with our method, applied to three different GANs. The white insets specify the edits using notation explained in Section~\ref{sec:layerwise}.
}
\end{figure*}
}

\newcommand{\figStyleMixing}{
\begin{figure*}[!b]
\includegraphics[width=\linewidth]{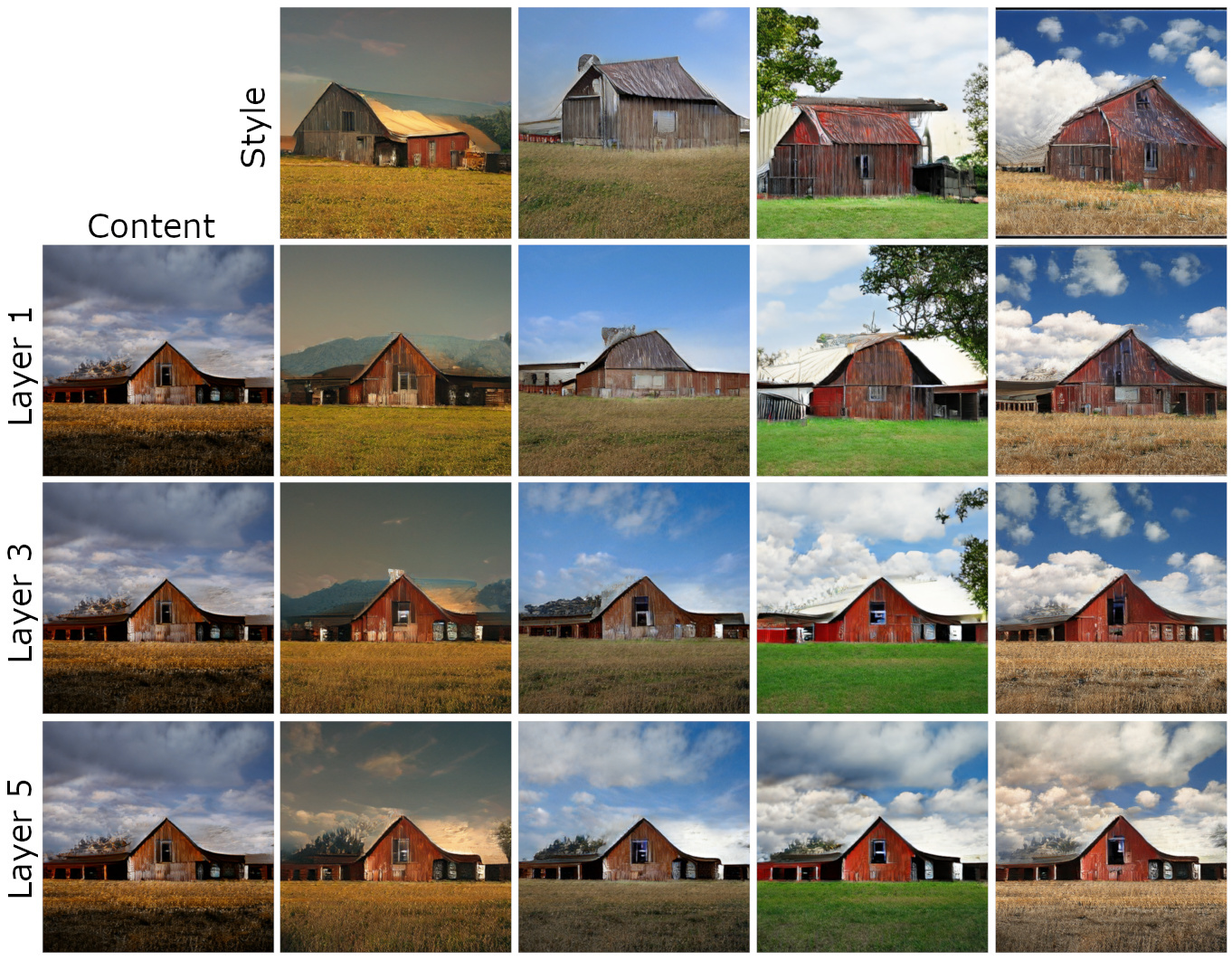}
\caption{\label{fig:BigganStyleMixing}Even though not explicitly trained to do so, BigGAN displays similar style-mixing characteristics to StyleGAN. Here, the latent vector of the content image is swapped for that of the style image starting at different layers.}
\end{figure*}
}

\newcommand{\figEditTransferability}{
\renewcommand{\h}{0.15\linewidth}
\renewcommand{\hh}{5mm}
\begin{figure*}[t]

\newcommand{\baseimg}{biggan_edit_transferability/geom_husky}
\makebox[\hh]{}\hfill
\makebox[\h]{Base}\hfill
\makebox[\h]{\editu{0}{all}}\hfill
\makebox[\h]{\editu{6}{all}}\hfill
\makebox[\hh]{}\hfill
\makebox[\h]{Base}\hfill
\makebox[\h]{\editu{54}{7-9}}\hfill
\makebox[\h]{\editu{33}{7-9}}\\
\rotatebox{90}{\parbox[c][\hh][c]{\h}{\centering Husky}}\hfill
\includegraphics[width=\h]{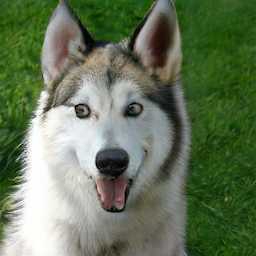}\hfill
\includegraphics[width=\h]{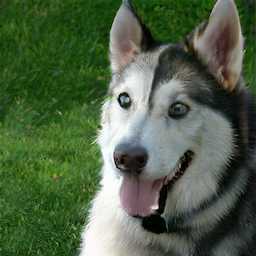}\hfill
\includegraphics[width=\h]{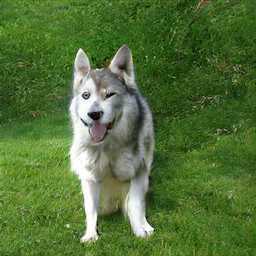}\hfill
\renewcommand{\baseimg}{biggan_edit_transferability/style_lighthouse}
\rotatebox{90}{\parbox[c][\hh][c]{\h}{\centering Lighthouse}}\hfill
\includegraphics[width=\h]{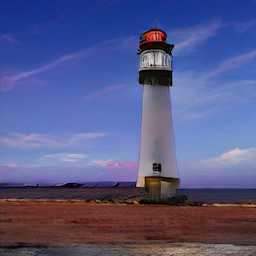}\hfill
\includegraphics[width=\h]{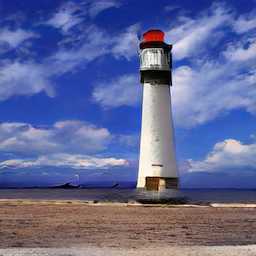}\hfill
\includegraphics[width=\h]{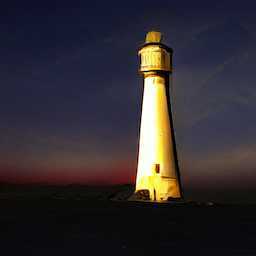}

\renewcommand{\baseimg}{biggan_edit_transferability/geom_castle}
\rotatebox{90}{\parbox[c][\hh][c]{\h}{\centering Castle}}\hfill
\includegraphics[width=\h]{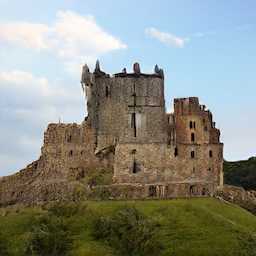}\hfill
\includegraphics[width=\h]{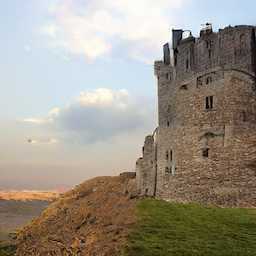}\hfill
\includegraphics[width=\h]{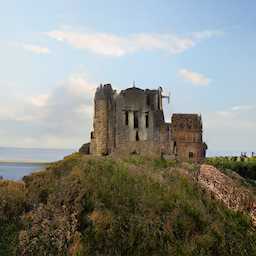}\hfill
\renewcommand{\baseimg}{biggan_edit_transferability/style_barn}
\rotatebox{90}{\parbox[c][\hh][c]{\h}{\centering Barn}}\hfill
\includegraphics[width=\h]{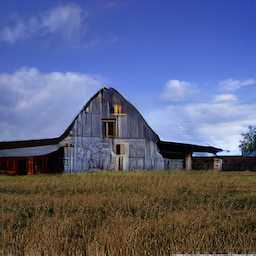}\hfill
\includegraphics[width=\h]{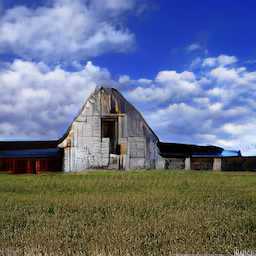}\hfill
\includegraphics[width=\h]{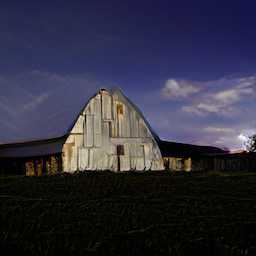}\vspace{-2mm}

\caption{\label{fig:Transferability}The latent space directions we discover often generalize between BigGAN classes. \emph{Left three columns:} Component 0 corresponds to translation and component 6 to zoom. The edit is applied globally to all layers. \emph{Right three columns:} Some later components, when applied to a subset of the layers, control specific textural aspects such as clouds or nighttime illumination of a central object. The components shown where all computed from the husky class.}
\end{figure*}
}

\newcommand{\sidelabelfour}[1]{\rotatebox{90}{\parbox[t][0.5em][c]{\h}{\centering #1}}}

\newcommand{\figSteerabilityComp}{
\renewcommand{\h}{0.095\linewidth}
\renewcommand{\hh}{0.49\linewidth}
\renewcommand{\hhh}{0.2\linewidth}
\begin{figure*}[tb]
\makebox[0.75em]{}\hfill
\makebox[\hh]{(a) BigGAN-512 Zoom \editu{6}{all}}\hfill
\makebox[\hh]{(b) BigGAN-512 Translate \editu{0}{all}}\\
\sidelabelfour{Ours}\hfill%
\imageRowFive{\h}{steerability_comparison/robin/zoom_ours_560157313}
\imageRowFive{\h}{steerability_comparison/golden_retriever/translate_ours_552411435}\\
\sidelabelfour{\cite{gansteerability}}\hfill%
\imageRowFive{\h}{steerability_comparison/robin/zoom_supervised_560157313}
\imageRowFive{\h}{steerability_comparison/golden_retriever/translate_supervised_552411435}\\
\makebox[\hh]{(c) StyleGAN1 FFHQ Pose \editv{9}{0-6}}\hfill
\makebox[\hh]{(d) StyleGAN1 FFHQ Smile \editv{44}{3}}\\
\sidelabelfour{Ours}\hfill%
\imageRowFive{\h}{steerability_comparison/ffhq/pose_ours_129888612}
\imageRowFive{\h}{steerability_comparison/ffhq/smile_ours_70163682}\\
\sidelabelfour{\cite{shen2019interpreting}}\hfill%
\imageRowFive{\h}{steerability_comparison/ffhq/pose_supervised_129888612}
\imageRowFive{\h}{steerability_comparison/ffhq/smile_supervised_70163682}\\
\makebox[\hh]{(e) StyleGAN1 CelebaHQ Glasses \editv{5}{0}}\hfill
\makebox[\hh]{(f) StyleGAN1 CelebaHQ Gender \editv{1}{0-1}}\\
\sidelabelfour{Ours}\hfill%
\imageRowFive{\h}{steerability_comparison/celebahq/glasses_ours_1919124025}
\imageRowFive{\h}{steerability_comparison/celebahq/gender_ours_264878205}\\
\sidelabelfour{\cite{shen2019interpreting}}\hfill%
\imageRowFive{\h}{steerability_comparison/celebahq/glasses_supervised_1919124025}
\imageRowFive{\h}{steerability_comparison/celebahq/gender_supervised_264878205}\\
\caption{\label{fig:SteerabilityComparison}
Comparison of edit directions found through PCA to those found in previous work using supervised methods~\cite{gansteerability,shen2019interpreting}. Some are visually very close~(a,~c). Others achieve a variant of the same effect~(d,~e,~f), sometimes with more entanglement~(d), and sometimes less~(b). In some cases, both produce highly entangled effects~(a,~f).  We also observe a few cases where strong effects introduce inconsistencies~(e) in our outputs. Still, the results are remarkably close given that our approach does not specify target transformations or use supervised learning. The corresponding edits were found manually using our interactive exploration software.
}
\end{figure*}
}

\renewcommand{\sidelabelfour}[1]{\rotatebox{90}{\parbox[t][0.5em][c]{\h}{\centering #1}}}
\newcommand{\figSteerabilityCompSupplementBigGAN}{
\renewcommand{\h}{0.095\linewidth}
\renewcommand{\hh}{0.49\linewidth}
\begin{figure*}[tb]
\makebox[0.75em]{}\hfill

\makebox[\hh]{(a) Zoom \editu{6}{all}}\hfill
\makebox[\hh]{(b) Translate \editu{0}{all}}\\
\sidelabelfour{Ours}\hfill%
\imageRowFive{\h}{steerability_comparison/ship/zoom_ours_107715983}
\imageRowFive{\h}{steerability_comparison/lemon/translate_ours_331582800}\\
\sidelabelfour{\cite{gansteerability}}\hfill%
\imageRowFive{\h}{steerability_comparison/ship/zoom_supervised_107715983}
\imageRowFive{\h}{steerability_comparison/lemon/translate_supervised_331582800}\\

\makebox[\hh]{(a) FFHQ Blueness \editu{2}{17}}\hfill
\makebox[\hh]{(b) FFHQ Greenness \editu{1}{17}}\\
\sidelabelfour{Ours}\hfill%
\imageRowFive{\h}{steerability_comparison/ffhq/blue_ours_5}
\imageRowFive{\h}{steerability_comparison/ffhq/green_ours_5}\\
\sidelabelfour{\cite{gansteerability}}\hfill%
\imageRowFive{\h}{steerability_comparison/ffhq/blue_supervised_5}
\imageRowFive{\h}{steerability_comparison/ffhq/green_supervised_5}\\

\makebox[\hh]{(a) Rotate \editu{0}{0}}\hfill
\makebox[\hh]{(b) ShiftY \editu{7}{1}}\\
\sidelabelfour{Ours}\hfill%
\imageRowFive{\h}{steerability_comparison/cars/rotate2d_ours_28}
\imageRowFive{\h}{steerability_comparison/cars/shifty_ours_0}\\
\sidelabelfour{\cite{gansteerability}}\hfill%
\imageRowFive{\h}{steerability_comparison/cars/rotate2d_supervised_28}
\imageRowFive{\h}{steerability_comparison/cars/shifty_supervised_0}\\
\sidelabelfour{Ours}\hfill%
\imageRowFive{\h}{steerability_comparison/cars/rotate2d_ours_46}
\imageRowFive{\h}{steerability_comparison/cars/shifty_ours_13}\\
\sidelabelfour{\cite{gansteerability}}\hfill%
\imageRowFive{\h}{steerability_comparison/cars/rotate2d_supervised_46}
\imageRowFive{\h}{steerability_comparison/cars/shifty_supervised_13}\\

\caption{\label{fig:SteerabilityCompSupplementBigGAN}Comparisons against \cite{gansteerability} for BigGAN512-deep, StyleGAN FFHQ, and StyleGAN Cars.}
\end{figure*}
}

\renewcommand{\sidelabelfour}[1]{\rotatebox{90}{\parbox[t][0.5em][c]{\h}{\centering #1}}}
\newcommand{\figSteerabilityCompSupplementFFHQ}{
\renewcommand{\h}{0.095\linewidth}
\renewcommand{\hh}{0.49\linewidth}
\begin{figure*}[tb]
\makebox[0.75em]{}\hfill

\makebox[\hh]{(c) FFHQ Pose \editv{9}{0-6}}\hfill
\makebox[\hh]{(d) FFHQ Gender \editv{0}{2-5}}\\
\sidelabelfour{Ours}\hfill%
\imageRowFive{\h}{steerability_comparison/ffhq/pose_ours_440608316}
\imageRowFive{\h}{steerability_comparison/ffhq/gender_ours_1302836080}\\
\sidelabelfour{\cite{shen2019interpreting}}\hfill%
\imageRowFive{\h}{steerability_comparison/ffhq/pose_supervised_440608316}
\imageRowFive{\h}{steerability_comparison/ffhq/gender_supervised_1302836080}\\
\sidelabelfour{Ours}\hfill%
\imageRowFive{\h}{steerability_comparison/ffhq/pose_ours_1811098088}
\imageRowFive{\h}{steerability_comparison/ffhq/gender_ours_1746672325}\\
\sidelabelfour{\cite{shen2019interpreting}}\hfill%
\imageRowFive{\h}{steerability_comparison/ffhq/pose_supervised_1811098088}
\imageRowFive{\h}{steerability_comparison/ffhq/gender_supervised_1746672325}\\

\makebox[\hh]{(e) FFHQ Smile \editv{44}{3}}\hfill
\makebox[\hh]{(f) FFHQ Glasses \editv{12}{0-1}}\\
\sidelabelfour{Ours}\hfill%
\imageRowFive{\h}{steerability_comparison/ffhq/smile_ours_1647189561}
\imageRowFive{\h}{steerability_comparison/ffhq/glasses_ours_1110182583}\\
\sidelabelfour{\cite{shen2019interpreting}}\hfill%
\imageRowFive{\h}{steerability_comparison/ffhq/smile_supervised_1647189561}
\imageRowFive{\h}{steerability_comparison/ffhq/glasses_supervised_1110182583}\\
\sidelabelfour{Ours}\hfill%
\imageRowFive{\h}{steerability_comparison/ffhq/smile_ours_1759734403}
\imageRowFive{\h}{steerability_comparison/ffhq/glasses_ours_1005764659}\\
\sidelabelfour{\cite{shen2019interpreting}}\hfill%
\imageRowFive{\h}{steerability_comparison/ffhq/smile_supervised_1759734403}
\imageRowFive{\h}{steerability_comparison/ffhq/glasses_supervised_1005764659}\\

\caption{\label{fig:SteerabilityCompSupplementFFHQ}Edits found with our method compared to those found by \cite{shen2019interpreting} for the StyleGAN FFHQ model.}
\end{figure*}
}

\renewcommand{\sidelabelfour}[1]{\rotatebox{90}{\parbox[t][0.5em][c]{\h}{\centering #1}}}
\newcommand{\figSteerabilityCompSupplementCelebaHQ}{
\renewcommand{\h}{0.095\linewidth}
\renewcommand{\hh}{0.49\linewidth}
\begin{figure*}[tb]
\makebox[0.75em]{}\hfill

\makebox[\hh]{(g) CelebaHQ Pose \editv{7}{0-6}}\hfill
\makebox[\hh]{(h) CelebaHQ Gender \editv{1}{0-1}}\\
\sidelabelfour{Ours}\hfill%
\imageRowFive{\h}{steerability_comparison/celebahq/pose_ours_329555154}
\imageRowFive{\h}{steerability_comparison/celebahq/gender_ours_967075839}\\
\sidelabelfour{\cite{shen2019interpreting}}\hfill%
\imageRowFive{\h}{steerability_comparison/celebahq/pose_supervised_329555154}
\imageRowFive{\h}{steerability_comparison/celebahq/gender_supervised_967075839}\\
\sidelabelfour{Ours}\hfill%
\imageRowFive{\h}{steerability_comparison/celebahq/pose_ours_1460055449}
\imageRowFive{\h}{steerability_comparison/celebahq/gender_ours_1144615644}\\
\sidelabelfour{\cite{shen2019interpreting}}\hfill%
\imageRowFive{\h}{steerability_comparison/celebahq/pose_supervised_1460055449}
\imageRowFive{\h}{steerability_comparison/celebahq/gender_supervised_1144615644}\\

\makebox[\hh]{(i) CelebaHQ Smile \editv{14}{3}}\hfill
\makebox[\hh]{(j) CelebaHQ Glasses \editv{5}{0}}\\
\sidelabelfour{Ours}\hfill%
\imageRowFive{\h}{steerability_comparison/celebahq/smile_ours_424805522}
\imageRowFive{\h}{steerability_comparison/celebahq/glasses_ours_991993380}\\
\sidelabelfour{\cite{shen2019interpreting}}\hfill%
\imageRowFive{\h}{steerability_comparison/celebahq/smile_supervised_424805522}
\imageRowFive{\h}{steerability_comparison/celebahq/glasses_supervised_991993380}\\
\sidelabelfour{Ours}\hfill%
\imageRowFive{\h}{steerability_comparison/celebahq/smile_ours_329187806}
\imageRowFive{\h}{steerability_comparison/celebahq/glasses_ours_594344173}\\
\sidelabelfour{\cite{shen2019interpreting}}\hfill%
\imageRowFive{\h}{steerability_comparison/celebahq/smile_supervised_329187806}
\imageRowFive{\h}{steerability_comparison/celebahq/glasses_supervised_594344173}\\

\caption{\label{fig:SteerabilityCompSupplementCelebaHQ}Edits found with our method compared to those found by \cite{shen2019interpreting} for the StyleGAN CelebaHQ model}
\end{figure*}
}

\newcommand{\figPCADiagram}{
\begin{figure*}[t]
\centering
\includegraphics[width=4in]{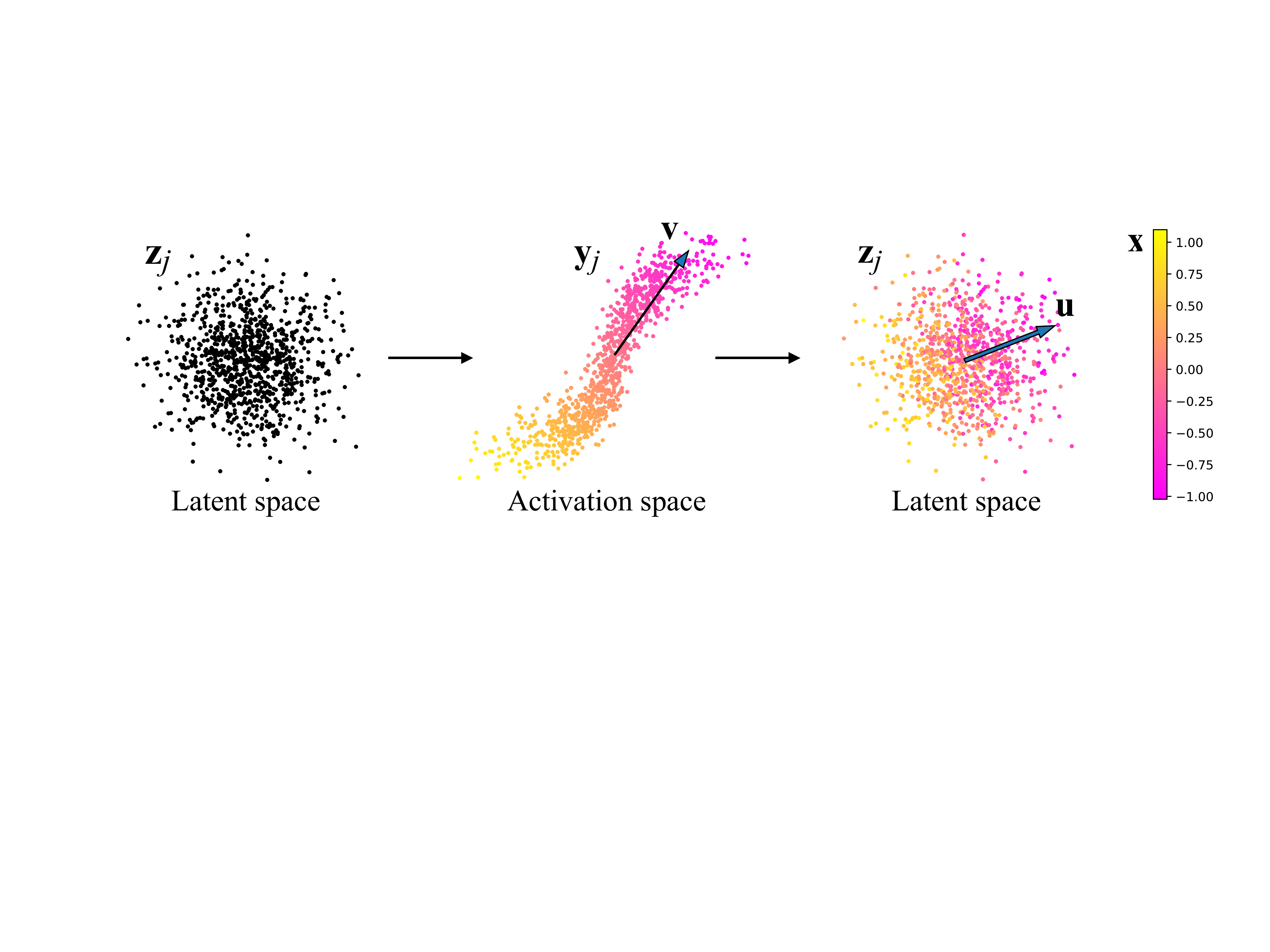}
\caption{\label{fig:PCAdiagram} 
2D Illustration of identifying a principal activation direction for BigGAN.
Random latent vectors $\bz_j$ are sampled, and converted to activations $\by_j$. The PCA direction $\bv$ is computed from the samples, and PCA coordinates $x_j$ computed, shown here by color-coding. Finally, back in the latent space, the direction $\bu$ is computed by regression from $\bz_j$ to $x_j$.
}
\end{figure*}
}

\renewcommand{\sidelabel}[1]{\rotatebox{90}{\parbox[t][1.75em][b]{\h}{\tiny \centering {#1}}}}
\newcommand{\figPCACleanup}{
\renewcommand{\h}{0.126\linewidth}
\renewcommand{\hh}{0.10\linewidth}
\begin{figure*}[h!]
\centering
\sidelabel{\editv{0}{all}\\gender\strut}%
\imageRowSevenFramed{\h}{pca_cleanup/global/366745668_pc0}{opacity=0}{red}{opacity=0}{opacity=0}{opacity=0}{opacity=0}{opacity=0}
\sidelabel{\editv{1}{all}\\rotate+gender\strut}%
\imageRowSevenFramed{\h}{pca_cleanup/global/366745668_pc1}{opacity=0}{opacity=0}{opacity=0}{red}{opacity=0}{opacity=0}{opacity=0}\\
\sidelabel{\editv{2}{all}\\rotate, age, gender, bkg\strut}%
\imageRowSevenFramed{\h}{pca_cleanup/global/366745668_pc2}{opacity=0}{red}{opacity=0}{opacity=0}{opacity=0}{opacity=0}{opacity=0}\\
\smallskip
\hrule
\smallskip
\sidelabel{\editv{1}{0-2}\\mostly rotate\strut}%
\imageRowSevenFramed{\h}{pca_cleanup/tuned/366745668_pc1_s0_e3}{opacity=0}{opacity=0}{opacity=0}{red}{opacity=0}{opacity=0}{opacity=0}\\
\sidelabel{\editv{10}{7-8}\\hair color\strut}%
\imageRowSevenFramed{\h}{pca_cleanup/tuned/366745668_pc10_s7_e9}{opacity=0}{opacity=0}{opacity=0}{opacity=0}{red}{opacity=0}{opacity=0}\vspace{-2mm}
  
\caption{\label{fig:PCs}\emph{Rows 1-3} illustrate the three largest principal components in the intermediate $\mathcal{W}$ latent space of StyleGAN2. They span the major variations expected of portrait photographs---such as gender and head rotation---with a few effects typically entangled together. The red square corresponds to location of the original image on each principal axis. \emph{Rows 4-5} demonstrate the effect of constraining the variation to a subset of the layers. For example, restricting the 2nd component to only layers 0-2, denoted \mbox{\editv{1}{0-2}}, leaves a relatively pure head rotation that changes gender expression and identity less (compare to 2nd row). Similarly, selective application of the principal components allows control of features such as hair color, aspects of hairstyle, and lighting. See \SMref{1} for a larger sampling.}%
\end{figure*}
}

\renewcommand{\sidelabel}[1]{\rotatebox{90}{\parbox[t][1.75em][b]{\h}{\tiny \centering {#1}}}}

\newcommand{\sidelabelthree}[2]{\resizebox{!}{#1}{\rotatebox{90}{\parbox[t][1.5em][c]{3.5cm}{\centering {#2}}}}}
\newcommand{\figEditZoo}{
\renewcommand{\h}{0.0925\linewidth}
\begin{figure*}[t]
\sidelabelthree{\h}{StyleGAN2 Cars\\\editv{18}{7-8}\\reflections}\hfill%
\imageRowFive{\h}{edit_zoo/StyleGAN2/car/reflections/cmp18_s7_e9_1498448887}
\sidelabelthree{\h}{BigGAN512-deep\\\editu{54}{6-14}\\add clouds}\hfill%
\imageRowFive{\h}{edit_zoo/BigGAN-512/clouds/cmp54_s6_e15_1826867440}
\sidelabelthree{\h}{BigGAN512-deep\\\editu{62}{3-14}\\season}\hfill%
\imageRowFive{\h}{edit_zoo/BigGAN-512/season/cmp62_s3_e15_1162727876}
\sidelabelthree{\h}{StyleGAN2 Cats\\\editv{27}{2-4}\\fluffiness}\hfill%
\imageRowFive{\h}{edit_zoo/StyleGAN2/cat/fluffiness/cmp27_s2_e5_740196857}
\sidelabelthree{\h}{StyleGAN2 Horse\\\editv{3}{3-4}\\rider}\hfill%
\imageRowFive{\h}{edit_zoo/StyleGAN2/horse/person/cmp3_s3_e5_944988831}
\sidelabelthree{\h}{StyleGAN2 FFHQ\\\editv{0}{8}\\makeup}\hfill%
\imageRowFive{\h}{edit_zoo/StyleGAN2/ffhq/makeup/cmp0_s8_e9_266415229}
\sidelabelthree{\h}{StyleGAN WikiArt\\\editv{7}{0-1}\\rotation}\hfill%
\imageRowFive{\h}{edit_zoo/StyleGAN/wikiart/head_rotation/cmp7_s0_e2_1819967864}
\sidelabelthree{\h}{StyleGAN WikiArt\\\editv{9}{8-14}\\stroke style}\hfill%
\imageRowFive{\h}{edit_zoo/StyleGAN/wikiart/simple_strokes/cmp9_s8_e15_1239190942}\\
\vspace{-2mm}\caption{\label{fig:EditZoo}A selection of interpretable edits discovered by selective application of latent edits across the layers of several pretrained GAN models. The reader is encouraged to zoom in on an electronic device. A larger selection is available in \SMref{1}.
}
\end{figure*}
}

\renewcommand{\sidelabelthree}[2]{\resizebox{!}{#1}{\rotatebox{90}{\parbox[t][1.5em][c]{3.5cm}{\centering {#2}}}}}
\newcommand{\figEditZooSupplemental}{
\renewcommand{\h}{0.0925\linewidth}
\begin{figure*}[t]
\sidelabelthree{\h}{StyleGAN2 Cars\\\editv{50}{8}\\season}\hfill%
\imageRowFive{\h}{edit_zoo/StyleGAN2/car/autumn/cmp50_s8_e9_329004386}
\sidelabelthree{\h}{StyleGAN2 Cars\\\editv{15}{0-3}\\focal length}\hfill%
\imageRowFive{\h}{edit_zoo/StyleGAN2/car/focal_depth/cmp15_s0_e4_587218105}
\sidelabelthree{\h}{StyleGAN2 Cars\\\editv{44}{0-8}\\car model}\hfill%
\imageRowFive{\h}{edit_zoo/StyleGAN2/car/low_rider/cmp44_s0_e9_1204444821}
\sidelabelthree{\h}{StyleGAN2 Cars\\\editv{18}{7-8}\\reflections}\hfill%
\imageRowFive{\h}{edit_zoo/StyleGAN2/car/reflections/cmp18_s7_e9_1498448887}
\sidelabelthree{\h}{BigGAN512-deep\\\editu{64}{6-9}\\add grass}\hfill%
\imageRowFive{\h}{edit_zoo/BigGAN-512/add_grass/cmp64_s6_e10_20736816}
\sidelabelthree{\h}{BigGAN512-deep\\\editu{54}{6-14}\\add clouds}\hfill%
\imageRowFive{\h}{edit_zoo/BigGAN-512/clouds/cmp54_s6_e15_1826867440}
\sidelabelthree{\h}{BigGAN512-deep\\\editu{37}{6-14}\\day-night}\hfill%
\imageRowFive{\h}{edit_zoo/BigGAN-512/nighttime/cmp37_s6_e15_1202948959}
\sidelabelthree{\h}{BigGAN512-deep\\\editu{62}{3-14}\\season}\hfill%
\imageRowFive{\h}{edit_zoo/BigGAN-512/season/cmp62_s3_e15_1162727876}
\sidelabelthree{\h}{StyleGAN2 Cats\\\editv{45}{5-7}\\eyes open}\hfill%
\imageRowFive{\h}{edit_zoo/StyleGAN2/cat/eyes/cmp45_s5_e8_81011138}
\sidelabelthree{\h}{StyleGAN2 Cats\\\editv{27}{2-4}\\fluffiness}\hfill%
\imageRowFive{\h}{edit_zoo/StyleGAN2/cat/fluffiness/cmp27_s2_e5_740196857}
\sidelabelthree{\h}{StyleGAN2 Horse\\\editv{11}{5-6}\\color}\hfill%
\imageRowFive{\h}{edit_zoo/StyleGAN2/horse/coloring/cmp11_s5_e7_897830797}
\sidelabelthree{\h}{StyleGAN2 Horse\\\editv{3}{3-4}\\rider}\hfill%
\imageRowFive{\h}{edit_zoo/StyleGAN2/horse/person/cmp3_s3_e5_944988831}
\sidelabelthree{\h}{StyleGAN2 Church\\\editv{20}{7-8}\\clouds}\hfill%
\imageRowFive{\h}{edit_zoo/StyleGAN2/church/clouds/cmp20_s7_e9_1360331956}
\sidelabelthree{\h}{StyleGAN2 Church\\\editv{8}{7-8}\\direct sun}\hfill%
\imageRowFive{\h}{edit_zoo/StyleGAN2/church/direct_sunlight/cmp8_s7_e9_1777321344}
\sidelabelthree{\h}{StyleGAN2 Church\\\editv{15}{8}\\sun direction}\hfill%
\imageRowFive{\h}{edit_zoo/StyleGAN2/church/sun_dir/cmp15_s8_e9_485108354}
\sidelabelthree{\h}{StyleGAN2 Church\\\editv{8}{12-13}\\vibrant}\hfill%
\imageRowFive{\h}{edit_zoo/StyleGAN2/church/vibrant/cmp8_s12_e14_373098621}
\sidelabelthree{\h}{StyleGAN Landscapes\\\editv{0}{1-16}\\verticality}\hfill%
\imageRowFive{\h}{edit_zoo/StyleGAN/landscapes/verticality/cmp0_s1_e17_1482275361}
\sidelabelthree{\h}{StyleGAN Landscapes\\\editv{1}{9-17}\\evening}\hfill%
\imageRowFive{\h}{edit_zoo/StyleGAN/landscapes/evening/cmp1_s9_e18_1924605759}
\sidelabelthree{\h}{StyleGAN Bedrooms\\\editv{31}{0-5}\\bed shape}\hfill%
\imageRowFive{\h}{edit_zoo/StyleGAN/bedrooms/height/cmp31_s0_e6_2073683729}
\sidelabelthree{\h}{StyleGAN Bedrooms\\\editv{5}{0-2}\\orientation}\hfill%
\imageRowFive{\h}{edit_zoo/StyleGAN/bedrooms/orientation/cmp5_s0_e3_96357868}
\sidelabelthree{\h}{StyleGAN2 FFHQ\\\editv{43}{6-7}\\smile-disgust}\hfill%
\imageRowFive{\h}{edit_zoo/StyleGAN2/ffhq/disgusted/cmp43_s6_e8_140658858}
\sidelabelthree{\h}{StyleGAN2 FFHQ\\\editv{0}{8}\\makeup}\hfill%
\imageRowFive{\h}{edit_zoo/StyleGAN2/ffhq/makeup/cmp0_s8_e9_266415229}
\sidelabelthree{\h}{StyleGAN WikiArt\\\editv{7}{0-1}\\rotation}\hfill%
\imageRowFive{\h}{edit_zoo/StyleGAN/wikiart/head_rotation/cmp7_s0_e2_1819967864}
\sidelabelthree{\h}{StyleGAN WikiArt\\\editv{9}{8-14}\\stroke style}\hfill%
\imageRowFive{\h}{edit_zoo/StyleGAN/wikiart/simple_strokes/cmp9_s8_e15_1239190942}
\sidelabelthree{\h}{StyleGAN WikiArt\\\editv{59}{9-14}\\skin tone}\hfill%
\imageRowFive{\h}{edit_zoo/StyleGAN/wikiart/skin_color/cmp59_s9_e15_1615931059}
\sidelabelthree{\h}{StyleGAN WikiArt\\\editv{36}{4-6}\\mouth shape}\hfill%
\imageRowFive{\h}{edit_zoo/StyleGAN/wikiart/mouth_shape/cmp36_s4_e7_333293845}
\sidelabelthree{\h}{StyleGAN WikiArt\\\editv{35}{2-3}\\eye spacing}\hfill%
\imageRowFive{\h}{edit_zoo/StyleGAN/wikiart/head_width/cmp35_s2_e4_1213732031}
\sidelabelthree{\h}{StyleGAN WikiArt\\\editv{31}{8-14}\\sharpness}\hfill%
\imageRowFive{\h}{edit_zoo/StyleGAN/wikiart/sharpness/cmp31_s8_e15_1489906162}%
\\

\caption{\label{fig:EditZooSupp}A selection of interpretable edits discovered by selective application of latent edits across the layers of several pretrained GAN models. The reader is encouraged to zoom in on an electronic device.}
\end{figure*}
}

\newcommand{\figRandomBaseline}{
\renewcommand{\h}{0.5\linewidth}
\renewcommand{\hh}{0.16\linewidth}

\begin{figure*}[t]
\footnotesize
\imageBlockSixA{\hh}{random_baseline/StyleGAN2_cat/keep_8_first_1866827965}
\imageBlockSixA{\hh}{random_baseline/StyleGAN2_cat/randomize_8_first_1866827965}\\
\imageBlockSixB{\hh}{random_baseline/StyleGAN2_cat/keep_8_first_1866827965}
\imageBlockSixB{\hh}{random_baseline/StyleGAN2_cat/randomize_8_first_1866827965}\\
\makebox[\h]{(a) Fix first 8 PCA coord., randomize remaining 504}
\makebox[\h]{(b) Randomize first 8 PCA coord., fix remaining 504}\\
\makebox[\h]{(Pose and camera fixed, appearance changes)}
\makebox[\h]{(Appearance fixed, pose changes)}\\

\imageBlockSixA{\hh}{random_baseline/StyleGAN2_cat/keep_8_first_random_1866827965}
\imageBlockSixA{\hh}{random_baseline/StyleGAN2_cat/randomize_8_first_random_1866827965}\\
\imageBlockSixB{\hh}{random_baseline/StyleGAN2_cat/keep_8_first_random_1866827965}
\imageBlockSixB{\hh}{random_baseline/StyleGAN2_cat/randomize_8_first_random_1866827965}\\
\makebox[\h]{(c) Fix 8 random basis coord., randomize the others}
\makebox[\h]{(d) Randomize 8 random basis coord., fix remaining 504}\hfill
\makebox[\h]{(Almost everything changes)}
\makebox[\h]{(Almost nothing changes)}
\caption{\label{fig:RandomBaseline}Illustration of the significance of the principal components as compared to random directions in the intermediate latent space $\mathcal{W}$ of StyleGAN2. Fixing and randomizing the early principal components shows a separation between pose and style (a, b). In contrast, fixing and randomizing randomly-chosen directions does not yield a similar meaningful decomposition (c, d).}
\end{figure*}
}

\newcommand{\figRandomBaselineCar}{
\renewcommand{\h}{0.5\linewidth}
\renewcommand{\hh}{0.16\linewidth}

\begin{figure*}[t]
\footnotesize

\makebox[\linewidth]{StyleGAN2 car}
\imageBlockSixA{\hh}{random_baseline/StyleGAN2_car/keep_5_first_1257084100}
\imageBlockSixA{\hh}{random_baseline/StyleGAN2_car/randomize_5_first_1257084100}\\
\imageBlockSixB{\hh}{random_baseline/StyleGAN2_car/keep_5_first_1257084100}
\imageBlockSixB{\hh}{random_baseline/StyleGAN2_car/randomize_5_first_1257084100}\\
\makebox[\h]{(a) Fix first 5 PCA coord., randomize rest}
\makebox[\h]{(b) Randomize first 5 PCA coord., fix rest}\\
\imageBlockSixA{\hh}{random_baseline/StyleGAN2_car/keep_5_first_random_1257084100}
\imageBlockSixA{\hh}{random_baseline/StyleGAN2_car/randomize_5_first_random_1257084100}\\
\imageBlockSixB{\hh}{random_baseline/StyleGAN2_car/keep_5_first_random_1257084100}
\imageBlockSixB{\hh}{random_baseline/StyleGAN2_car/randomize_5_first_random_1257084100}\\
\makebox[\h]{(c) Fix 5 random basis coord., randomize rest}
\makebox[\h]{(d) Randomize 5 random basis coord., fix rest}\hfill

\caption{\label{fig:RandomBaselineCar}The PCA basis displays a content-style separation not present in random bases.}
\end{figure*}
}

\newcommand{\figRandomBaselineDuck}{
\renewcommand{\h}{0.5\linewidth}
\renewcommand{\hh}{0.16\linewidth}

\begin{figure*}[t]
\footnotesize

\makebox[\linewidth]{BigGAN256-deep duck}
\imageBlockSixA{\hh}{random_baseline/BigGAN-256_duck/keep_10_first_1134462557}
\imageBlockSixA{\hh}{random_baseline/BigGAN-256_duck/randomize_10_first_1134462557}\\
\imageBlockSixB{\hh}{random_baseline/BigGAN-256_duck/keep_10_first_1134462557}
\imageBlockSixB{\hh}{random_baseline/BigGAN-256_duck/randomize_10_first_1134462557}\\
\makebox[\h]{(a) Fix first 10 PCA coord., randomize rest}
\makebox[\h]{(b) Randomize first 10 PCA coord., fix rest}\\
\imageBlockSixA{\hh}{random_baseline/BigGAN-256_duck/keep_10_first_random_1134462557}
\imageBlockSixA{\hh}{random_baseline/BigGAN-256_duck/randomize_10_first_random_1134462557}\\
\imageBlockSixB{\hh}{random_baseline/BigGAN-256_duck/keep_10_first_random_1134462557}
\imageBlockSixB{\hh}{random_baseline/BigGAN-256_duck/randomize_10_first_random_1134462557}\\
\makebox[\h]{(c) Fix 10 random basis coord., randomize rest}
\makebox[\h]{(d) Randomize 10 random basis coord., fix rest}\hfill

\caption{\label{fig:RandomBaselineDuck}The PCA basis displays a content-style separation not present in random bases.}
\end{figure*}
}

\newcommand{\figRandomBaseBedroom}{
\renewcommand{\h}{0.5\linewidth}
\renewcommand{\hh}{0.16\linewidth}

\begin{figure*}[t]
\footnotesize

\makebox[\linewidth]{StyleGAN bedrooms}
\imageBlockSixA{\hh}{random_baseline/StyleGAN_bedrooms/keep_10_first_1382244162}
\imageBlockSixA{\hh}{random_baseline/StyleGAN_bedrooms/randomize_10_first_1382244162}\\
\imageBlockSixB{\hh}{random_baseline/StyleGAN_bedrooms/keep_10_first_1382244162}
\imageBlockSixB{\hh}{random_baseline/StyleGAN_bedrooms/randomize_10_first_1382244162}\\
\makebox[\h]{(a) Fix first 10 PCA coord., randomize rest}
\makebox[\h]{(b) Randomize first 10 PCA coord., fix rest}\\
\imageBlockSixA{\hh}{random_baseline/StyleGAN_bedrooms/keep_10_first_random_1382244162}
\imageBlockSixA{\hh}{random_baseline/StyleGAN_bedrooms/randomize_10_first_random_1382244162}\\
\imageBlockSixB{\hh}{random_baseline/StyleGAN_bedrooms/keep_10_first_random_1382244162}
\imageBlockSixB{\hh}{random_baseline/StyleGAN_bedrooms/randomize_10_first_random_1382244162}\\
\makebox[\h]{(c) Fix 10 random basis coord., randomize rest}
\makebox[\h]{(d) Randomize 10 random basis coord., fix rest}\hfill

\caption{\label{fig:RandomBaselineBedroom}The PCA basis displays a content-style separation not present in random bases.}
\end{figure*}
}

\newcommand{\figRandomBaseFFHQ}{
\renewcommand{\h}{0.5\linewidth}
\renewcommand{\hh}{0.16\linewidth}

\begin{figure*}[t]
\footnotesize

\makebox[\linewidth]{StyleGAN ffhq}
\imageBlockSixA{\hh}{random_baseline/StyleGAN_ffhq/keep_10_first_598174413}
\imageBlockSixA{\hh}{random_baseline/StyleGAN_ffhq/randomize_10_first_598174413}\\
\imageBlockSixB{\hh}{random_baseline/StyleGAN_ffhq/keep_10_first_598174413}
\imageBlockSixB{\hh}{random_baseline/StyleGAN_ffhq/randomize_10_first_598174413}\\
\makebox[\h]{(a) Fix first 10 PCA coord., randomize rest}
\makebox[\h]{(b) Randomize first 10 PCA coord., fix rest}\\
\imageBlockSixA{\hh}{random_baseline/StyleGAN_ffhq/keep_10_first_random_598174413}
\imageBlockSixA{\hh}{random_baseline/StyleGAN_ffhq/randomize_10_first_random_598174413}\\
\imageBlockSixB{\hh}{random_baseline/StyleGAN_ffhq/keep_10_first_random_598174413}
\imageBlockSixB{\hh}{random_baseline/StyleGAN_ffhq/randomize_10_first_random_598174413}\\
\makebox[\h]{(c) Fix 10 random basis coord., randomize rest}
\makebox[\h]{(d) Randomize 10 random basis coord., fix rest}\hfill

\caption{\label{fig:RandomBaselineFFHQ}The first few principal components often encode style changes in addition to geometry in spatially aligned datasets, as seen in the change of identity in the top right quadrant.}
\end{figure*}
}

\newcommand{\pcarowless}[2]{\includegraphics[width=#1]{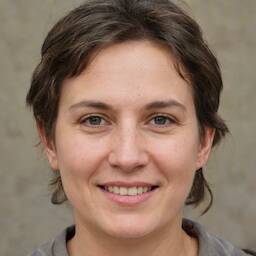}\hfill\includegraphics[width=#1]{reduced_images/gan_#2_1.jpg}\hfill\includegraphics[width=#1]{reduced_images/gan_#2_5.jpg}\hfill\includegraphics[width=#1]{reduced_images/gan_#2_10.jpg}\hfill\includegraphics[width=#1]{reduced_images/gan_#2_20.jpg}\hfill\includegraphics[width=#1]{reduced_images/gan_#2_100.jpg}\hfill\includegraphics[width=#1]{reduced_images/gan_#2_512.jpg}}

\newcommand{\figPCAReduced}
{
\begin{figure}
    \centering
    \pcarowless{0.75in}{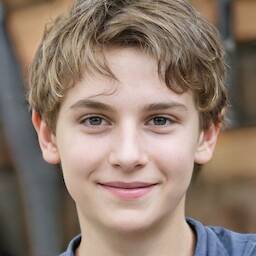} \\
    \pcarowless{0.75in}{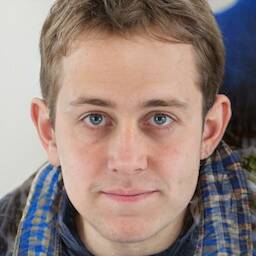} \\
    \pcarowless{0.75in}{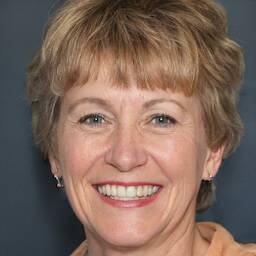} \\
    \pcarowless{0.75in}{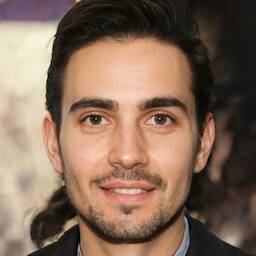} \\
    \pcarowless{0.75in}{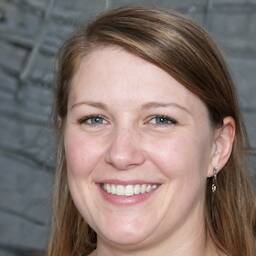} \\
    \pcarowless{0.75in}{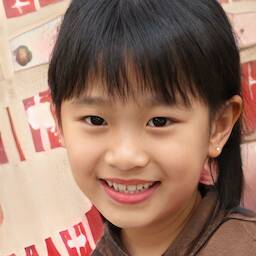} \\
    \pcarowless{0.75in}{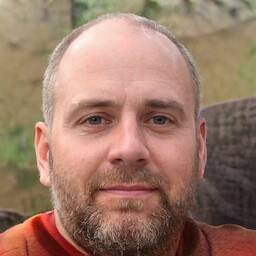} \\
    \pcarowless{0.75in}{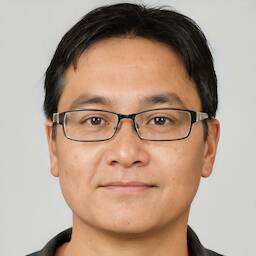} \\
    \pcarowless{0.75in}{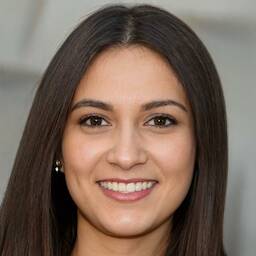} \\
    \makebox[0.75in]{0}\hfill\makebox[0.75in]{1}\hfill\makebox[0.75in]{5}\hfill\makebox[0.75in]{10}\hfill\makebox[0.75in]{20}\hfill\makebox[0.75in]{100}\hfill\makebox[0.75in]{512}\hfill\\
    \caption{Randomly sampled images, projected onto reduced numbers of PCA dimensions: 0, 1, 5, 10, 20, 100, 512 (full dimensional).}
    \label{fig:reduced_images}
\end{figure}
}

\newcommand{\fwid}{0.75in}

\newcommand{\figCombiningEdits}{
\begin{figure}
\centering
	\includegraphics[width=\fwid]{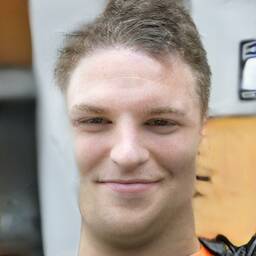}\hfill%
	\includegraphics[width=\fwid]{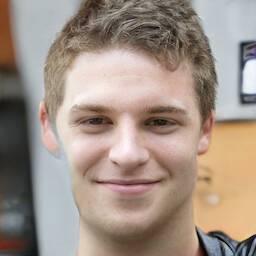}\hfill%
	\includegraphics[width=\fwid]{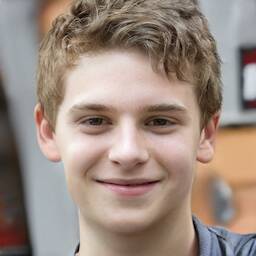}\hfill%
	\includegraphics[width=\fwid]{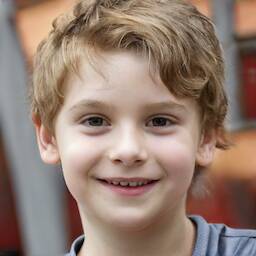}\hfill%
	\includegraphics[width=\fwid]{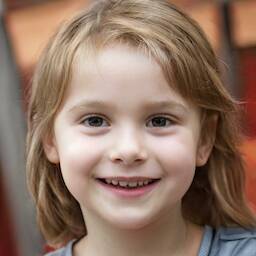}\hfill%
	\includegraphics[width=\fwid]{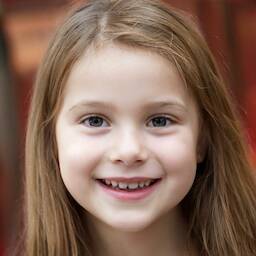}\hfill%
	\includegraphics[width=\fwid]{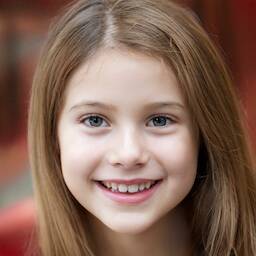}
\\
	\includegraphics[width=\fwid]{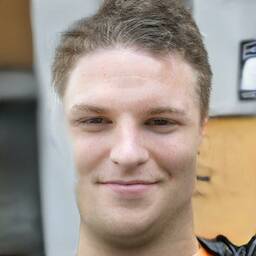}\hfill%
	\includegraphics[width=\fwid]{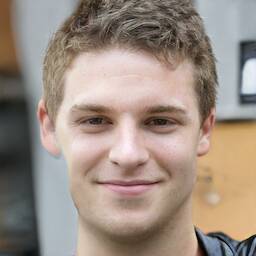}\hfill%
	\includegraphics[width=\fwid]{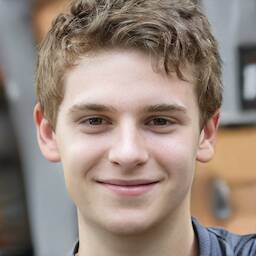}\hfill%
	\includegraphics[width=\fwid]{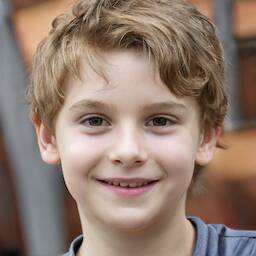}\hfill%
	\includegraphics[width=\fwid]{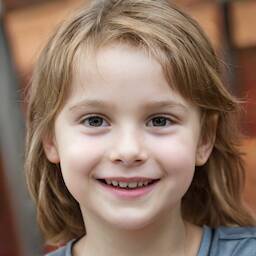}\hfill%
	\includegraphics[width=\fwid]{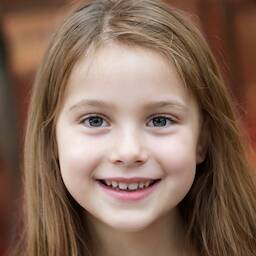}\hfill%
	\includegraphics[width=\fwid]{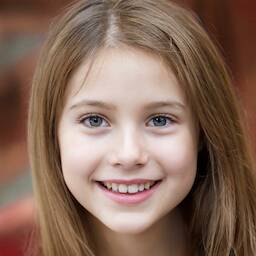}
\\
	\includegraphics[width=\fwid]{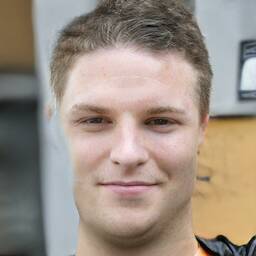}\hfill%
	\includegraphics[width=\fwid]{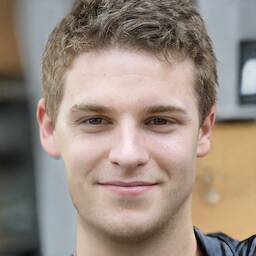}\hfill%
	\includegraphics[width=\fwid]{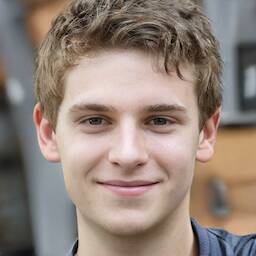}\hfill%
	\includegraphics[width=\fwid]{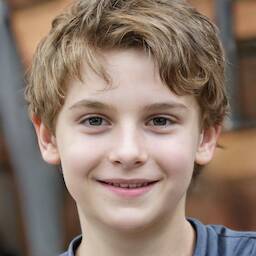}\hfill%
	\includegraphics[width=\fwid]{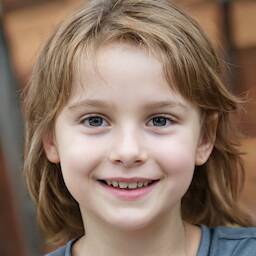}\hfill%
	\includegraphics[width=\fwid]{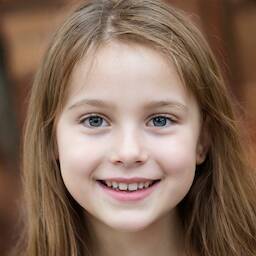}\hfill%
	\includegraphics[width=\fwid]{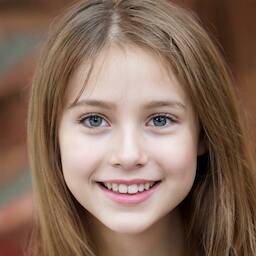}
\\
	\includegraphics[width=\fwid]{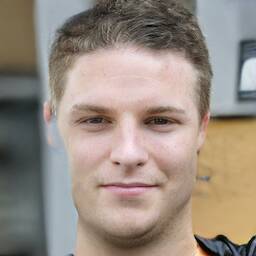}\hfill%
	\includegraphics[width=\fwid]{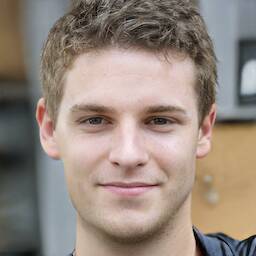}\hfill%
	\includegraphics[width=\fwid]{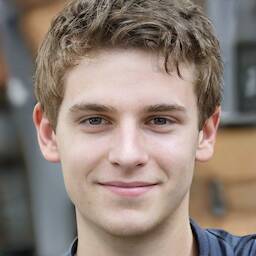}\hfill%
	\includegraphics[width=\fwid]{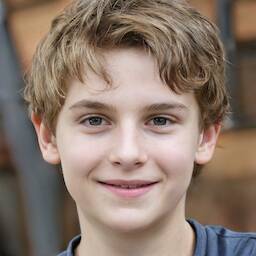}\hfill%
	\includegraphics[width=\fwid]{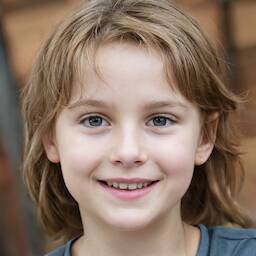}\hfill%
	\includegraphics[width=\fwid]{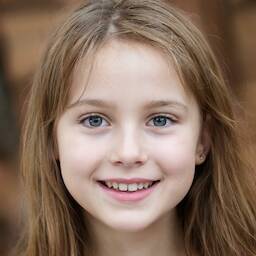}\hfill%
	\includegraphics[width=\fwid]{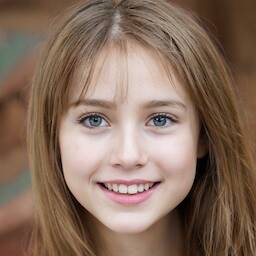}
\\
	\includegraphics[width=\fwid]{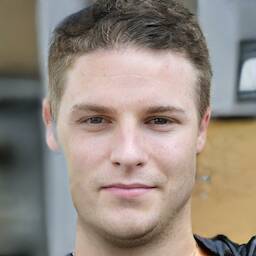}\hfill%
	\includegraphics[width=\fwid]{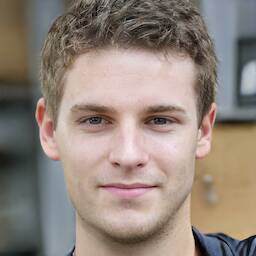}\hfill%
	\includegraphics[width=\fwid]{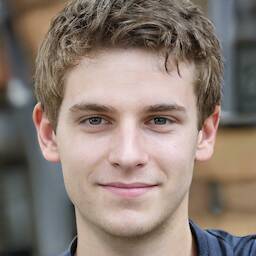}\hfill%
	\includegraphics[width=\fwid]{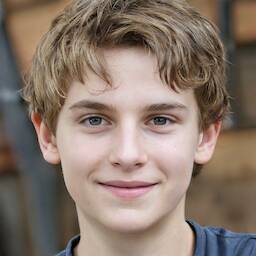}\hfill%
	\includegraphics[width=\fwid]{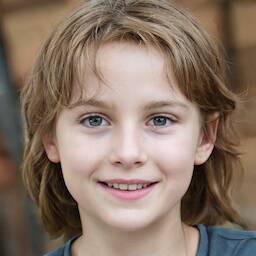}\hfill%
	\includegraphics[width=\fwid]{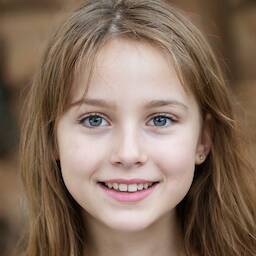}\hfill%
	\includegraphics[width=\fwid]{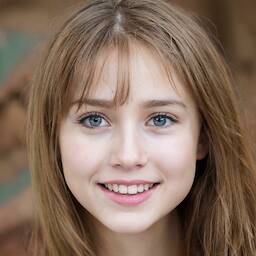}
\\
	\includegraphics[width=\fwid]{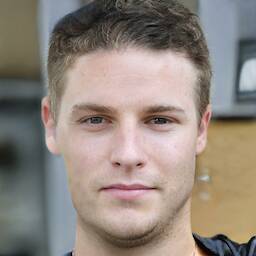}\hfill%
	\includegraphics[width=\fwid]{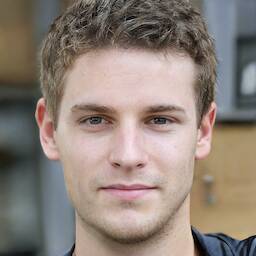}\hfill%
	\includegraphics[width=\fwid]{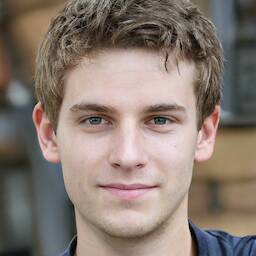}\hfill%
	\includegraphics[width=\fwid]{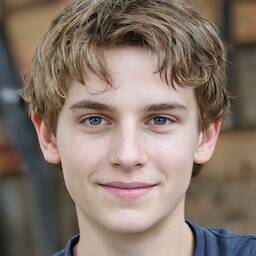}\hfill%
	\includegraphics[width=\fwid]{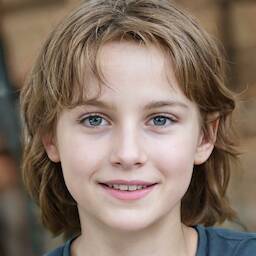}\hfill%
	\includegraphics[width=\fwid]{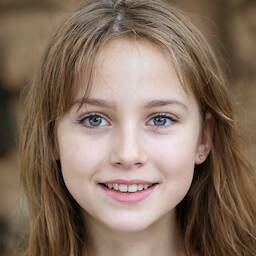}\hfill%
	\includegraphics[width=\fwid]{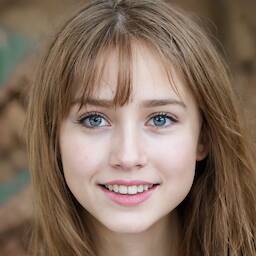}
\\
	\includegraphics[width=\fwid]{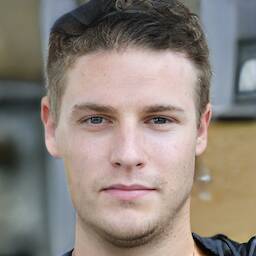}\hfill%
	\includegraphics[width=\fwid]{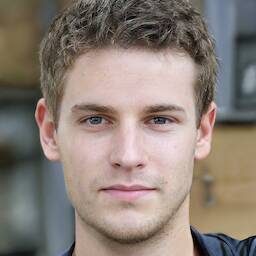}\hfill%
	\includegraphics[width=\fwid]{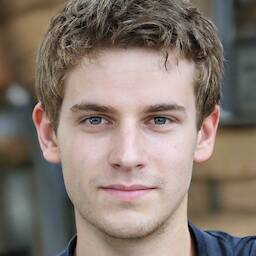}\hfill%
	\includegraphics[width=\fwid]{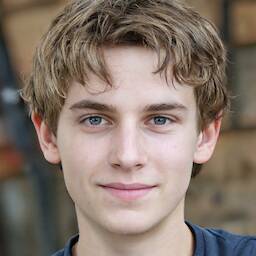}\hfill%
	\includegraphics[width=\fwid]{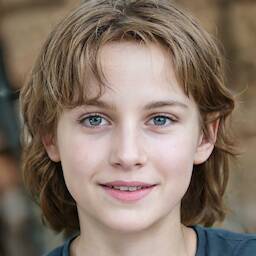}\hfill%
	\includegraphics[width=\fwid]{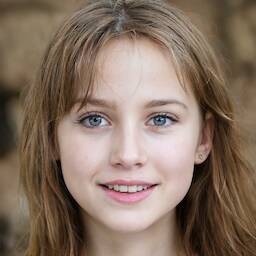}\hfill%
	\includegraphics[width=\fwid]{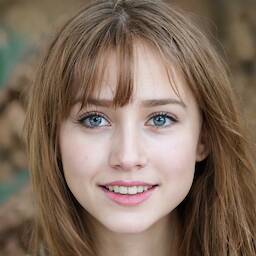}
\caption{Combining edits. Starting with the center image, the horizontal axis corresponds to adding or removing elements of $x_0$, in the range $\Delta x_0\in[-3,3]$. The horizontal axis is adding/removing elements of $x_{18}$.
Note that the horizontal axis roughly corresponds to "masculinity" and the vertical to "age." The components operate independently, except that the model does not produce a "masculine little boy" in the upper-left.}
\label{fig:combining}
\end{figure}
}

\newcommand{\figEntanglementBaldness}
{
\begin{figure}
\centering
	\includegraphics[width=\fwid]{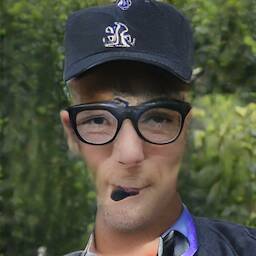}\hfill%
	\includegraphics[width=\fwid]{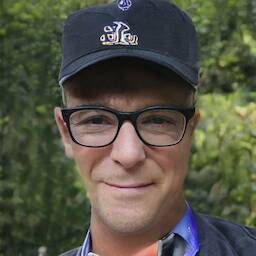}\hfill%
	\includegraphics[width=\fwid]{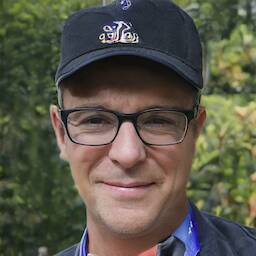}\hfill%
	\includegraphics[width=\fwid]{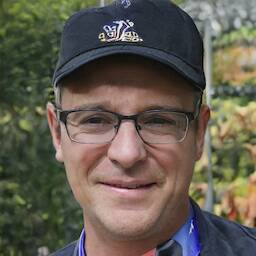}\hfill%
	\includegraphics[width=\fwid]{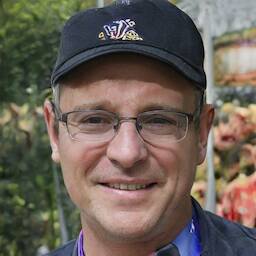}\hfill%
	\includegraphics[width=\fwid]{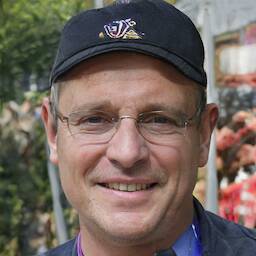}\hfill%
	\includegraphics[width=\fwid]{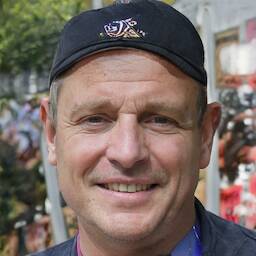} \\
	\includegraphics[width=\fwid]{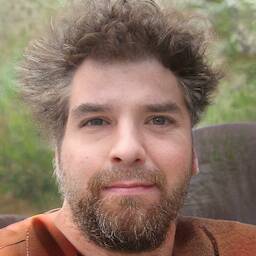}\hfill%
	\includegraphics[width=\fwid]{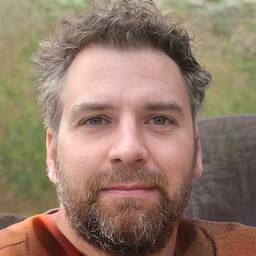}\hfill%
	\includegraphics[width=\fwid]{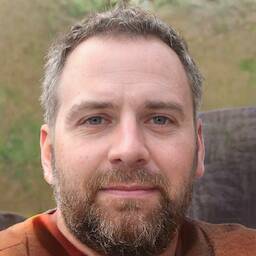}\hfill%
	\includegraphics[width=\fwid]{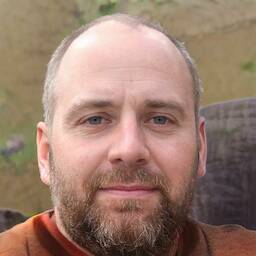}\hfill%
	\includegraphics[width=\fwid]{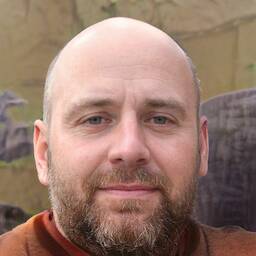}\hfill%
	\includegraphics[width=\fwid]{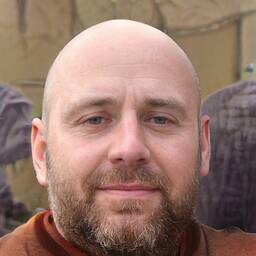}\hfill%
	\includegraphics[width=\fwid]{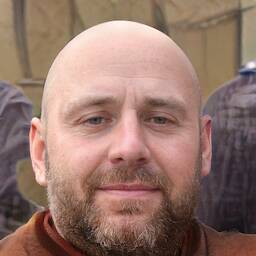}\\
	\includegraphics[width=\fwid]{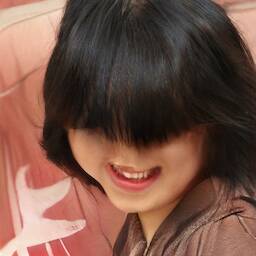}\hfill%
	\includegraphics[width=\fwid]{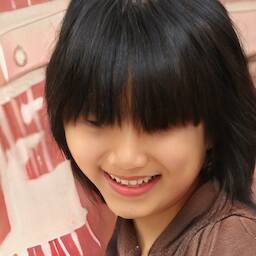}\hfill%
	\includegraphics[width=\fwid]{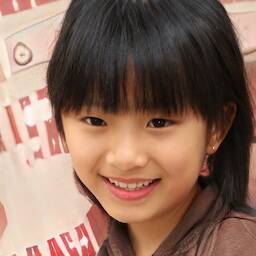}\hfill%
	\includegraphics[width=\fwid]{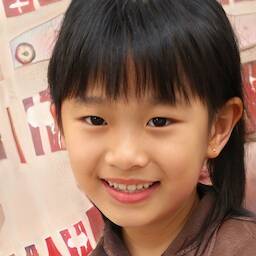}\hfill%
	\includegraphics[width=\fwid]{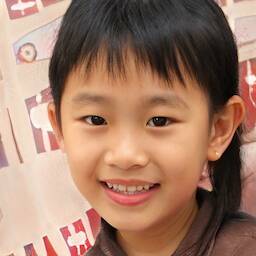}\hfill%
	\includegraphics[width=\fwid]{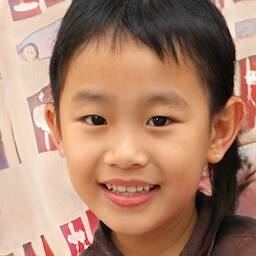}\hfill%
	\includegraphics[width=\fwid]{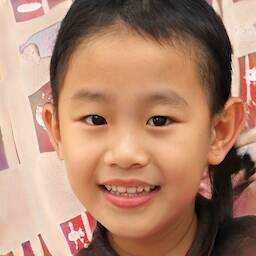}\\
	\includegraphics[width=\fwid]{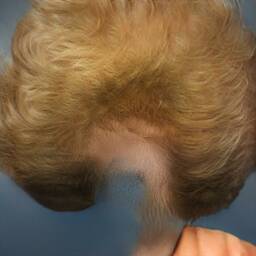}\hfill%
	\includegraphics[width=\fwid]{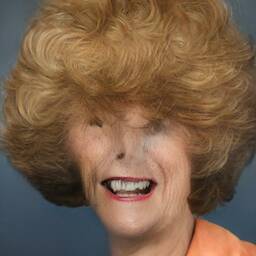}\hfill%
	\includegraphics[width=\fwid]{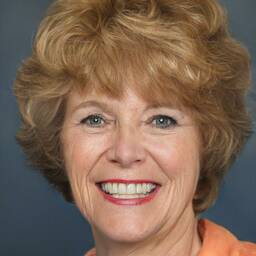}\hfill%
	\includegraphics[width=\fwid]{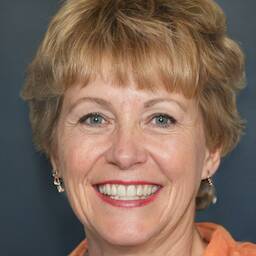}\hfill%
	\includegraphics[width=\fwid]{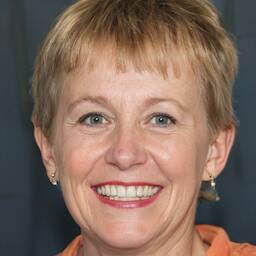}\hfill%
	\includegraphics[width=\fwid]{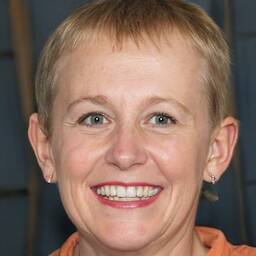}\hfill%
	\includegraphics[width=\fwid]{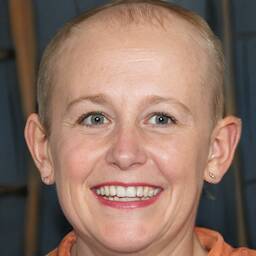}
	\caption{An example of edit direction dependence on input face: StyleGAN FFHQ direction that we labeled as ``baldness'' (\editv{21}{2-4}).}
\label{fig:Baldness}
\end{figure}
}

\newcommand{\figEntanglementMakeup}
{
\begin{figure}
\centering
	\includegraphics[width=\fwid]{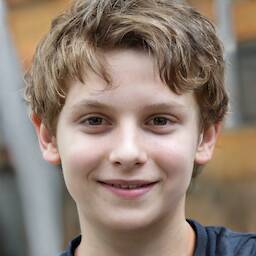}\hfill%
	\includegraphics[width=\fwid]{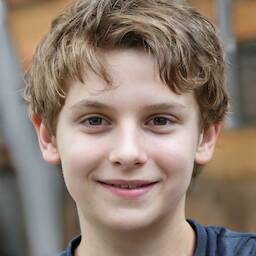}\hfill%
	\includegraphics[width=\fwid]{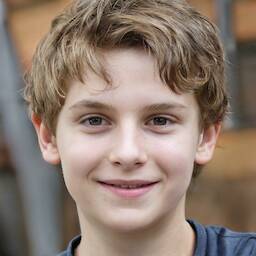}\hfill%
	\includegraphics[width=\fwid]{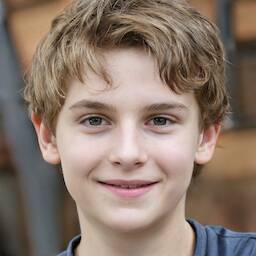}\hfill%
	\includegraphics[width=\fwid]{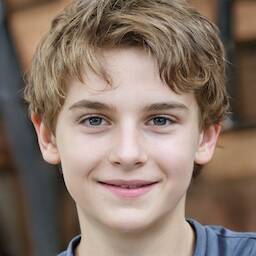}\hfill%
	\includegraphics[width=\fwid]{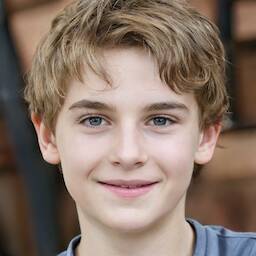}\hfill%
	\includegraphics[width=\fwid]{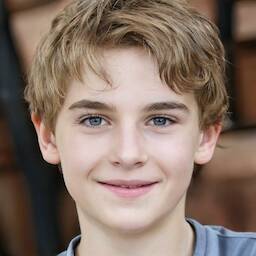} \\
	\includegraphics[width=\fwid]{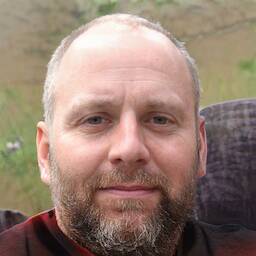}\hfill%
	\includegraphics[width=\fwid]{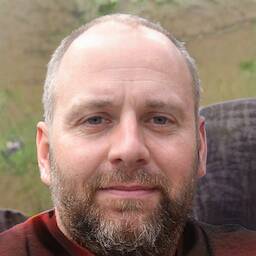}\hfill%
	\includegraphics[width=\fwid]{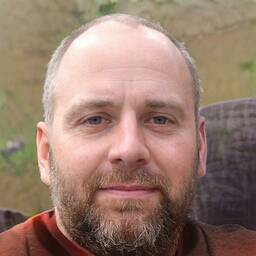}\hfill%
	\includegraphics[width=\fwid]{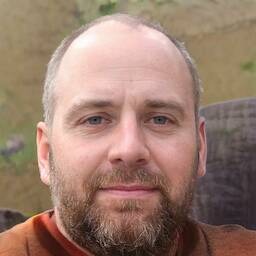}\hfill%
	\includegraphics[width=\fwid]{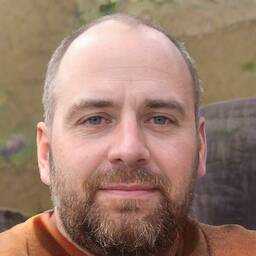}\hfill%
	\includegraphics[width=\fwid]{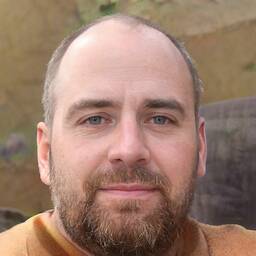}\hfill%
	\includegraphics[width=\fwid]{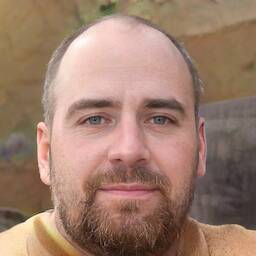}\\
	\includegraphics[width=\fwid]{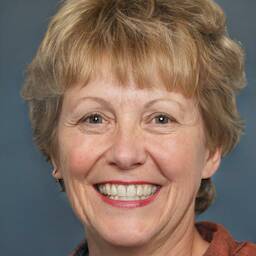}\hfill%
	\includegraphics[width=\fwid]{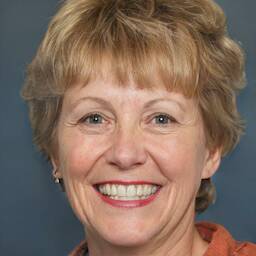}\hfill%
	\includegraphics[width=\fwid]{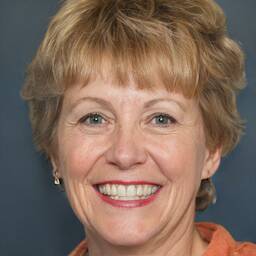}\hfill%
	\includegraphics[width=\fwid]{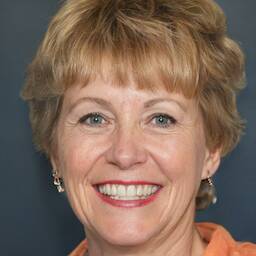}\hfill%
	\includegraphics[width=\fwid]{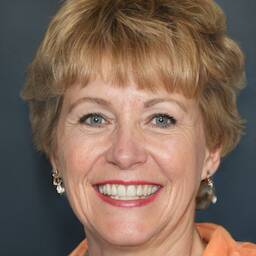}\hfill%
	\includegraphics[width=\fwid]{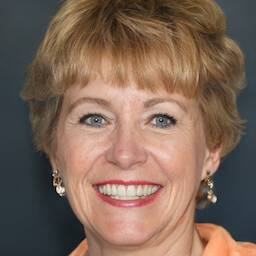}\hfill%
	\includegraphics[width=\fwid]{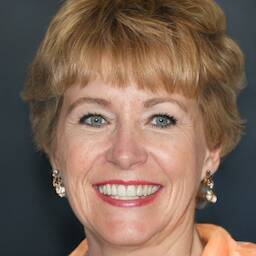}\\
	\includegraphics[width=\fwid]{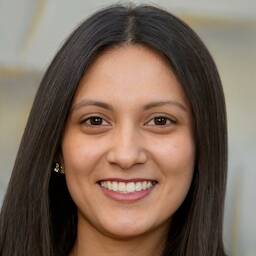}\hfill%
	\includegraphics[width=\fwid]{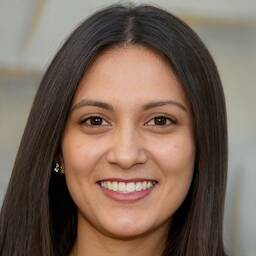}\hfill%
	\includegraphics[width=\fwid]{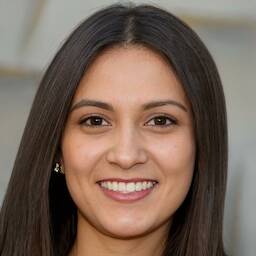}\hfill%
	\includegraphics[width=\fwid]{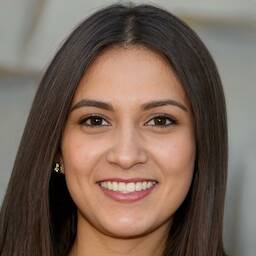}\hfill%
	\includegraphics[width=\fwid]{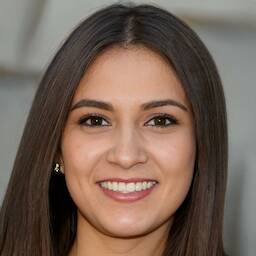}\hfill%
	\includegraphics[width=\fwid]{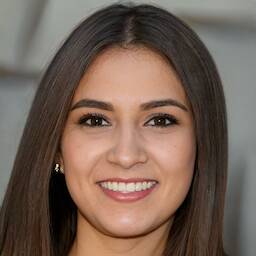}\hfill%
	\includegraphics[width=\fwid]{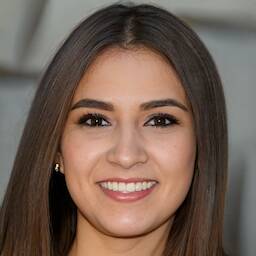}
	\caption{An example of edit direction dependence on input face: StyleGAN FFHQ direction that we labeled as ``makeup'' (\editv{0}{8}).
	}
\label{fig:Makeup}
\end{figure}
}

\newcommand{\figEntanglementWhiteHair}
{
\begin{figure}
\centering
	\includegraphics[width=\fwid]{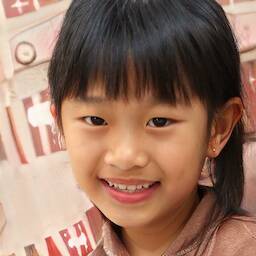}\hfill%
	\includegraphics[width=\fwid]{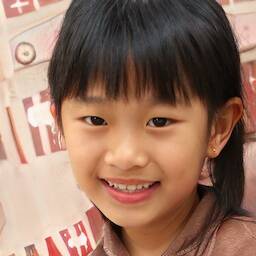}\hfill%
	\includegraphics[width=\fwid]{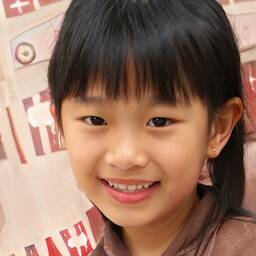}\hfill%
	\includegraphics[width=\fwid]{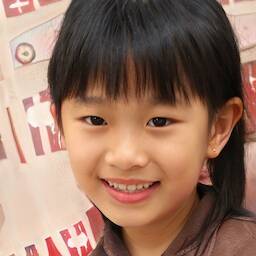}\hfill%
	\includegraphics[width=\fwid]{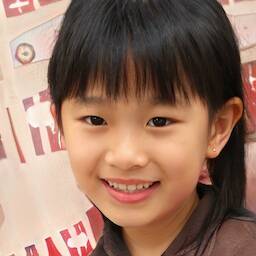}\hfill%
	\includegraphics[width=\fwid]{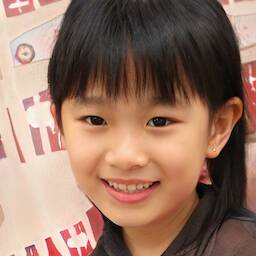}\hfill%
	\includegraphics[width=\fwid]{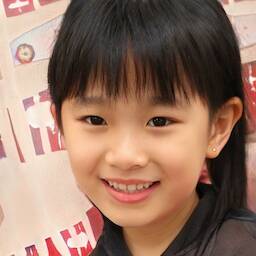} \\
		\includegraphics[width=\fwid]{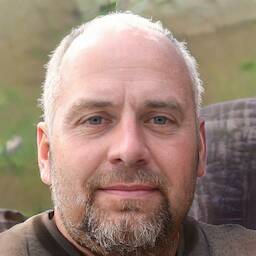}\hfill%
	\includegraphics[width=\fwid]{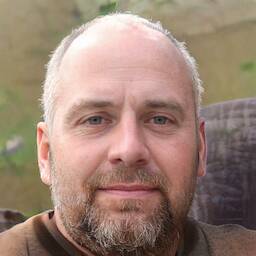}\hfill%
	\includegraphics[width=\fwid]{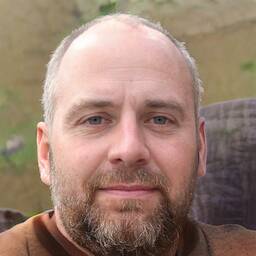}\hfill%
	\includegraphics[width=\fwid]{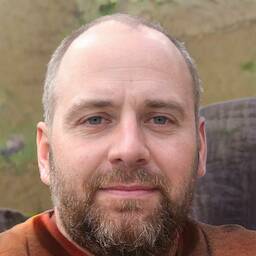}\hfill%
	\includegraphics[width=\fwid]{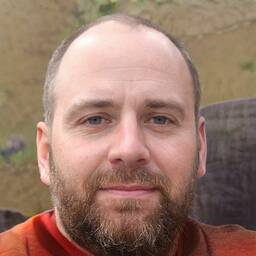}\hfill%
	\includegraphics[width=\fwid]{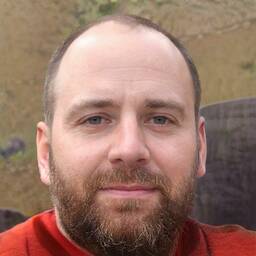}\hfill%
	\includegraphics[width=\fwid]{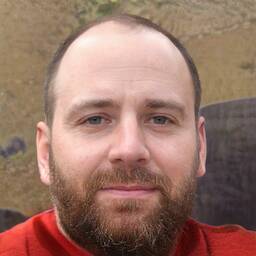} \\
	\includegraphics[width=\fwid]{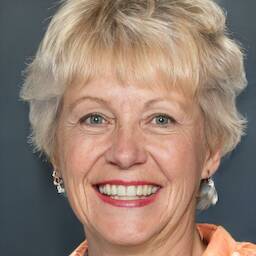}\hfill%
	\includegraphics[width=\fwid]{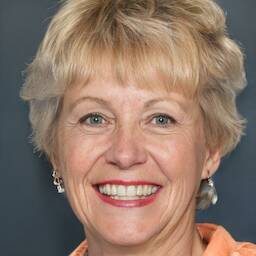}\hfill%
	\includegraphics[width=\fwid]{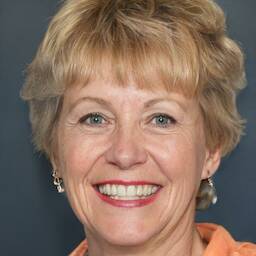}\hfill%
	\includegraphics[width=\fwid]{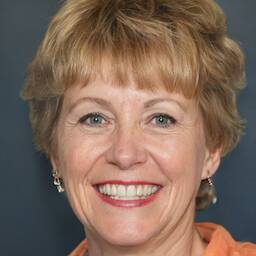}\hfill%
	\includegraphics[width=\fwid]{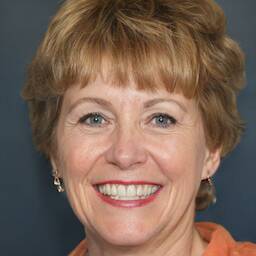}\hfill%
	\includegraphics[width=\fwid]{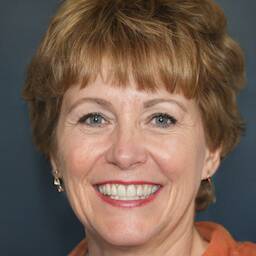}\hfill%
	\includegraphics[width=\fwid]{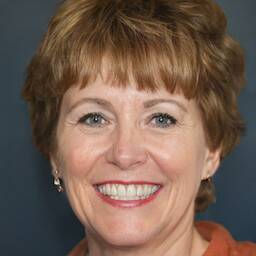} \\
	\includegraphics[width=\fwid]{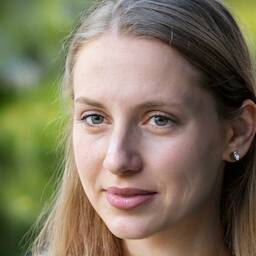}\hfill%
	\includegraphics[width=\fwid]{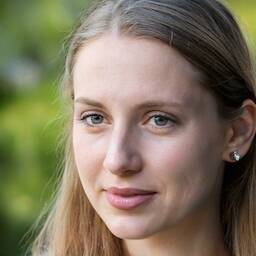}\hfill%
	\includegraphics[width=\fwid]{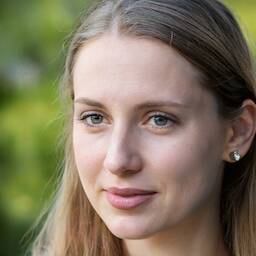}\hfill%
	\includegraphics[width=\fwid]{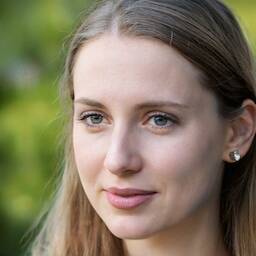}\hfill%
	\includegraphics[width=\fwid]{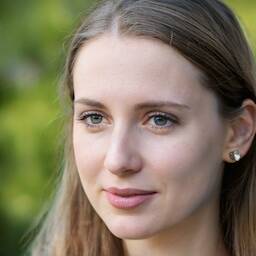}\hfill%
	\includegraphics[width=\fwid]{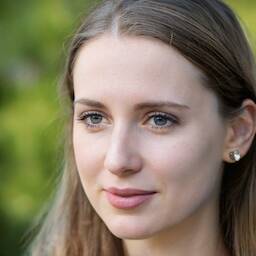}\hfill%
	\includegraphics[width=\fwid]{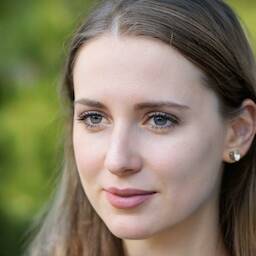}
	\caption{\label{fig:white_hair} An example of edit direction dependence on input face: StyleGAN FFHQ direction that we labeled as ``white hair'' (\editv{57}{7-9}).}
\end{figure}
}

\newcommand{\figWrinkles}{
\begin{figure}
\centering
	\includegraphics[width=\fwid]{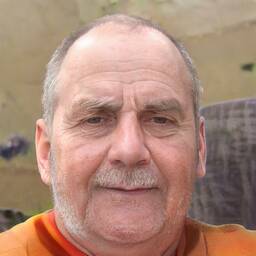}\hfill%
	\includegraphics[width=\fwid]{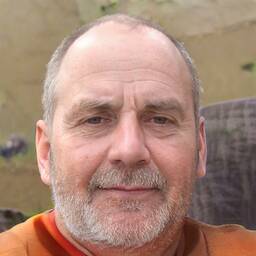}\hfill%
	\includegraphics[width=\fwid]{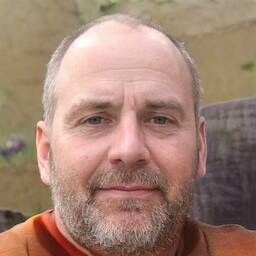}\hfill%
	\includegraphics[width=\fwid]{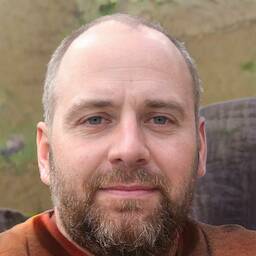}\hfill%
	\includegraphics[width=\fwid]{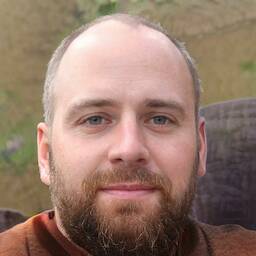}\hfill%
	\includegraphics[width=\fwid]{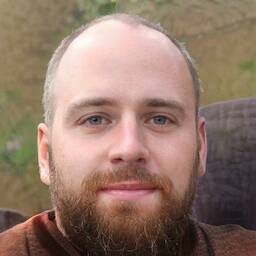}\hfill%
	\includegraphics[width=\fwid]{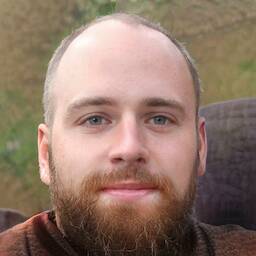} \\
	\includegraphics[width=\fwid]{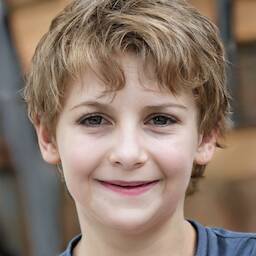}\hfill%
	\includegraphics[width=\fwid]{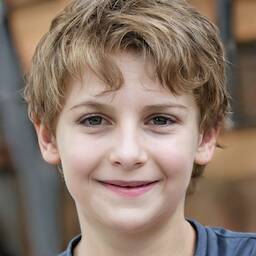}\hfill%
	\includegraphics[width=\fwid]{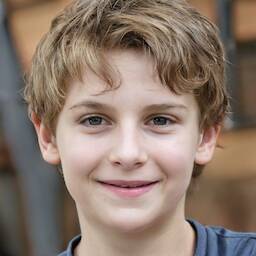}\hfill%
	\includegraphics[width=\fwid]{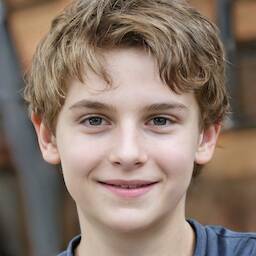}\hfill%
	\includegraphics[width=\fwid]{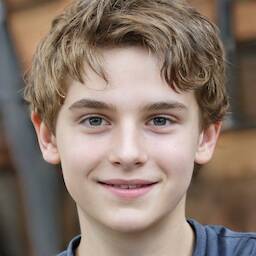}\hfill%
	\includegraphics[width=\fwid]{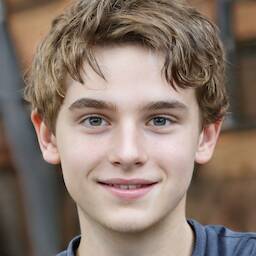}\hfill%
	\includegraphics[width=\fwid]{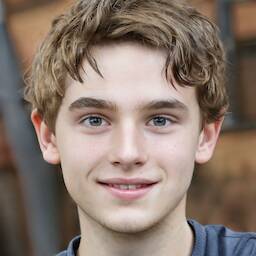} \\
	\includegraphics[width=\fwid]{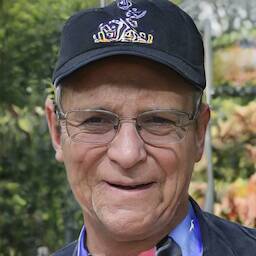}\hfill%
	\includegraphics[width=\fwid]{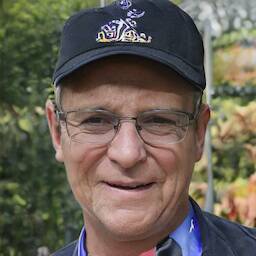}\hfill%
	\includegraphics[width=\fwid]{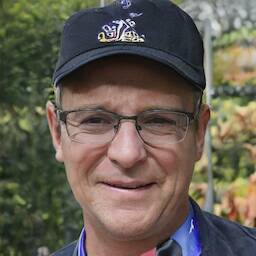}\hfill%
	\includegraphics[width=\fwid]{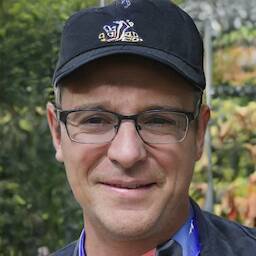}\hfill%
	\includegraphics[width=\fwid]{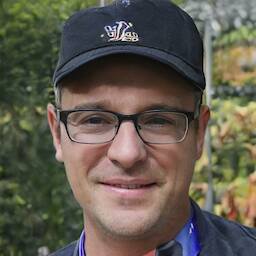}\hfill%
	\includegraphics[width=\fwid]{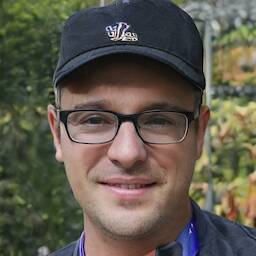}\hfill%
	\includegraphics[width=\fwid]{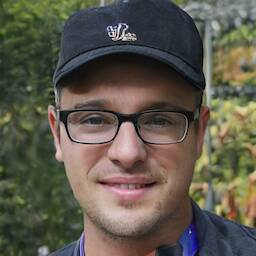} \\
	\includegraphics[width=\fwid]{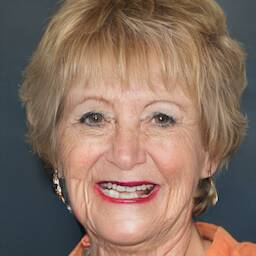}\hfill%
	\includegraphics[width=\fwid]{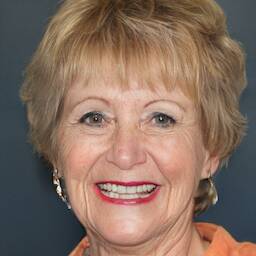}\hfill%
	\includegraphics[width=\fwid]{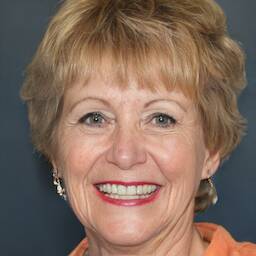}\hfill%
	\includegraphics[width=\fwid]{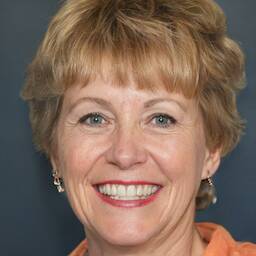}\hfill%
	\includegraphics[width=\fwid]{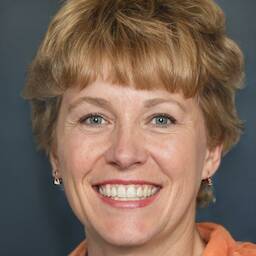}\hfill%
	\includegraphics[width=\fwid]{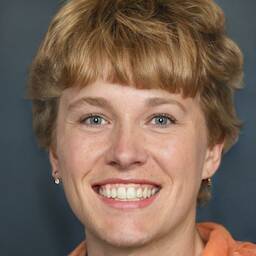}\hfill%
	\includegraphics[width=\fwid]{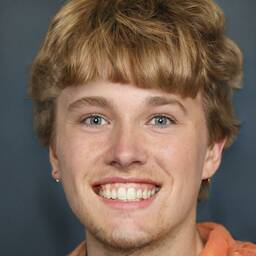}
\caption{An example of edit direction dependence on input face: StyleGAN FFHQ direction that we labeled as ``wrinkles'' (\editv{20}{6}).}
\label{fig:Wrinkles}
\end{figure}
}

\newcommand{\figTopPCsSGtwoFFHQ}{
\renewcommand{\h}{0.38\linewidth}
\begin{figure*}[t]
\centering
\includegraphics[width=\h]{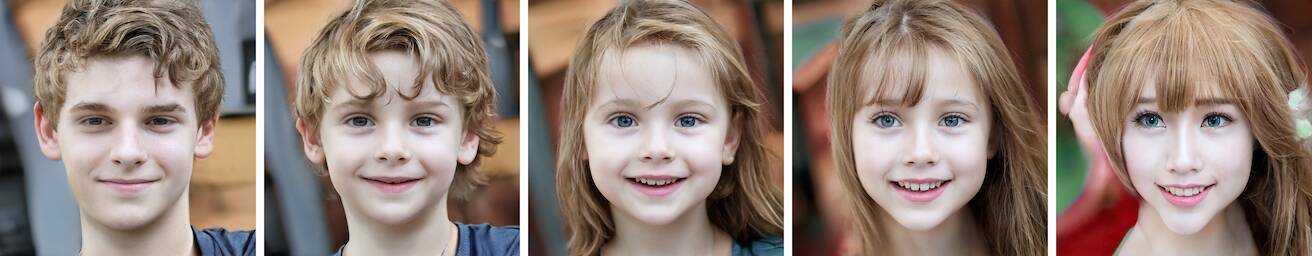}\hspace*{1cm}
\includegraphics[width=\h]{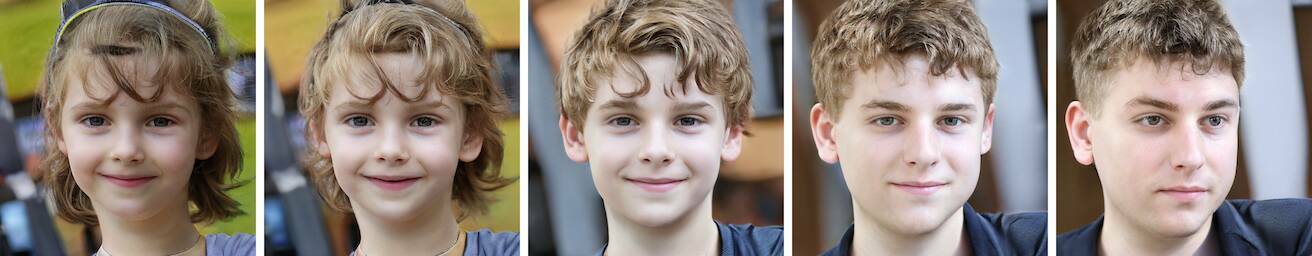}\\
\includegraphics[width=\h]{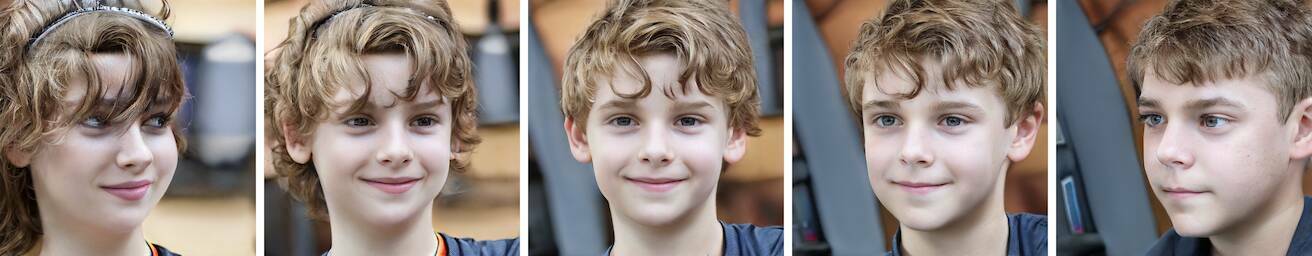}\hspace*{1cm}
\includegraphics[width=\h]{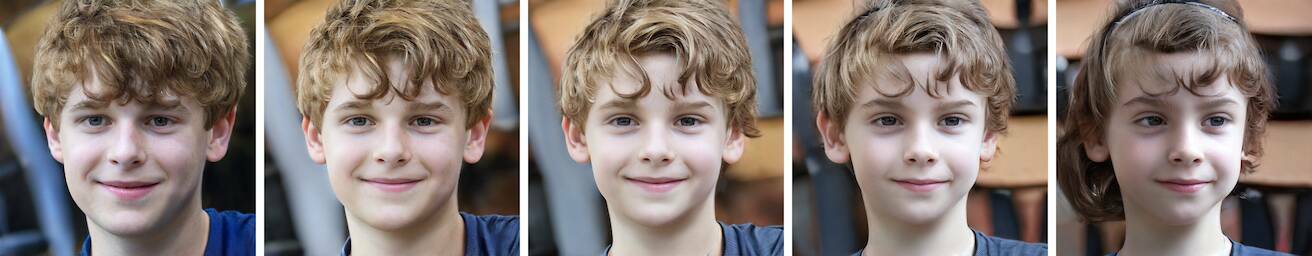}\\
\includegraphics[width=\h]{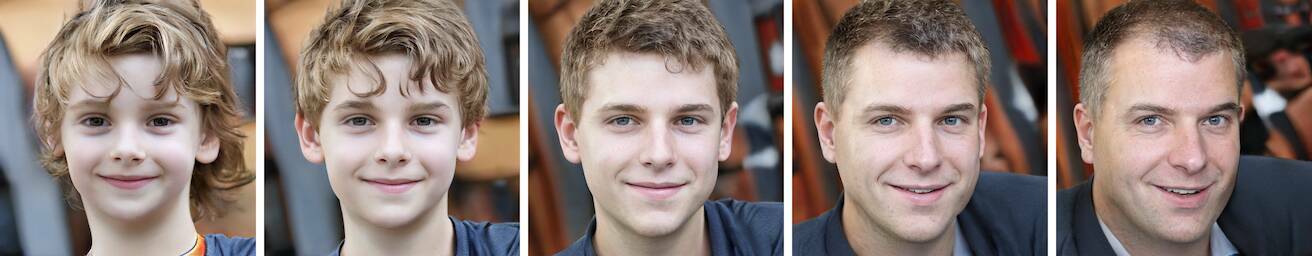}\hspace*{1cm}
\includegraphics[width=\h]{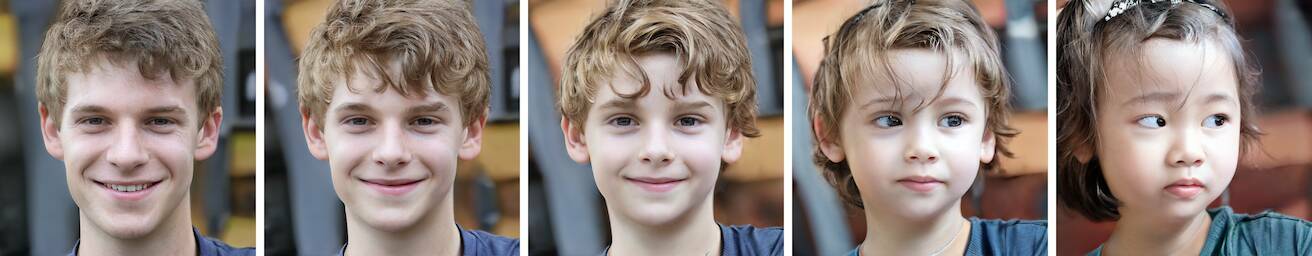}\\
\includegraphics[width=\h]{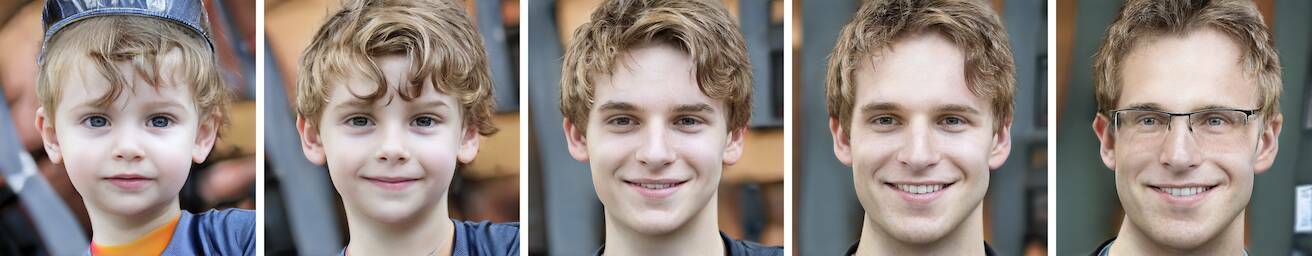}\hspace*{1cm}
\includegraphics[width=\h]{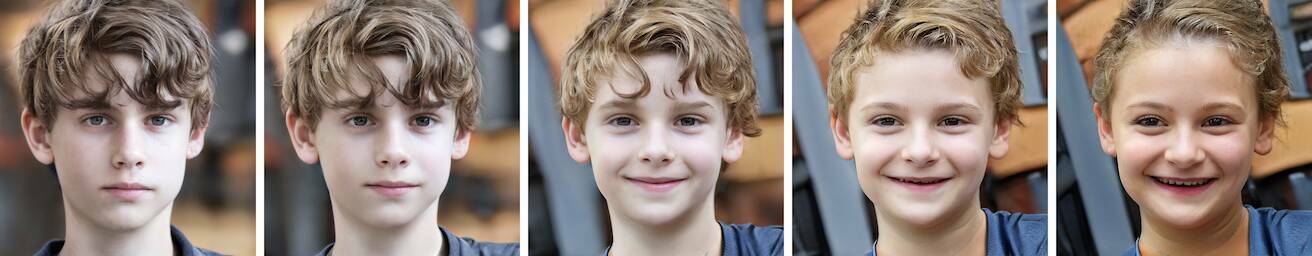}\\
\includegraphics[width=\h]{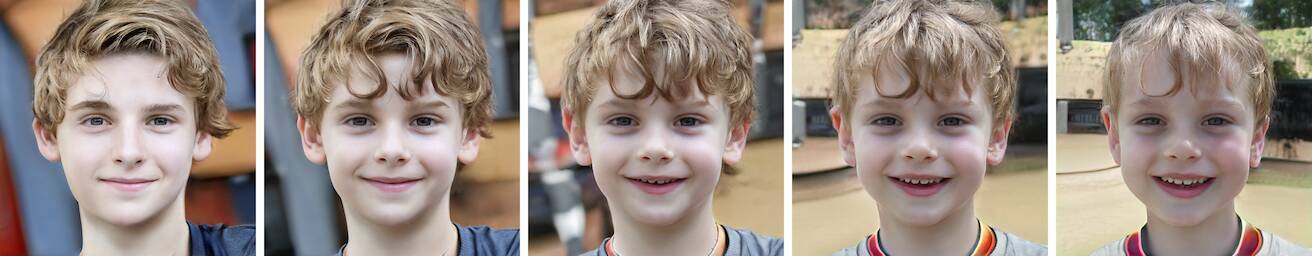}\hspace*{1cm}
\includegraphics[width=\h]{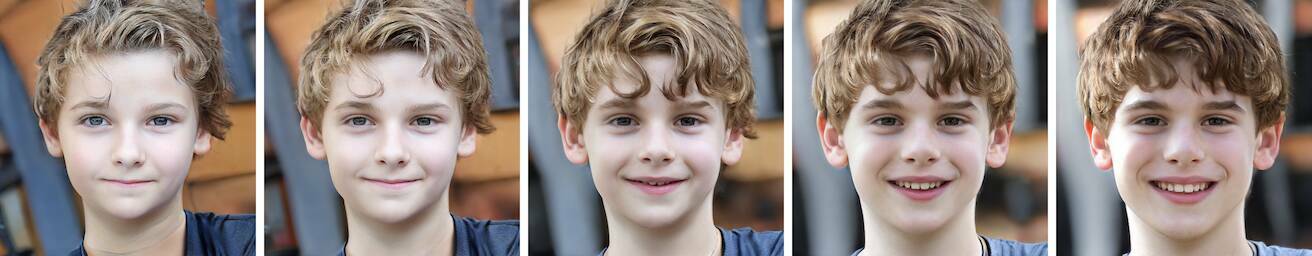}\\
\includegraphics[width=\h]{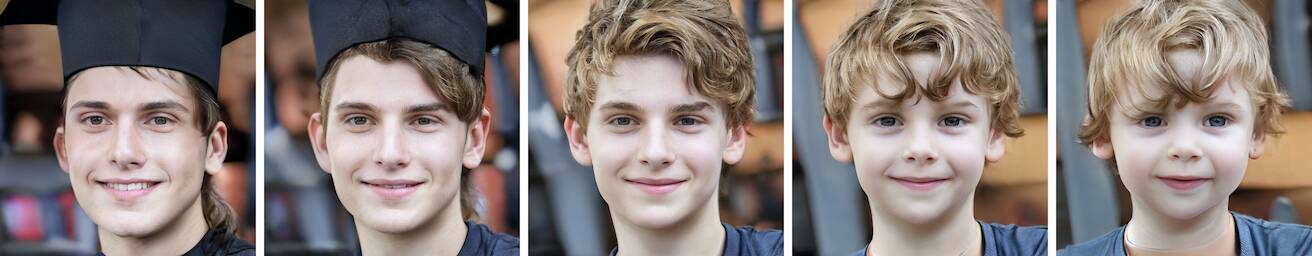}\hspace*{1cm}
\includegraphics[width=\h]{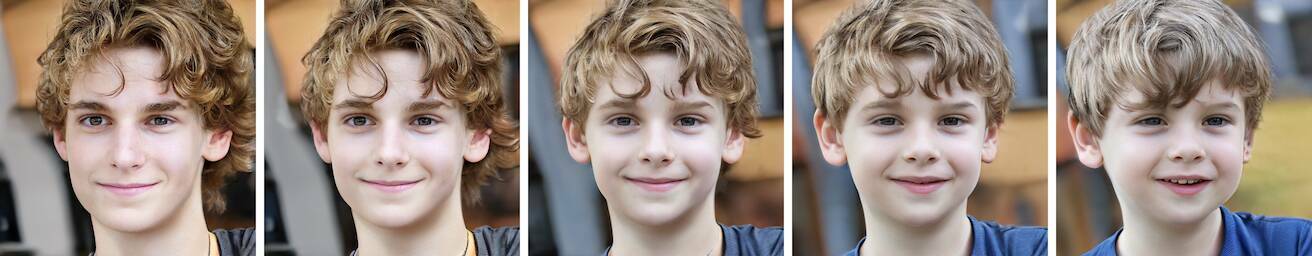}\\
\includegraphics[width=\h]{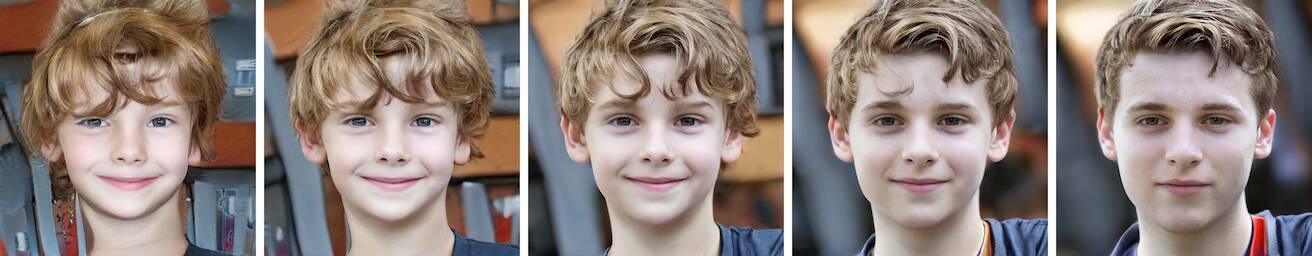}\hspace*{1cm}
\includegraphics[width=\h]{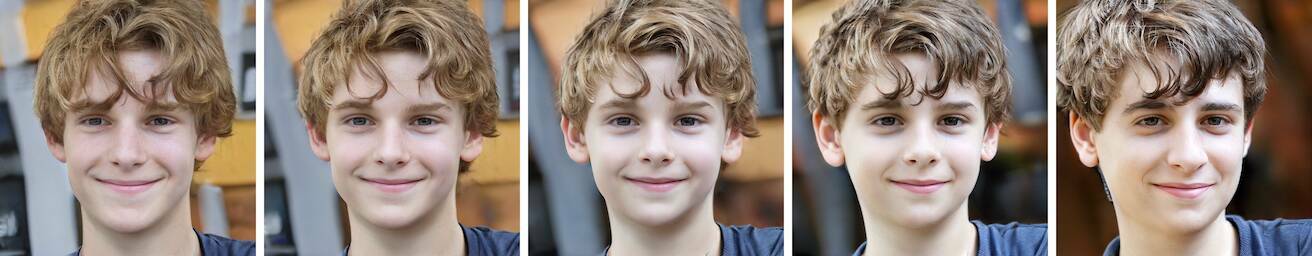}\\
\includegraphics[width=\h]{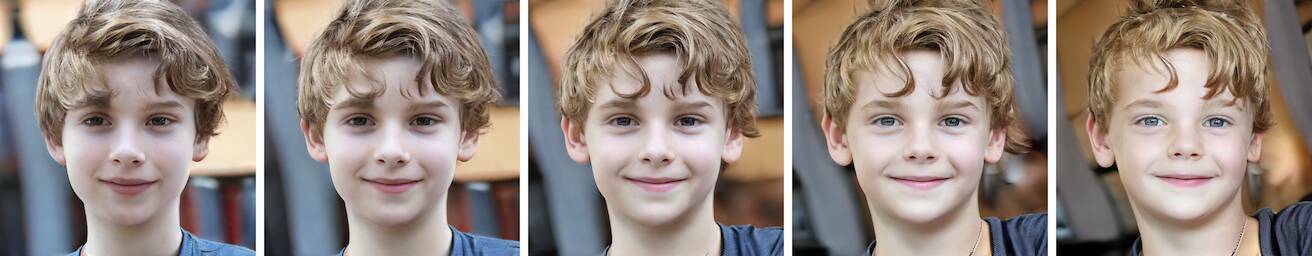}\hspace*{1cm}
\includegraphics[width=\h]{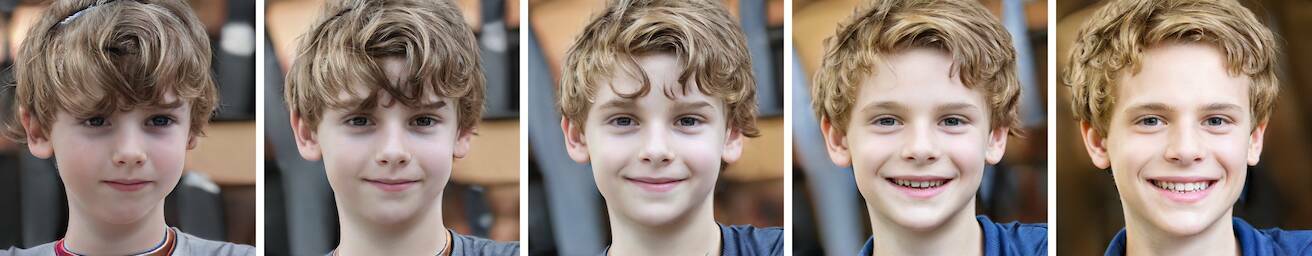}\\
\includegraphics[width=\h]{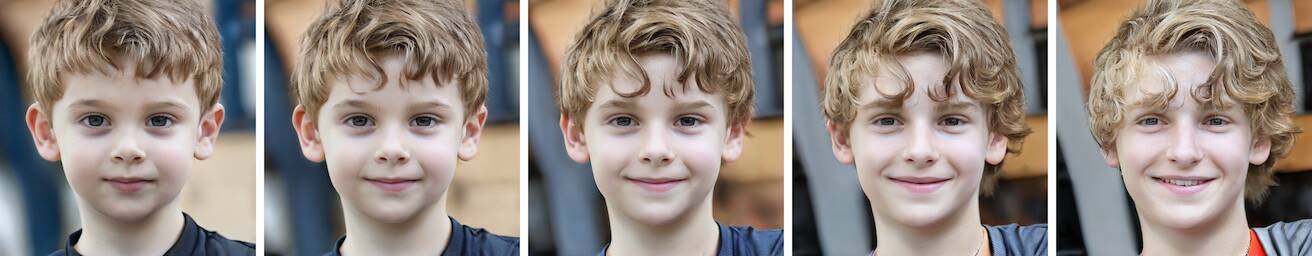}\hspace*{1cm}
\includegraphics[width=\h]{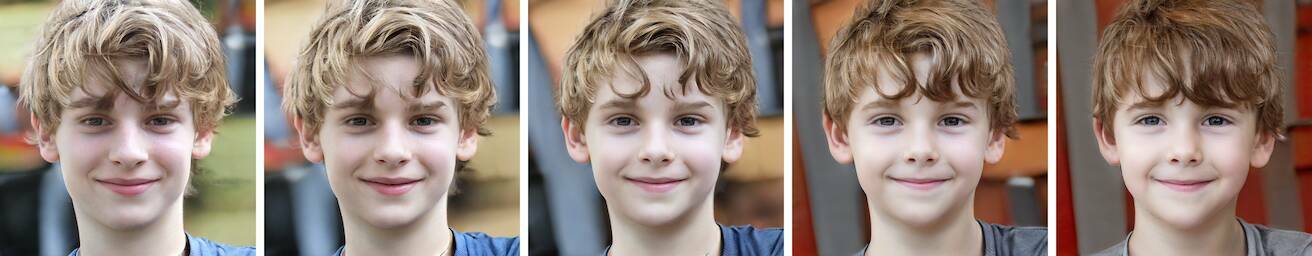}\\
\includegraphics[width=\h]{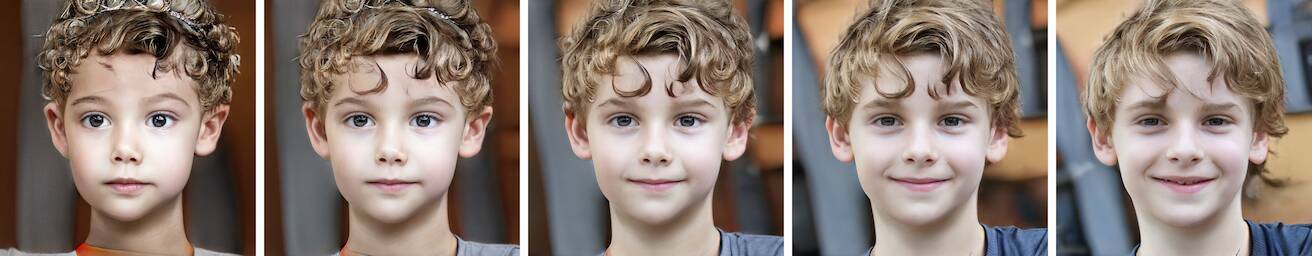}\hspace*{1cm}
\includegraphics[width=\h]{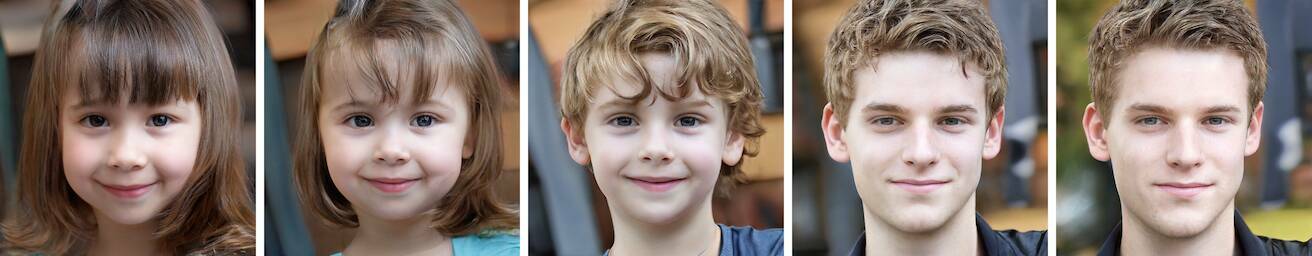}\\
\includegraphics[width=\h]{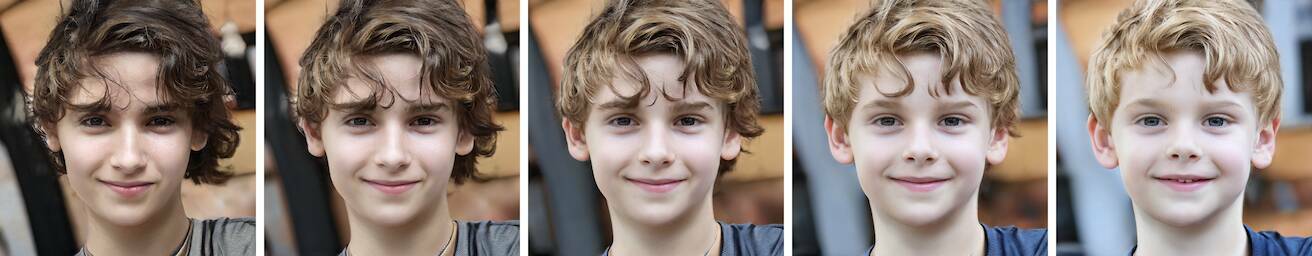}\hspace*{1cm}
\includegraphics[width=\h]{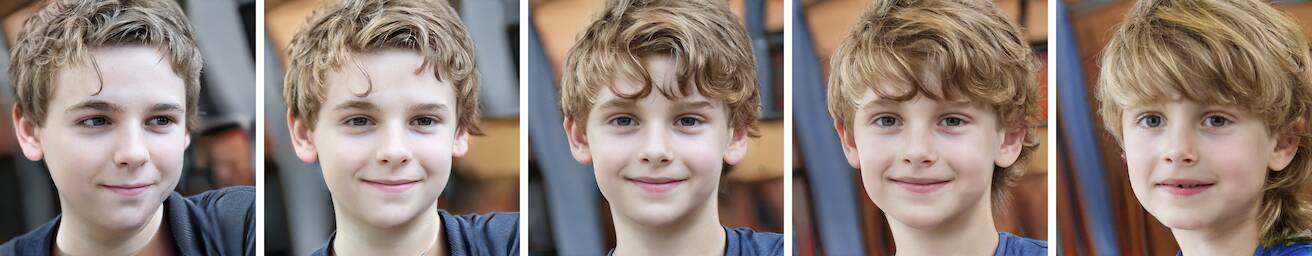}\\
\includegraphics[width=\h]{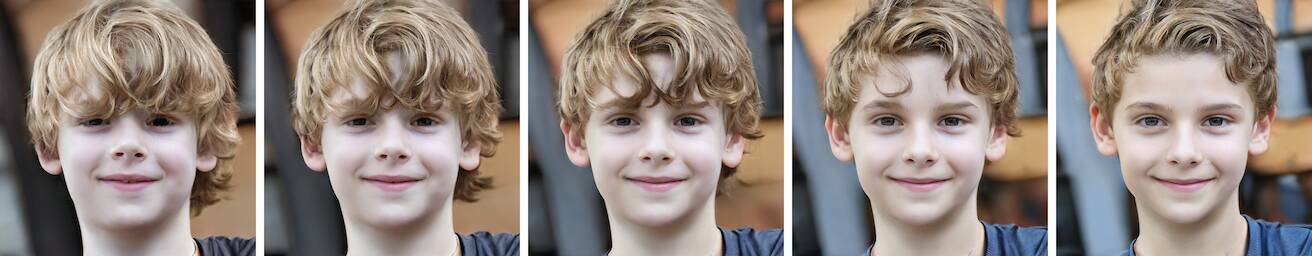}\hspace*{1cm}
\includegraphics[width=\h]{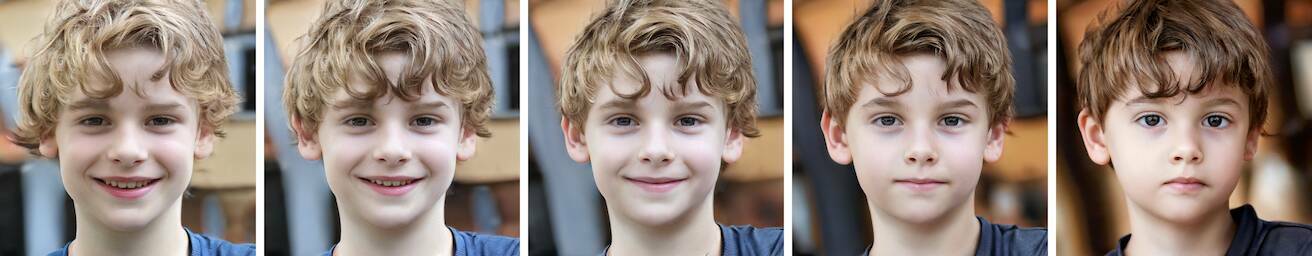}\\
\includegraphics[width=\h]{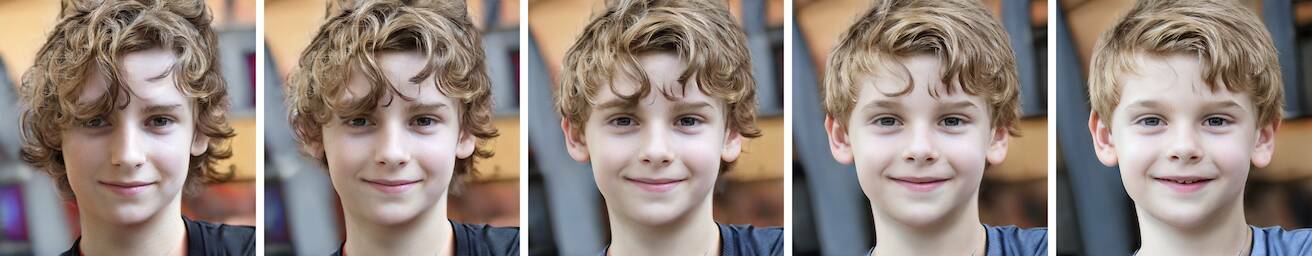}\hspace*{1cm}
\includegraphics[width=\h]{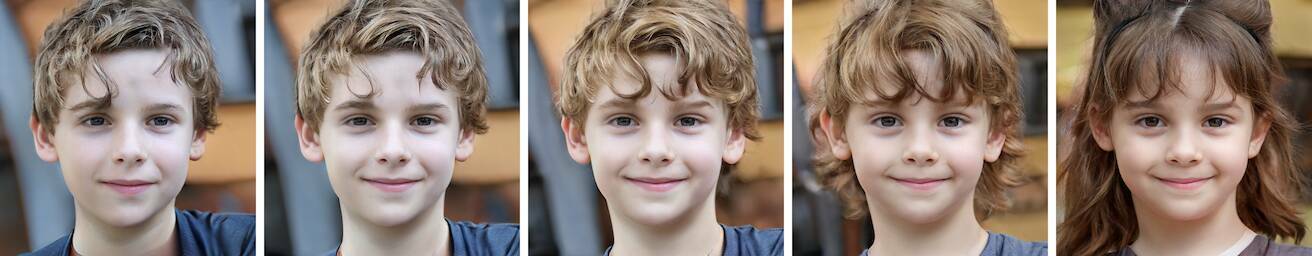}\\
\includegraphics[width=\h]{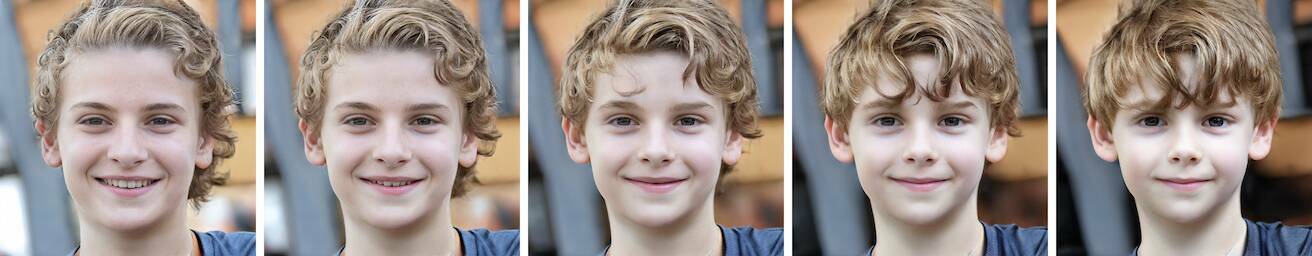}\hspace*{1cm}
\includegraphics[width=\h]{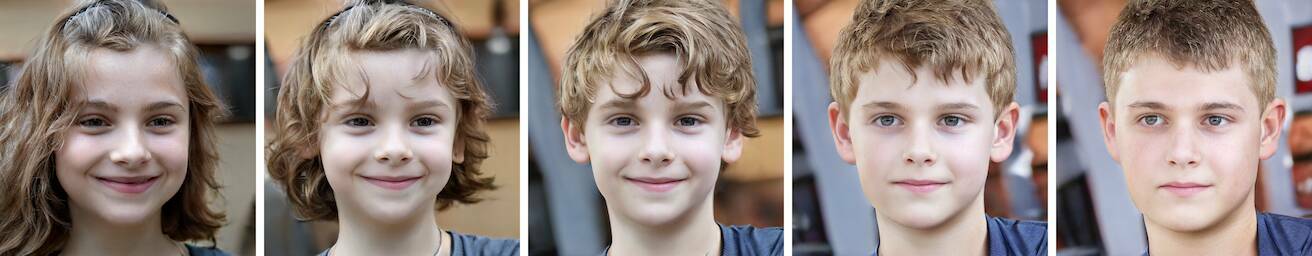}\\
\includegraphics[width=\h]{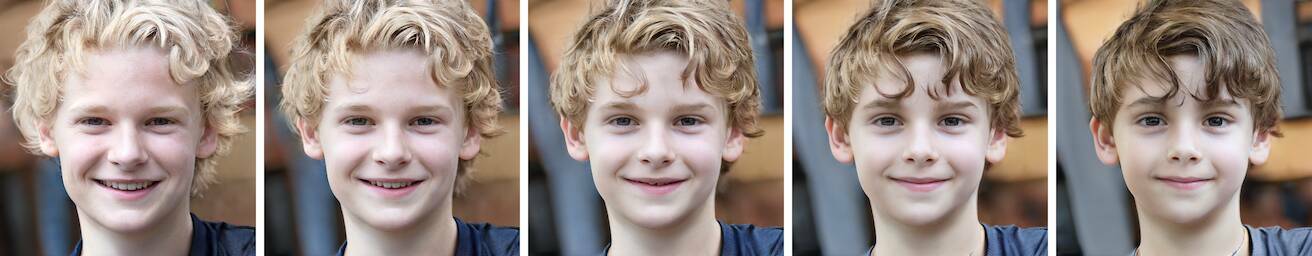}\hspace*{1cm}
\includegraphics[width=\h]{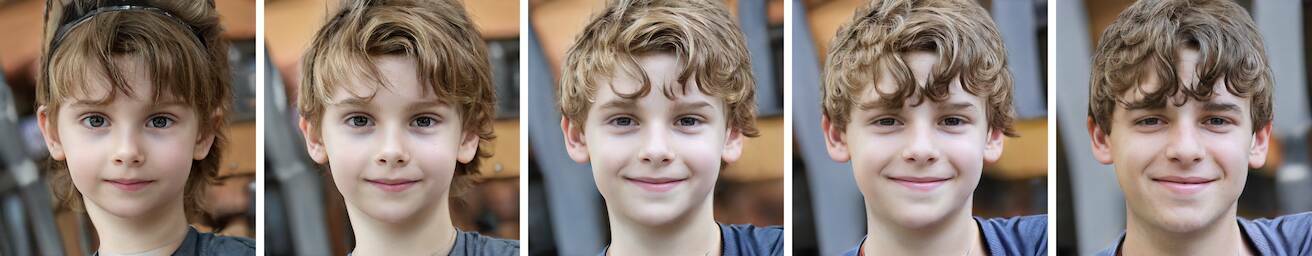}\\
\includegraphics[width=\h]{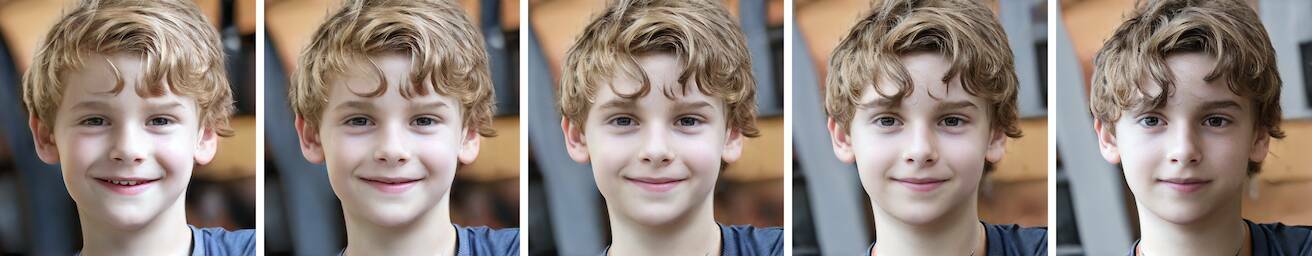}\hspace*{1cm}
\includegraphics[width=\h]{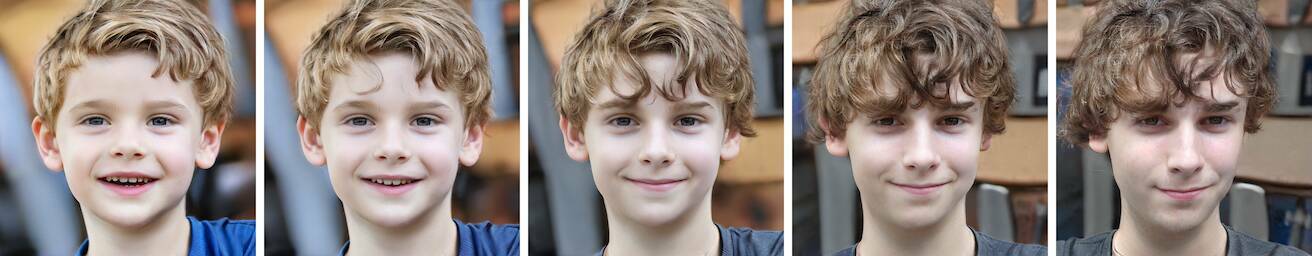}\\
\includegraphics[width=\h]{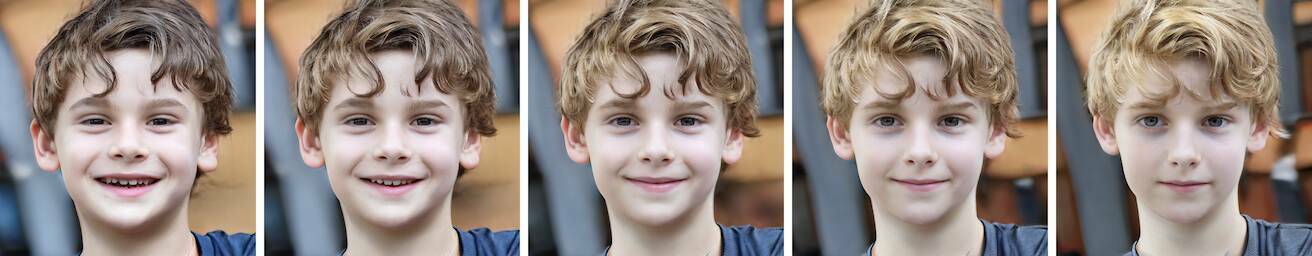}\hspace*{1cm}
\includegraphics[width=\h]{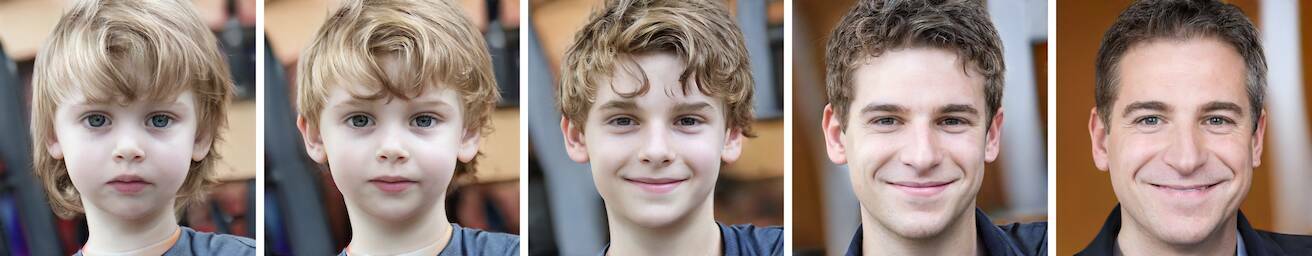}\\
\includegraphics[width=\h]{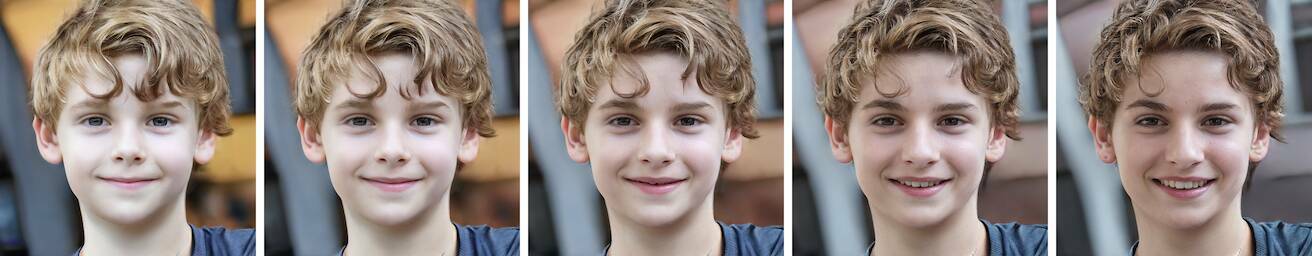}\hspace*{1cm}
\includegraphics[width=\h]{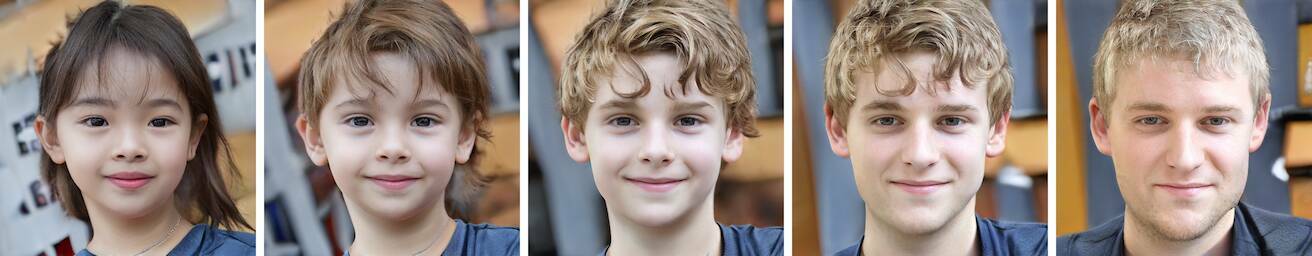}\\
\includegraphics[width=\h]{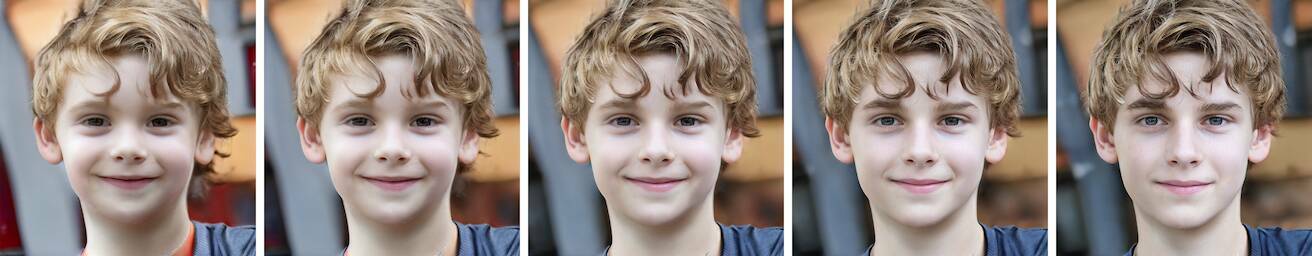}\hspace*{1cm}
\includegraphics[width=\h]{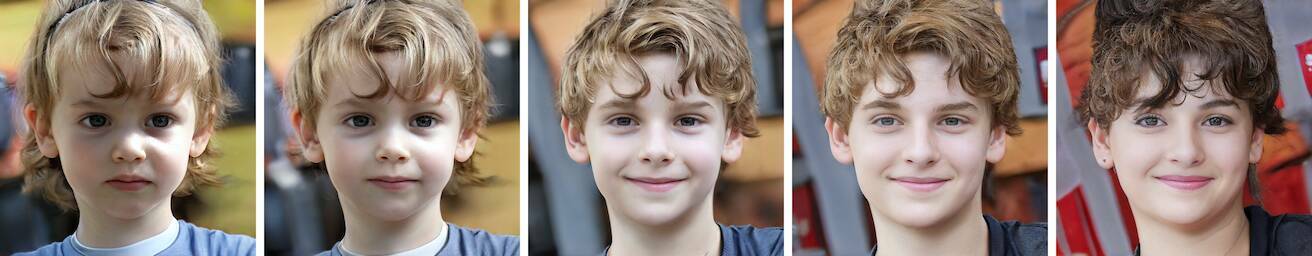}\\
\includegraphics[width=\h]{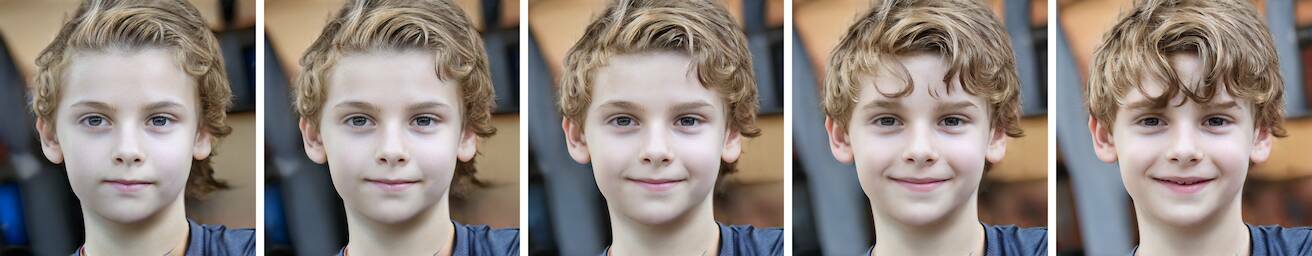}\hspace*{1cm}
\includegraphics[width=\h]{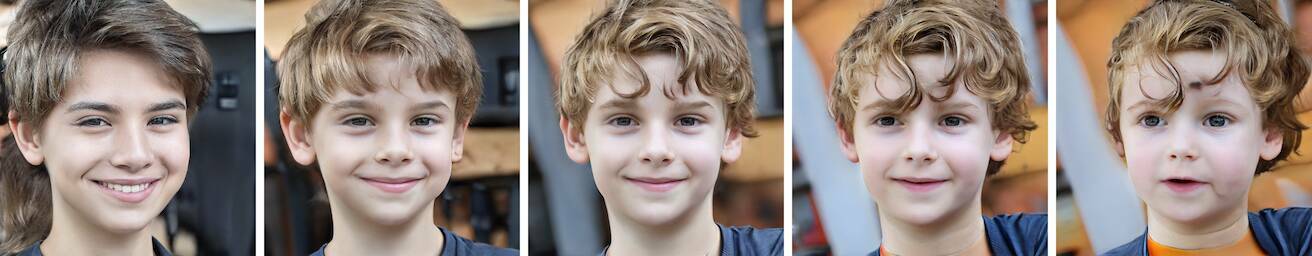}\\
\makebox[0.4\linewidth]{(a) Principal components $\mathbf{v_{0}}-\mathbf{v_{19}}$, $\pm 2\sigma$}\hspace*{1cm}
\makebox[0.4\linewidth]{(b) Normally distributed directions in $\mathcal{Z}$, $\pm 10\hat{r_i}$}%

\caption{\label{fig:topPCsFFHQ} A visualization of the first 20 principal components of StyleGAN2 FFHQ (a), and of 20 isotropic Gaussian directions in $\mathcal{Z}$ (b). The random directions are scaled to emphasize their effect.}
\end{figure*}
}

\newcommand{\figTopPCsSGtwoCats}{
\renewcommand{\h}{0.38\linewidth}
\begin{figure*}[t]
\centering
\includegraphics[width=\h]{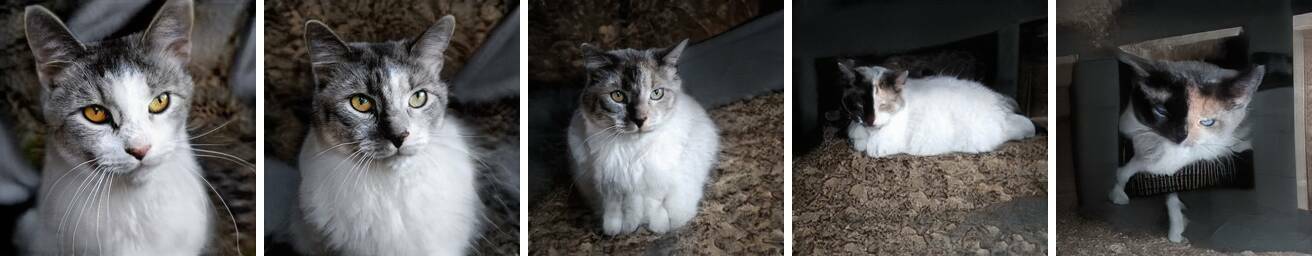}\hspace*{1cm}
\includegraphics[width=\h]{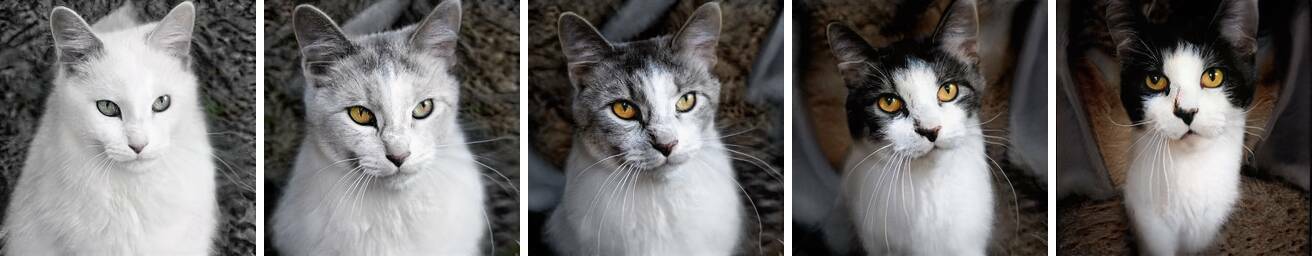}\\
\includegraphics[width=\h]{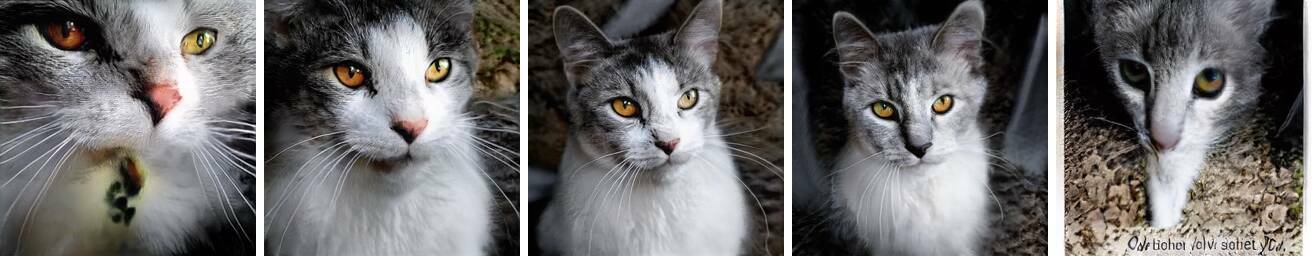}\hspace*{1cm}
\includegraphics[width=\h]{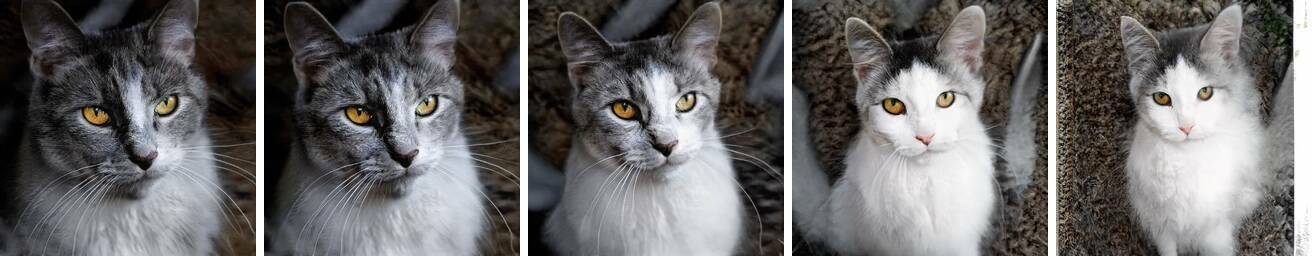}\\
\includegraphics[width=\h]{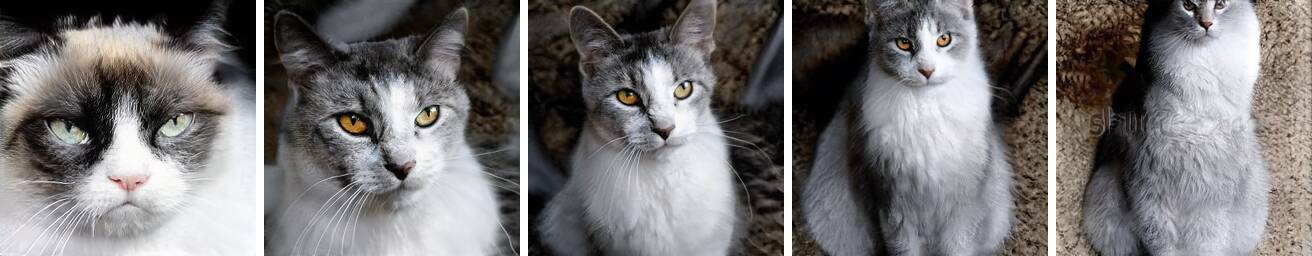}\hspace*{1cm}
\includegraphics[width=\h]{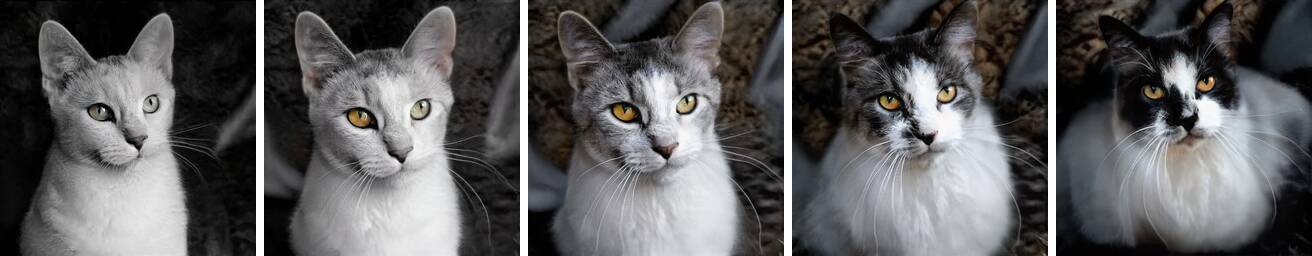}\\
\includegraphics[width=\h]{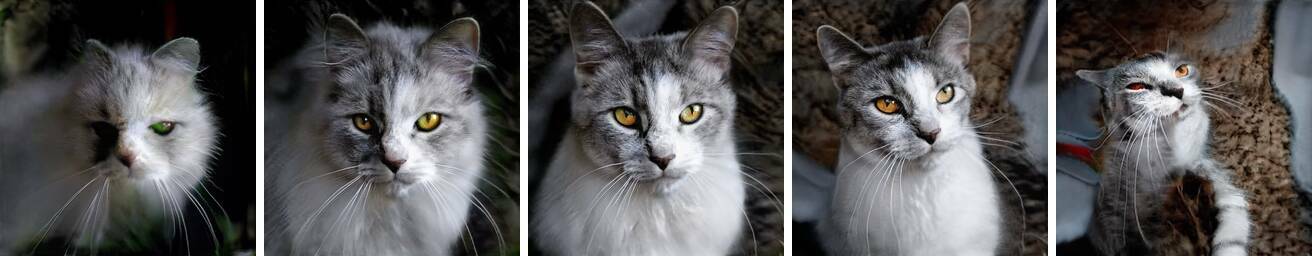}\hspace*{1cm}
\includegraphics[width=\h]{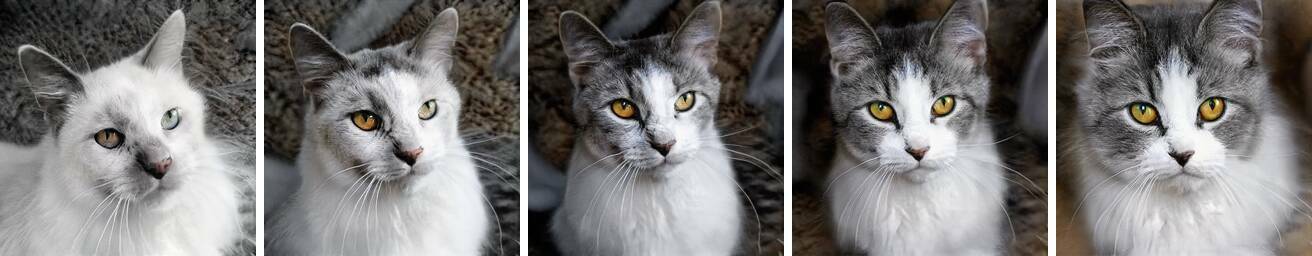}\\
\includegraphics[width=\h]{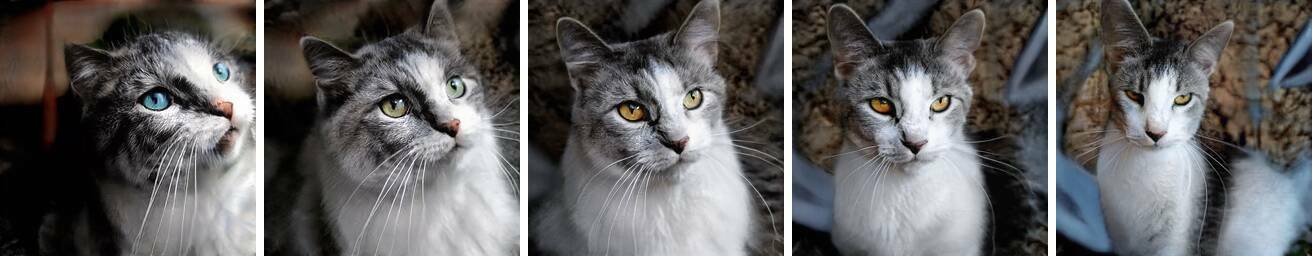}\hspace*{1cm}
\includegraphics[width=\h]{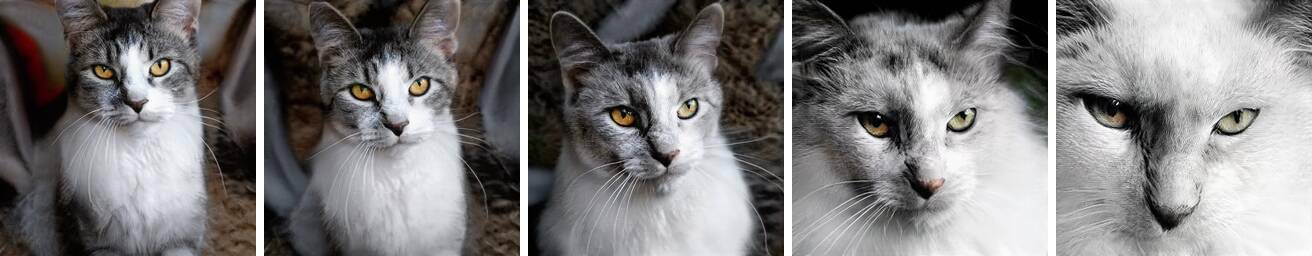}\\
\includegraphics[width=\h]{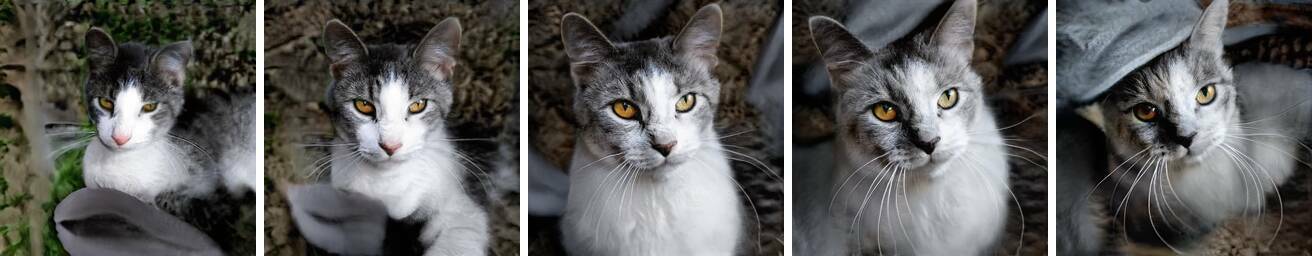}\hspace*{1cm}
\includegraphics[width=\h]{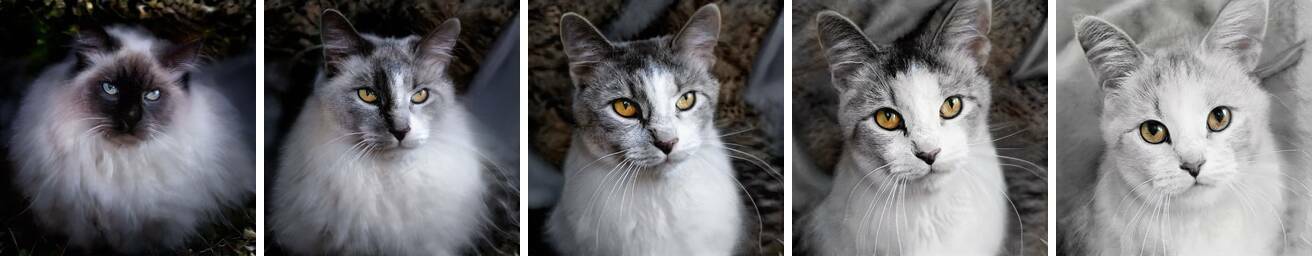}\\
\includegraphics[width=\h]{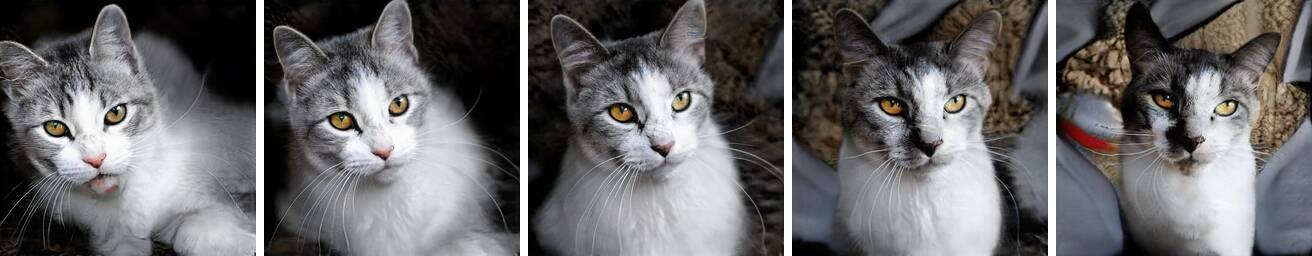}\hspace*{1cm}
\includegraphics[width=\h]{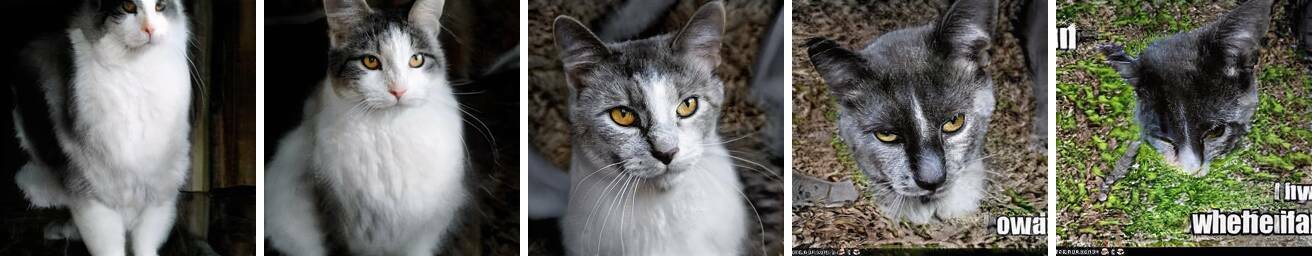}\\
\includegraphics[width=\h]{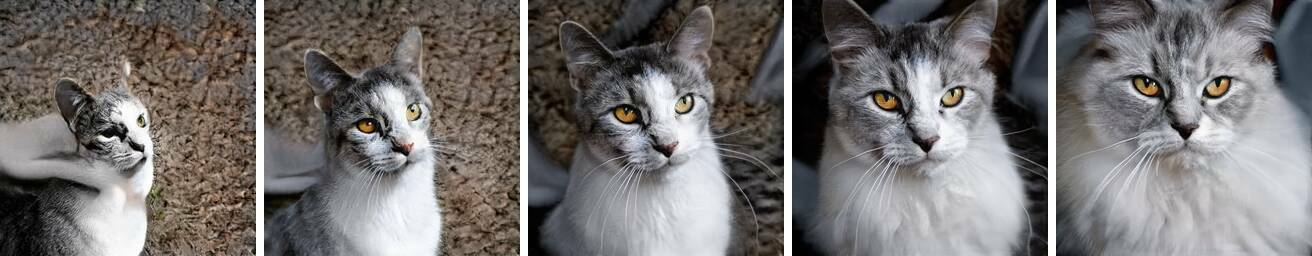}\hspace*{1cm}
\includegraphics[width=\h]{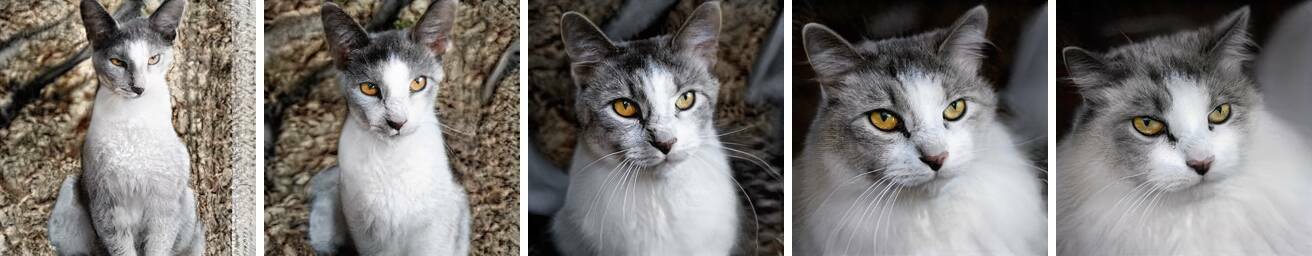}\\
\includegraphics[width=\h]{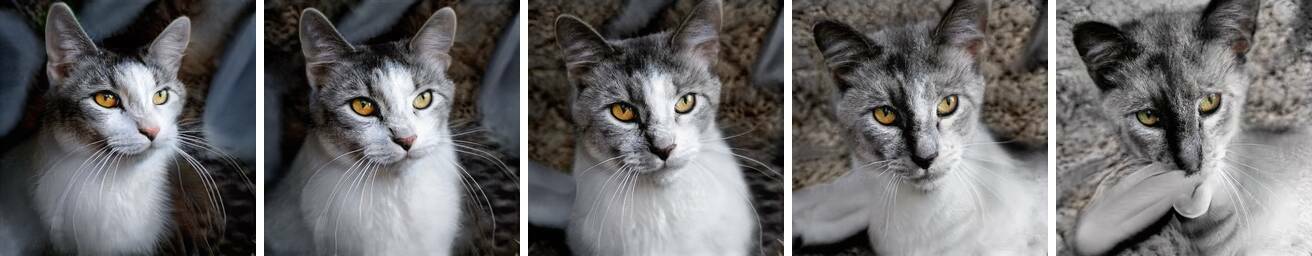}\hspace*{1cm}
\includegraphics[width=\h]{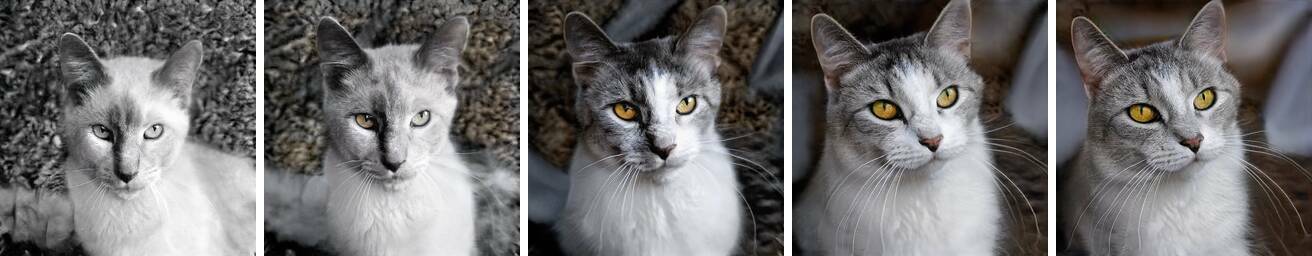}\\
\includegraphics[width=\h]{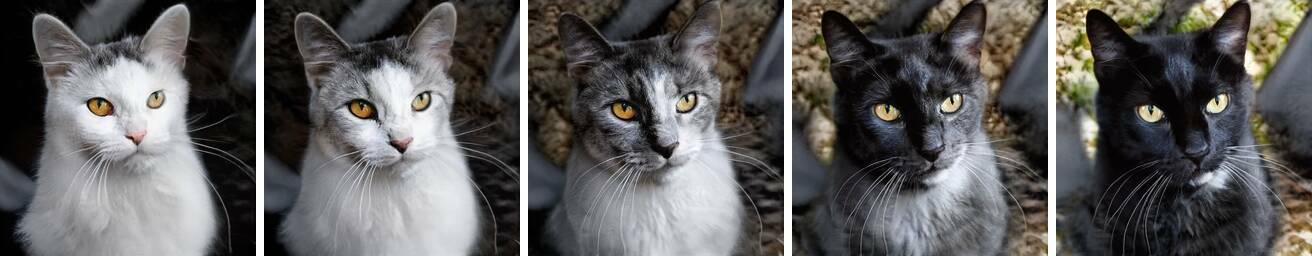}\hspace*{1cm}
\includegraphics[width=\h]{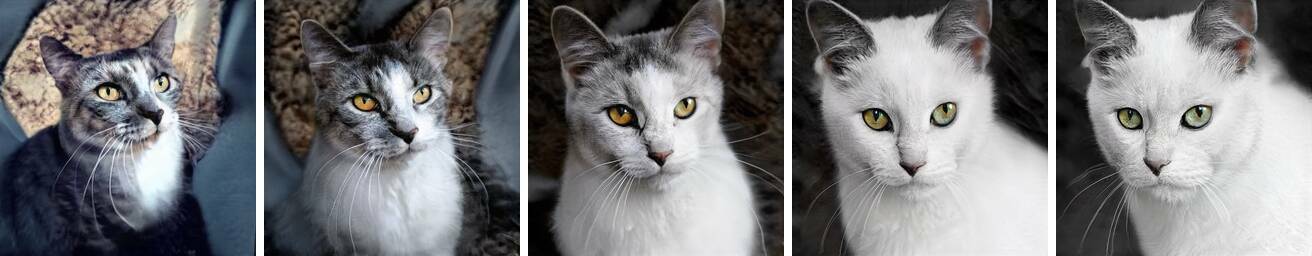}\\
\includegraphics[width=\h]{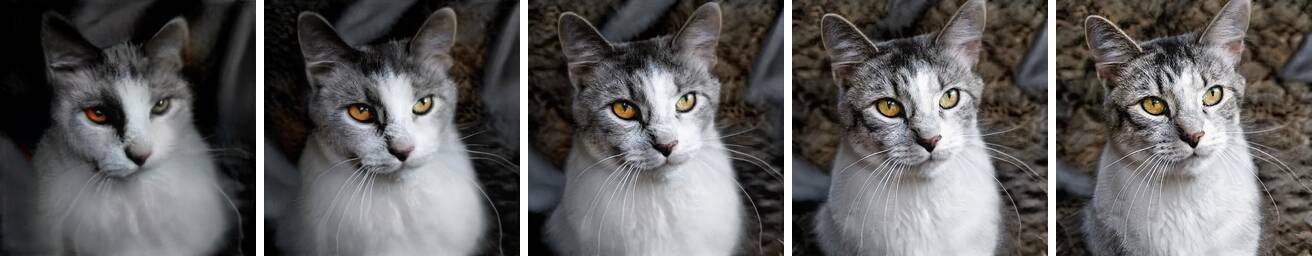}\hspace*{1cm}
\includegraphics[width=\h]{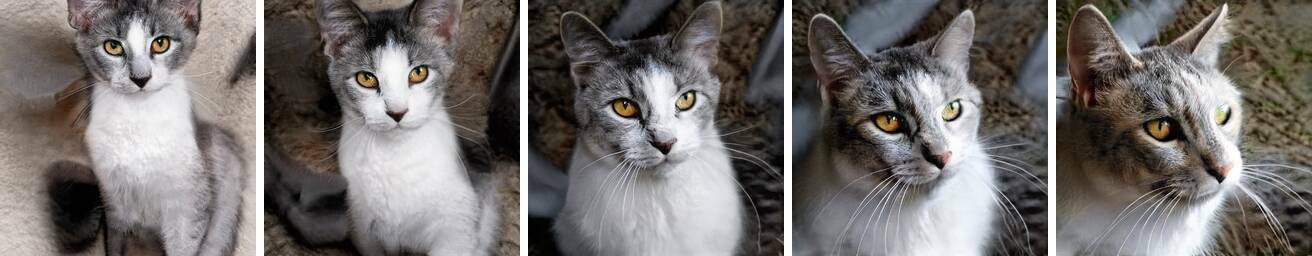}\\
\includegraphics[width=\h]{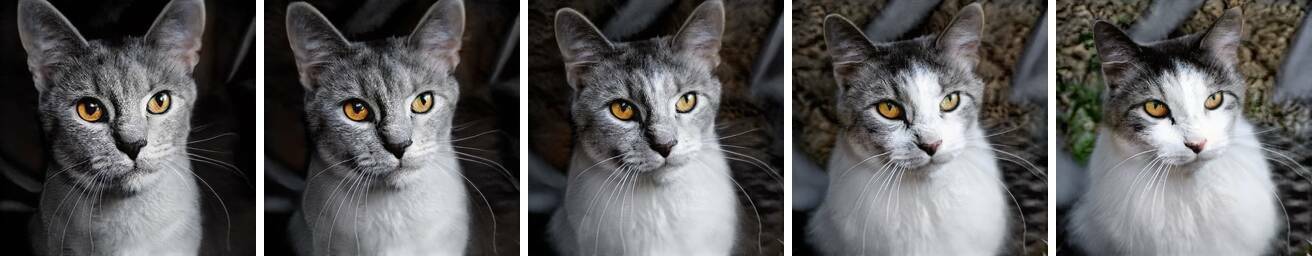}\hspace*{1cm}
\includegraphics[width=\h]{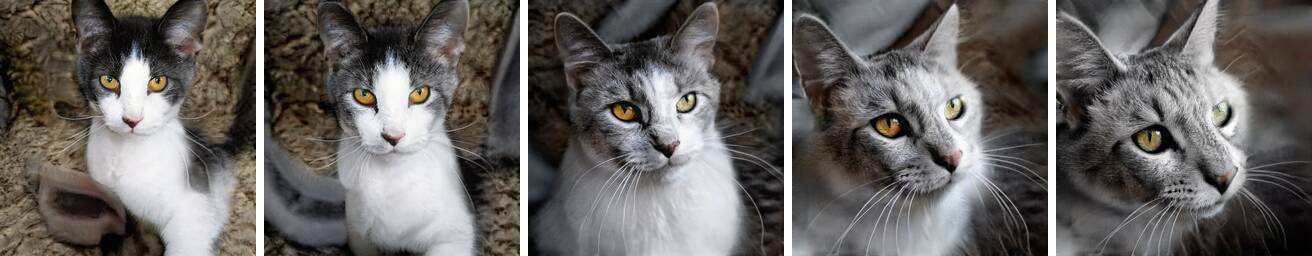}\\
\includegraphics[width=\h]{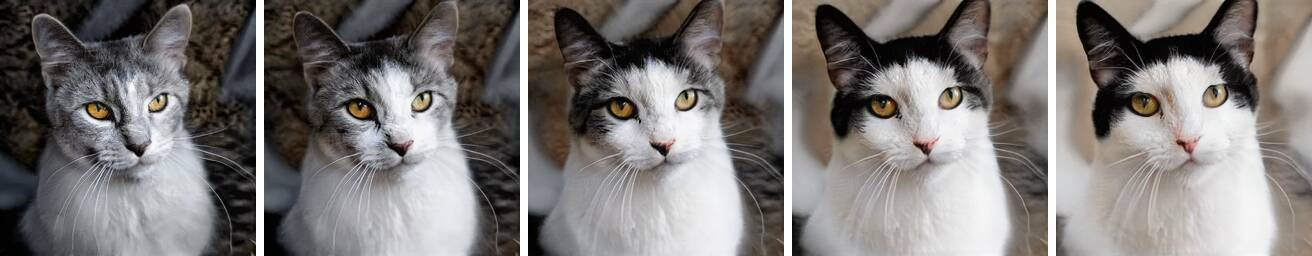}\hspace*{1cm}
\includegraphics[width=\h]{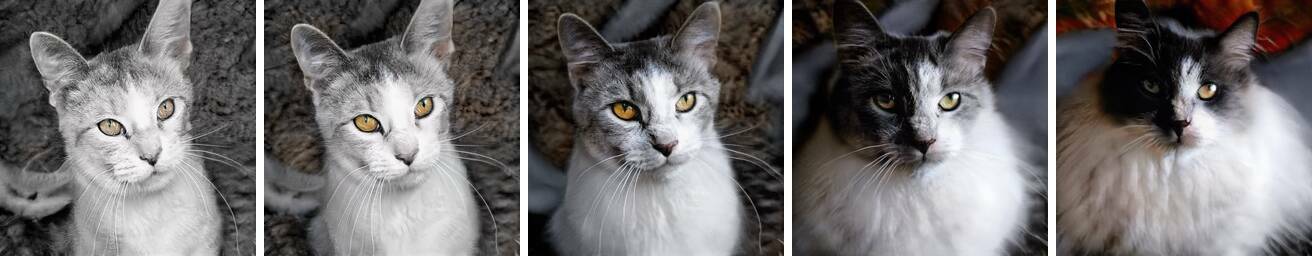}\\
\includegraphics[width=\h]{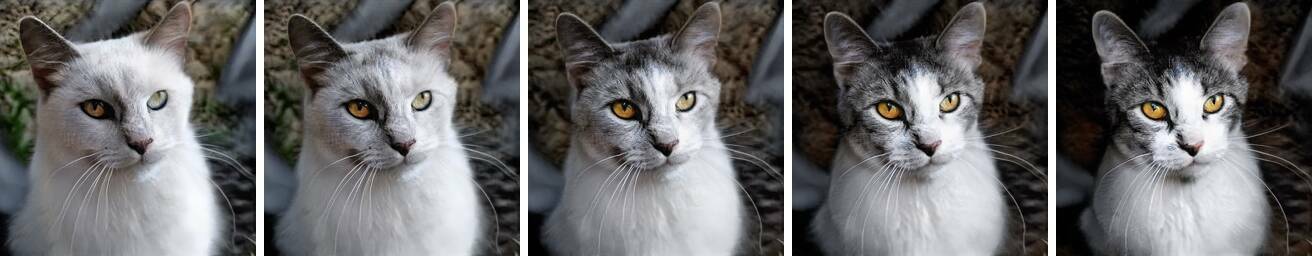}\hspace*{1cm}
\includegraphics[width=\h]{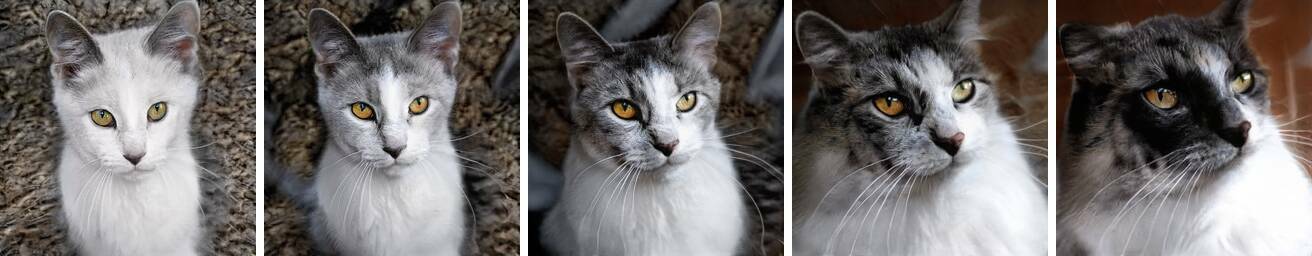}\\
\includegraphics[width=\h]{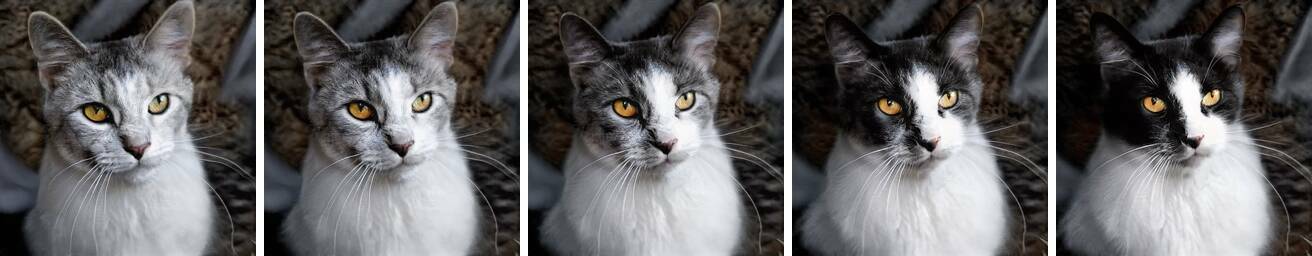}\hspace*{1cm}
\includegraphics[width=\h]{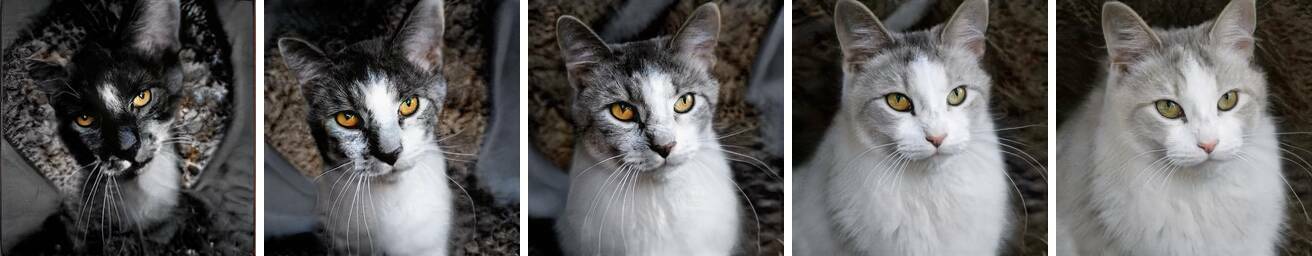}\\
\includegraphics[width=\h]{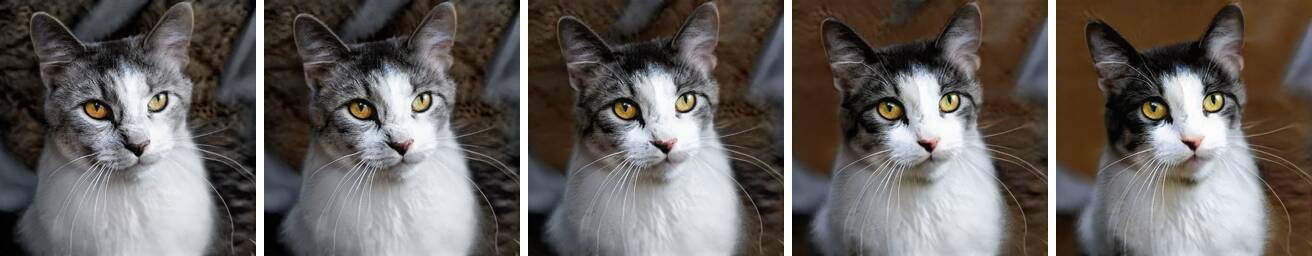}\hspace*{1cm}
\includegraphics[width=\h]{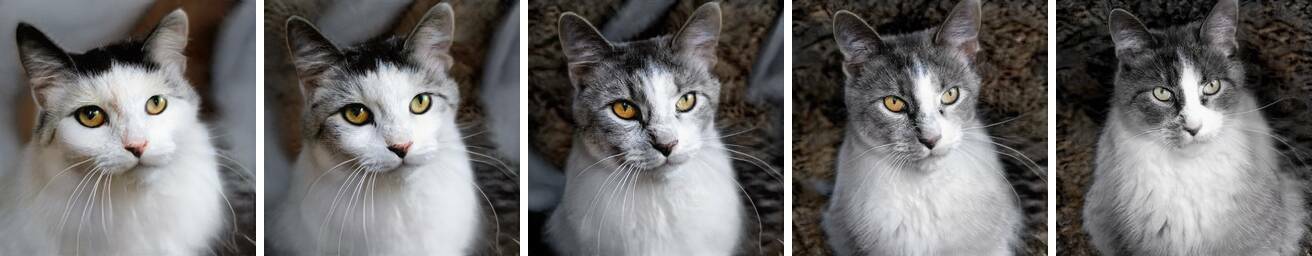}\\
\includegraphics[width=\h]{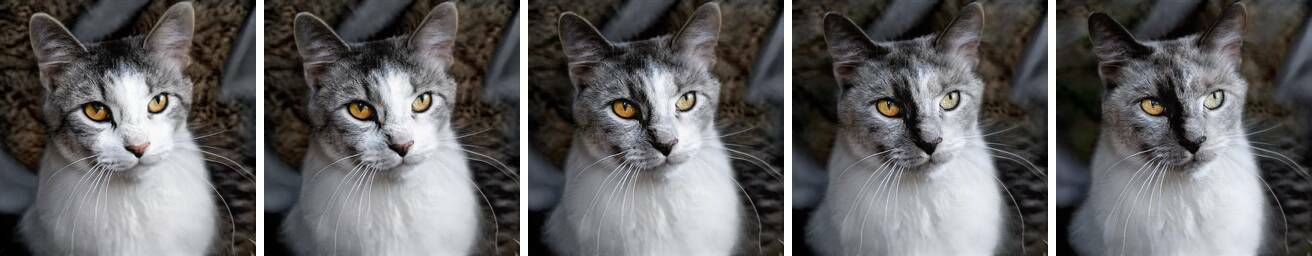}\hspace*{1cm}
\includegraphics[width=\h]{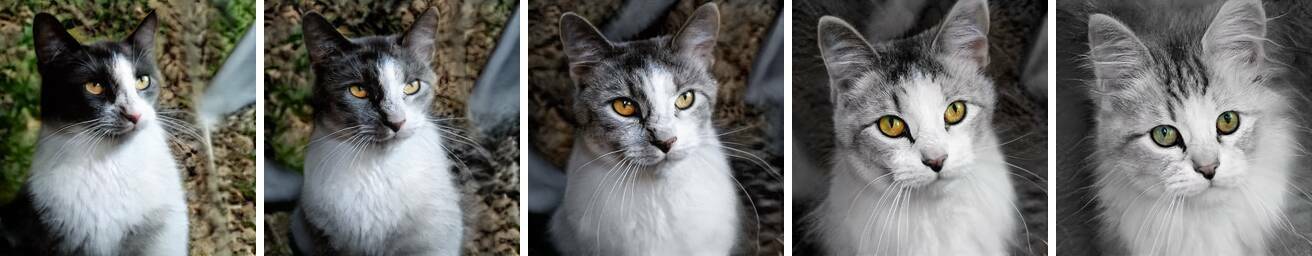}\\
\includegraphics[width=\h]{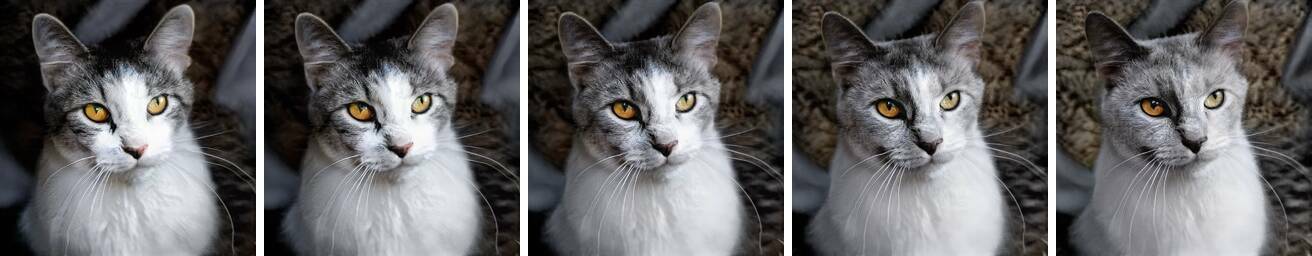}\hspace*{1cm}
\includegraphics[width=\h]{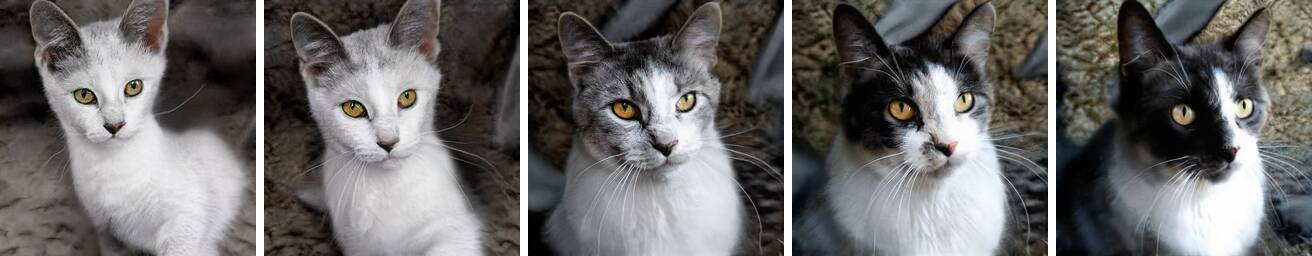}\\
\includegraphics[width=\h]{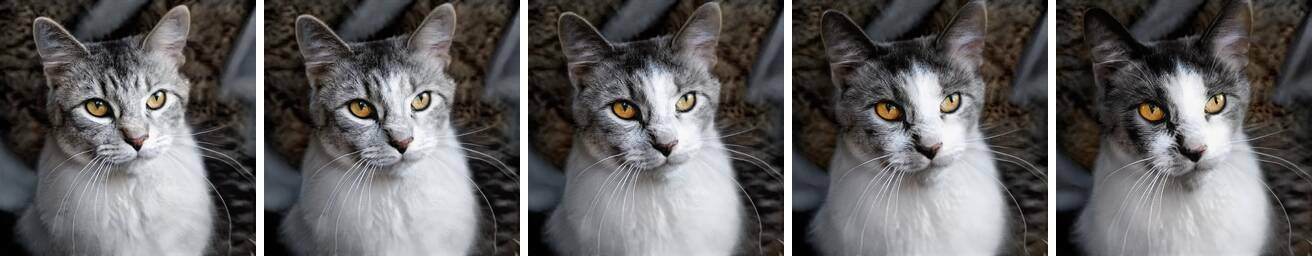}\hspace*{1cm}
\includegraphics[width=\h]{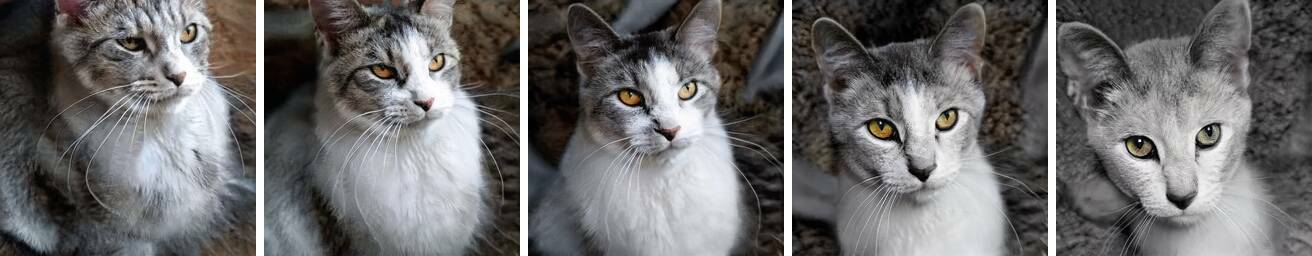}\\
\includegraphics[width=\h]{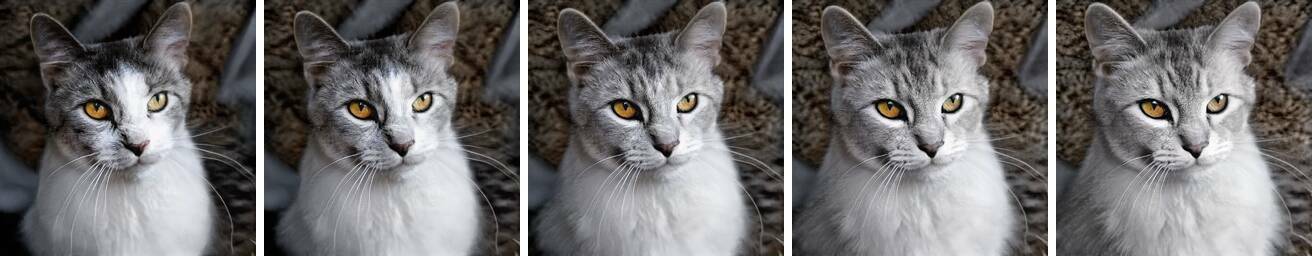}\hspace*{1cm}
\includegraphics[width=\h]{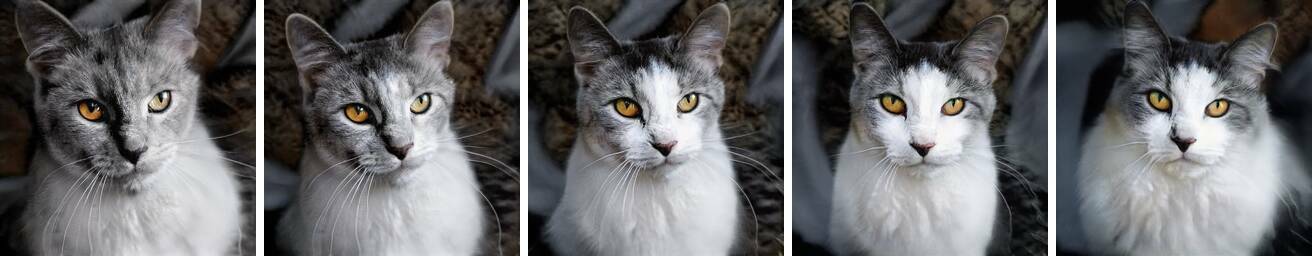}\\
\makebox[0.4\linewidth]{(a) Principal components $\mathbf{v_{0}}-\mathbf{v_{19}}$, $\pm 2\sigma$}\hspace*{1cm}
\makebox[0.4\linewidth]{(b) Normally distributed directions in $\mathcal{Z}$, $\pm 10\hat{r_i}$}%

\caption{\label{fig:topPCsCats} A visualization of the first 20 principal components of StyleGAN2 Cats (a), and of 20 isotropic Gaussian directions in $\mathcal{Z}$ (b). The random directions are scaled to emphasize their effect.}
\end{figure*}
}

\newcommand{\figTopPCsSGtwoCars}{
\renewcommand{\h}{0.38\linewidth}
\begin{figure*}[t]
\centering
\includegraphics[width=\h]{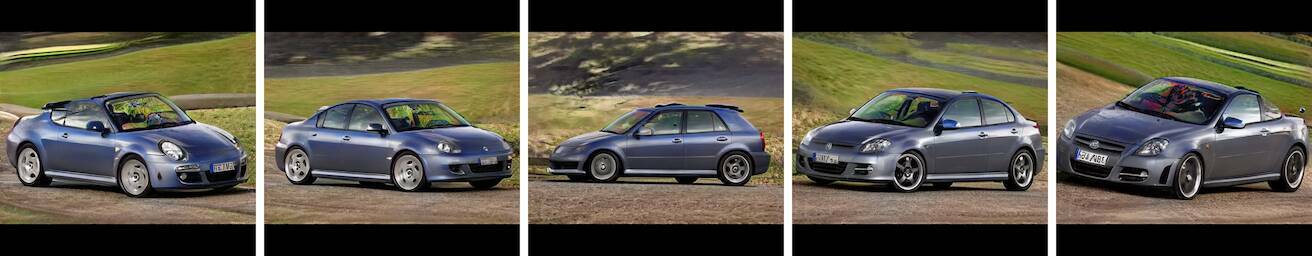}\hspace*{1cm}
\includegraphics[width=\h]{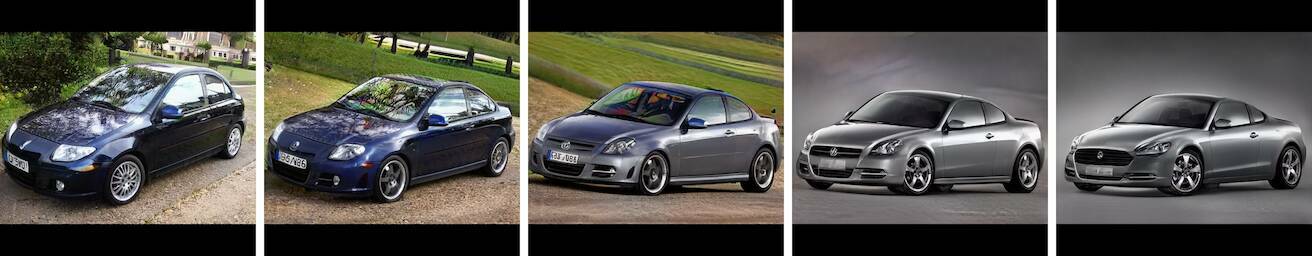}\\
\includegraphics[width=\h]{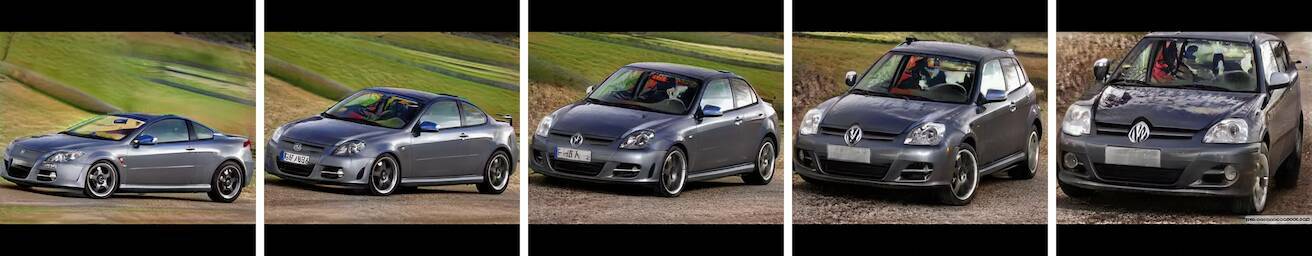}\hspace*{1cm}
\includegraphics[width=\h]{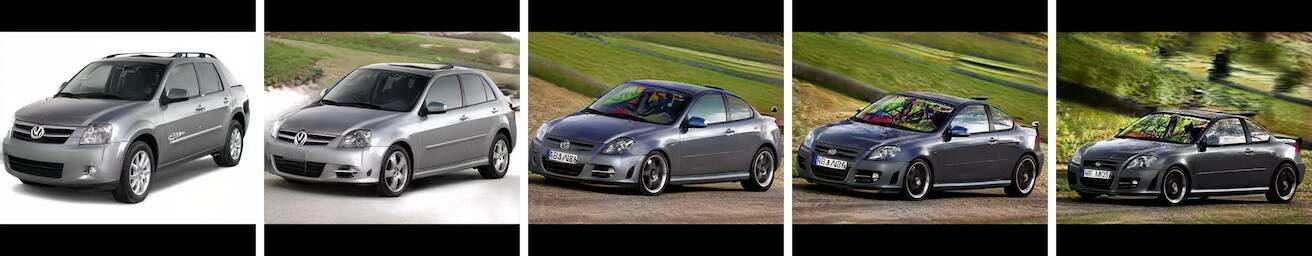}\\
\includegraphics[width=\h]{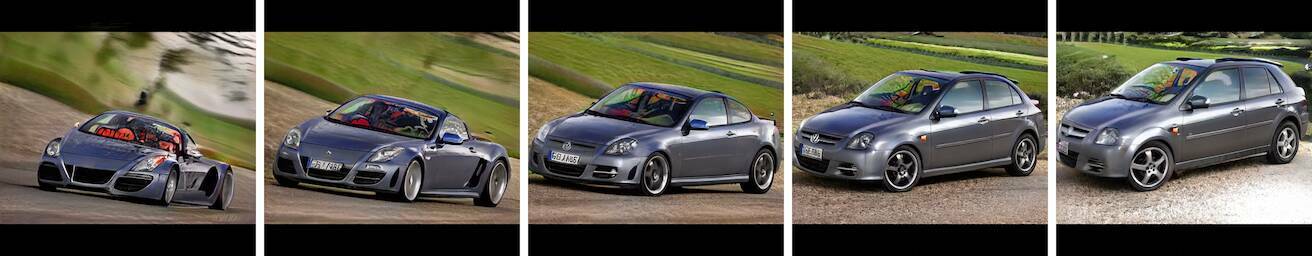}\hspace*{1cm}
\includegraphics[width=\h]{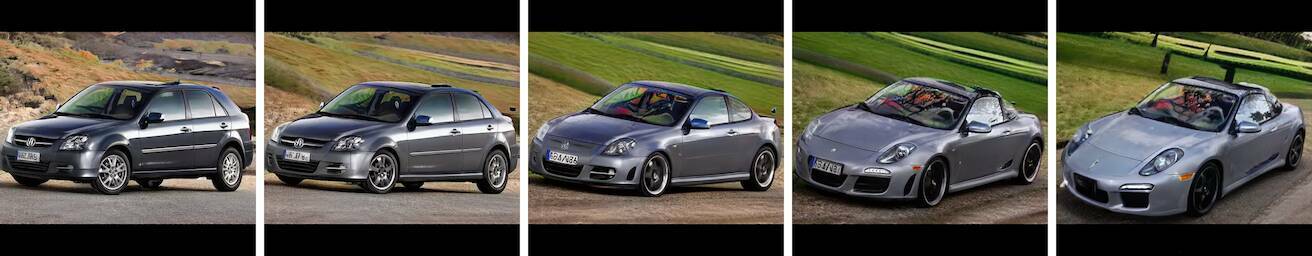}\\
\includegraphics[width=\h]{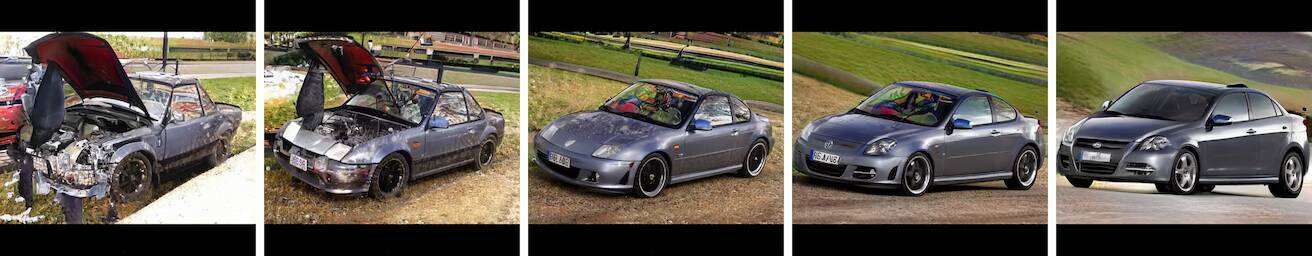}\hspace*{1cm}
\includegraphics[width=\h]{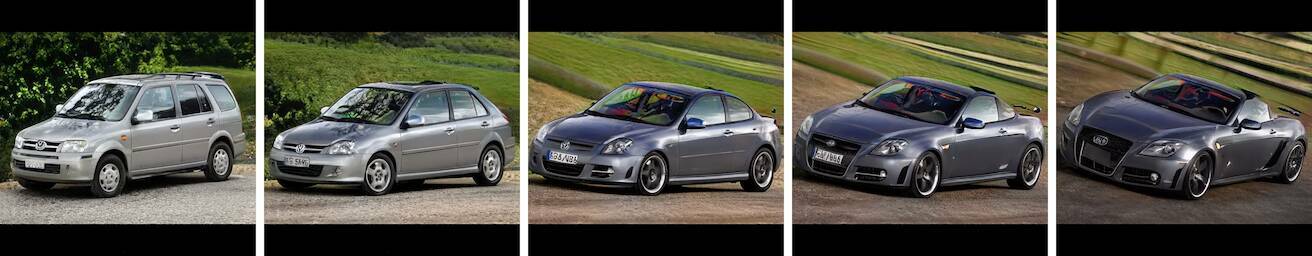}\\
\includegraphics[width=\h]{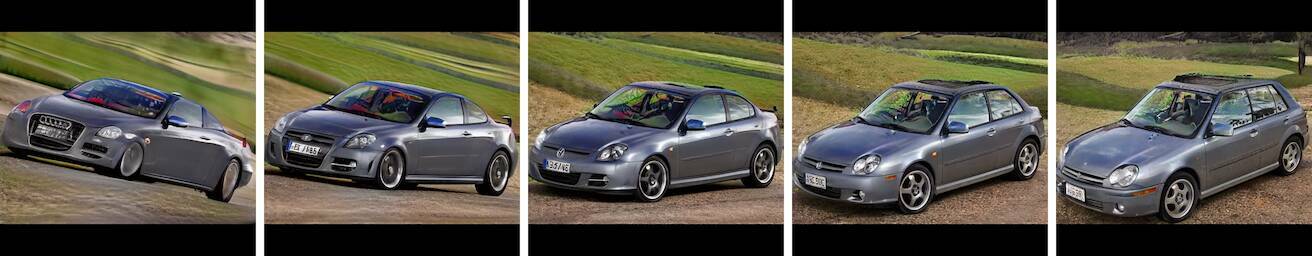}\hspace*{1cm}
\includegraphics[width=\h]{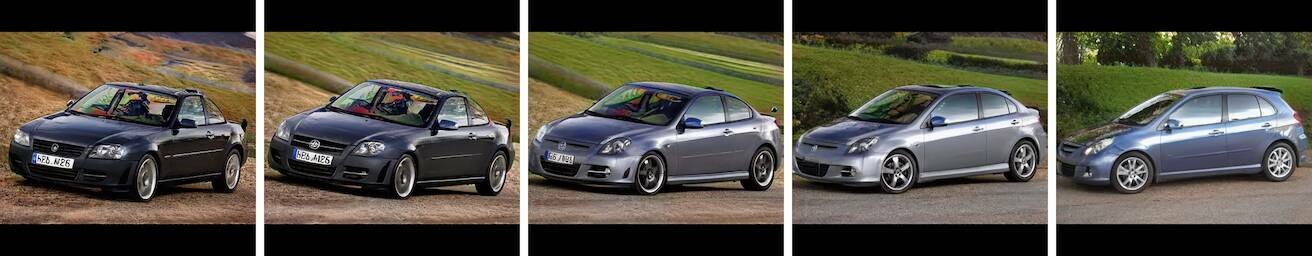}\\
\includegraphics[width=\h]{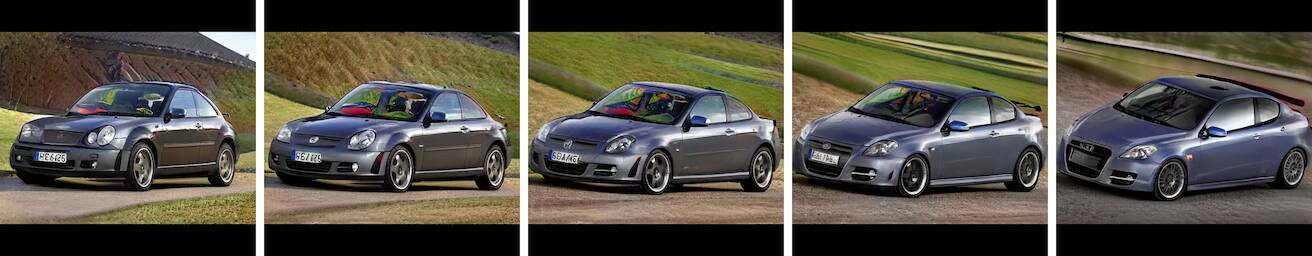}\hspace*{1cm}
\includegraphics[width=\h]{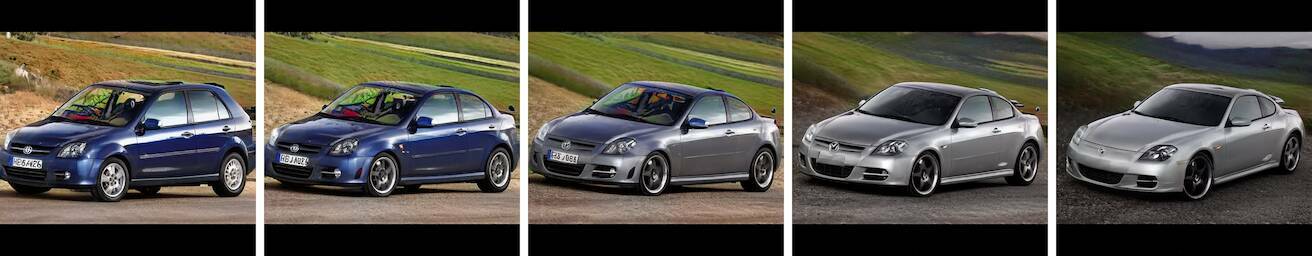}\\
\includegraphics[width=\h]{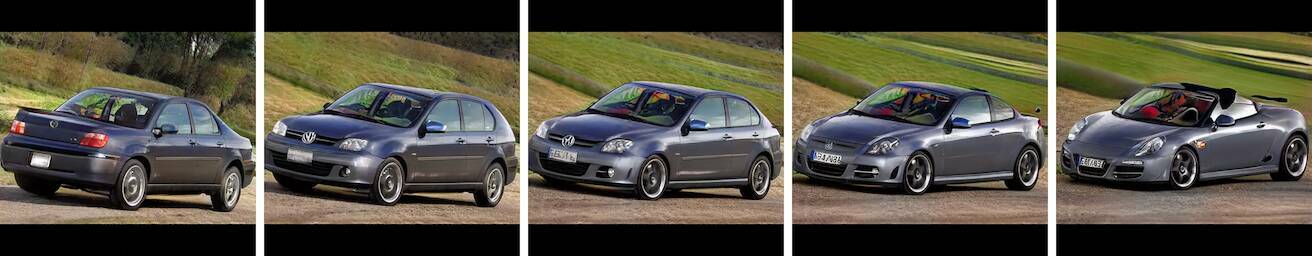}\hspace*{1cm}
\includegraphics[width=\h]{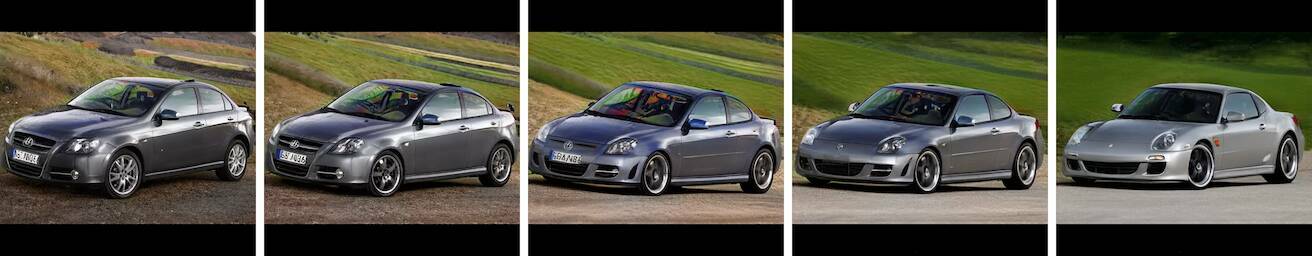}\\
\includegraphics[width=\h]{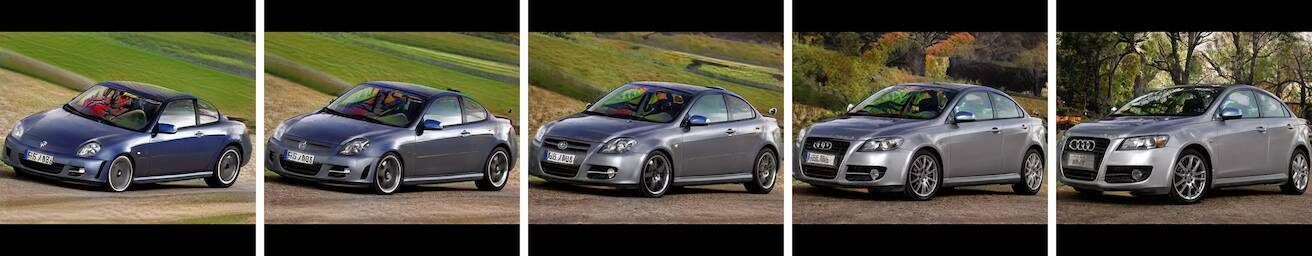}\hspace*{1cm}
\includegraphics[width=\h]{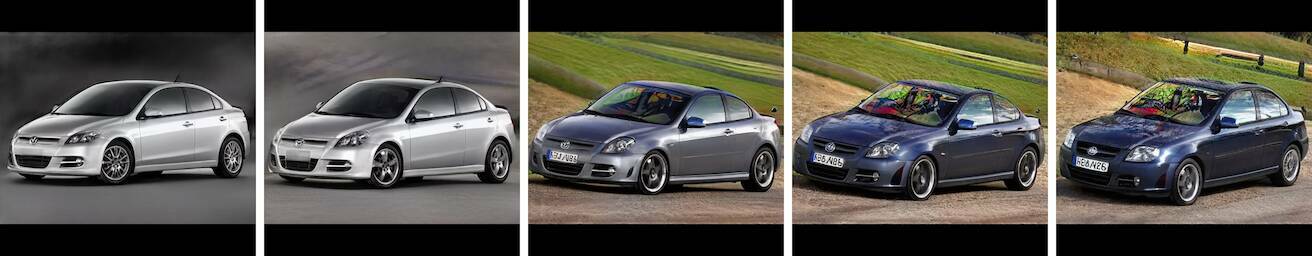}\\
\includegraphics[width=\h]{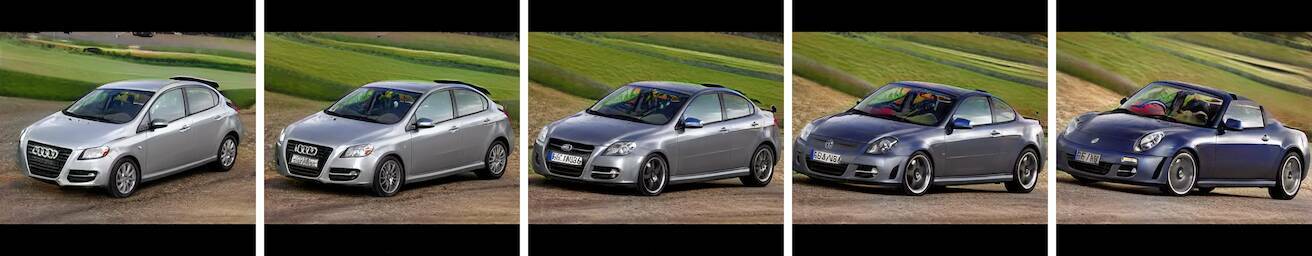}\hspace*{1cm}
\includegraphics[width=\h]{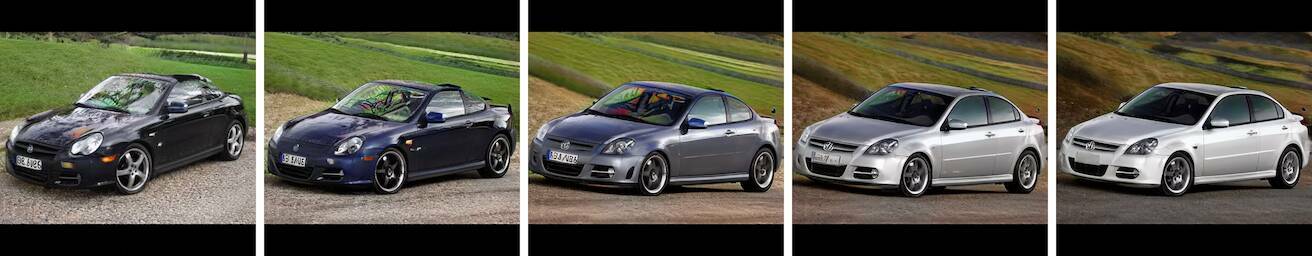}\\
\includegraphics[width=\h]{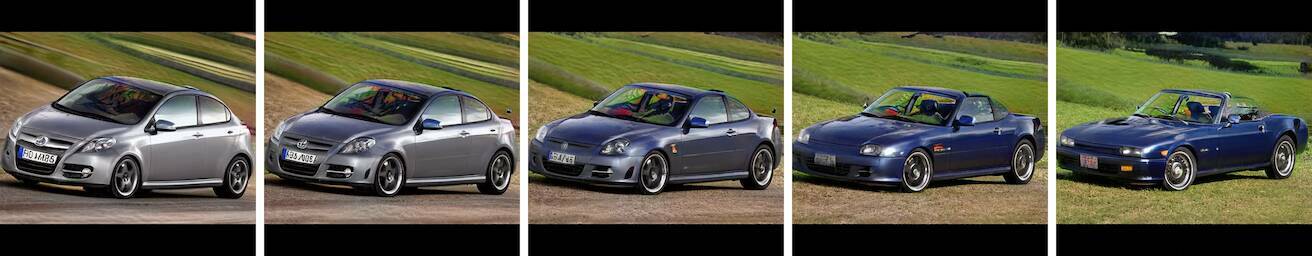}\hspace*{1cm}
\includegraphics[width=\h]{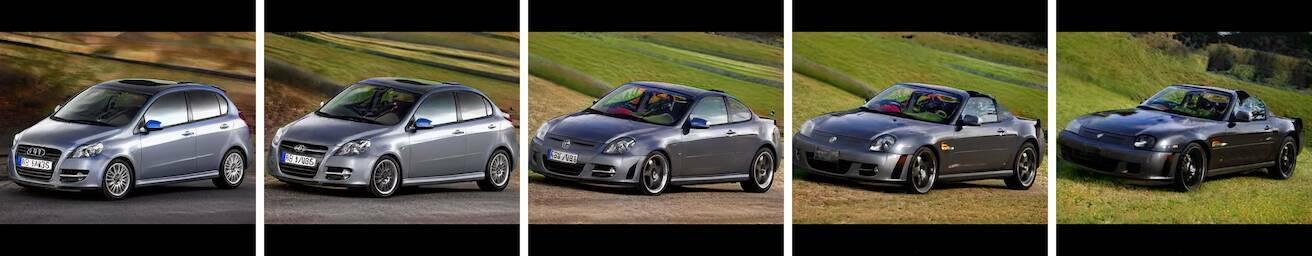}\\
\includegraphics[width=\h]{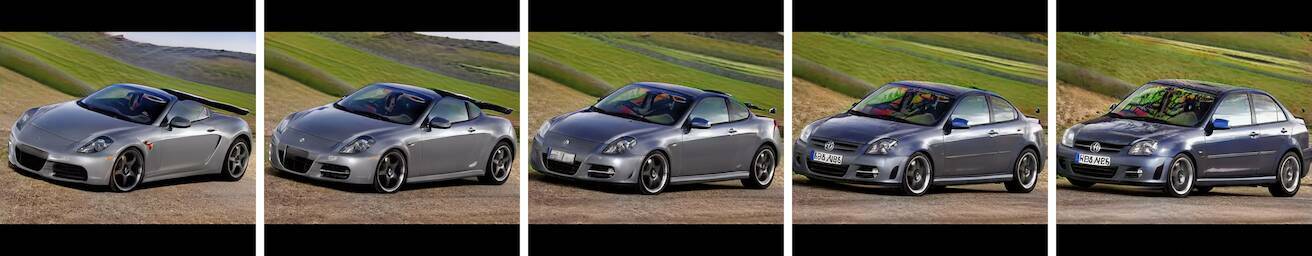}\hspace*{1cm}
\includegraphics[width=\h]{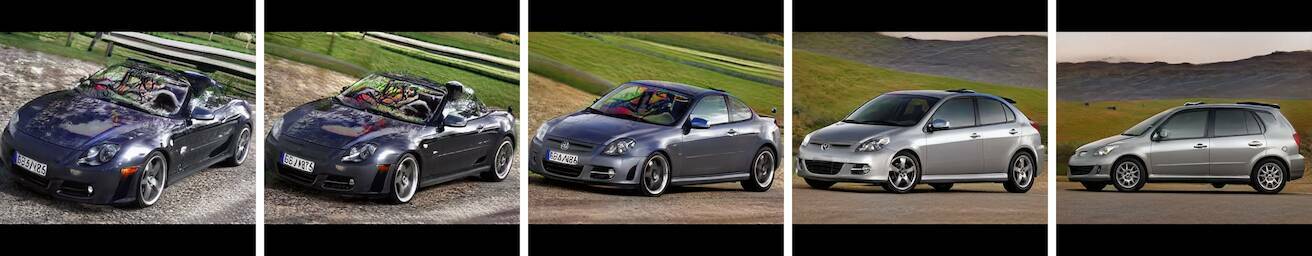}\\
\includegraphics[width=\h]{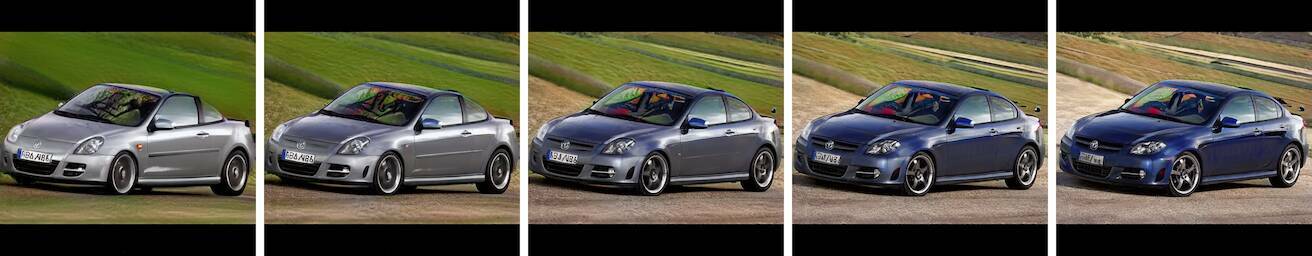}\hspace*{1cm}
\includegraphics[width=\h]{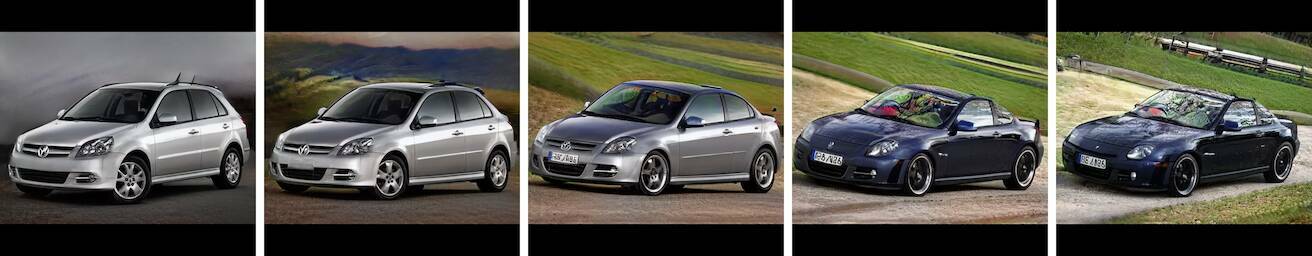}\\
\includegraphics[width=\h]{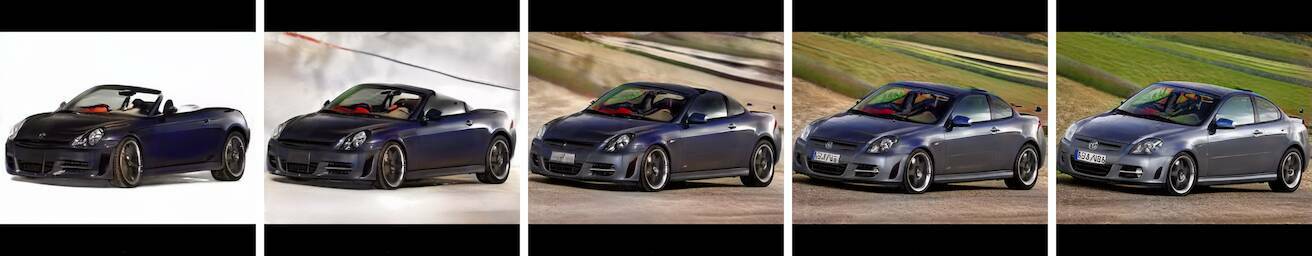}\hspace*{1cm}
\includegraphics[width=\h]{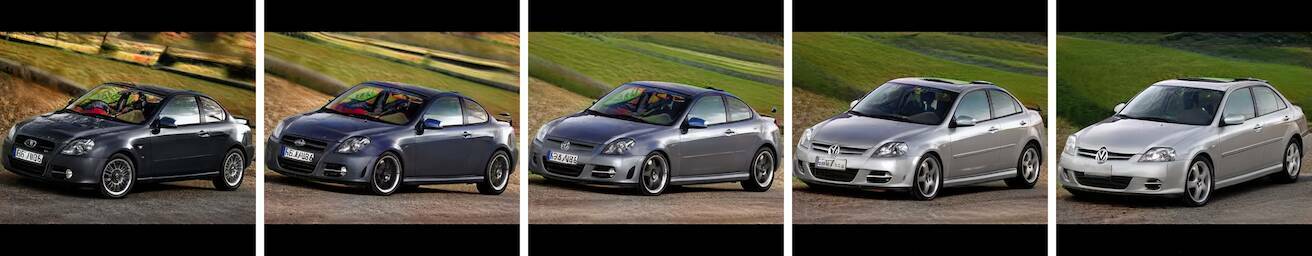}\\
\includegraphics[width=\h]{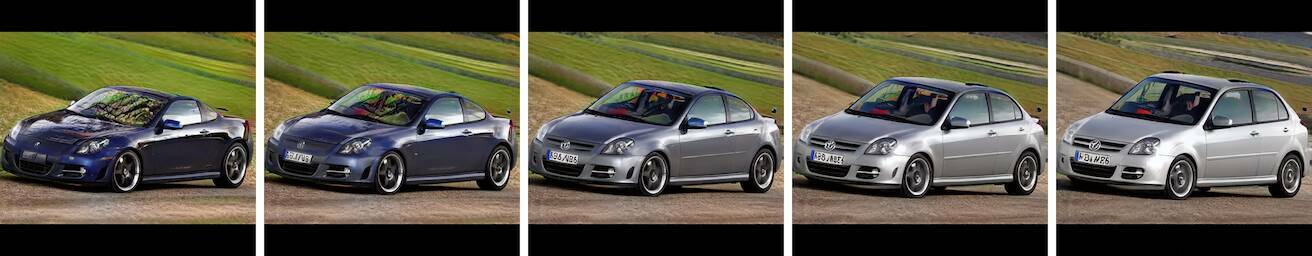}\hspace*{1cm}
\includegraphics[width=\h]{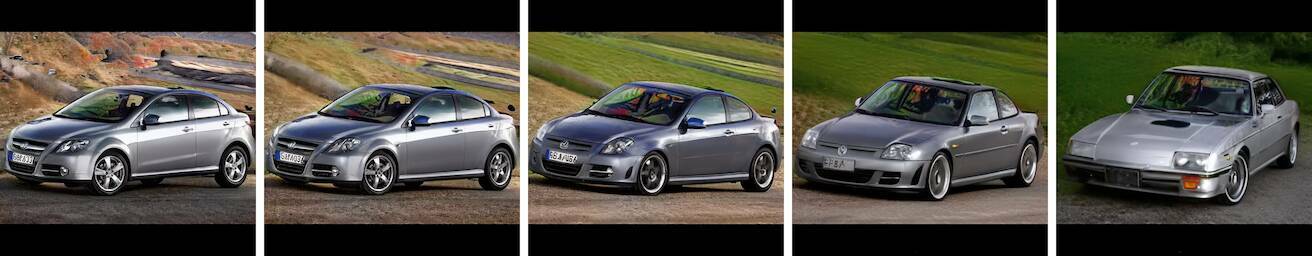}\\
\includegraphics[width=\h]{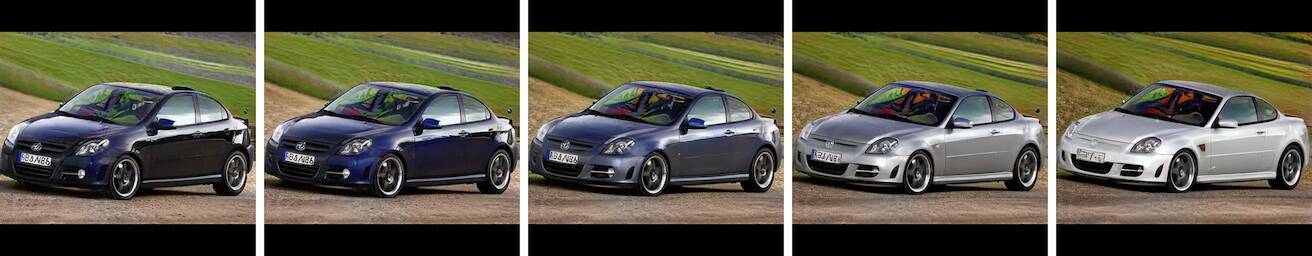}\hspace*{1cm}
\includegraphics[width=\h]{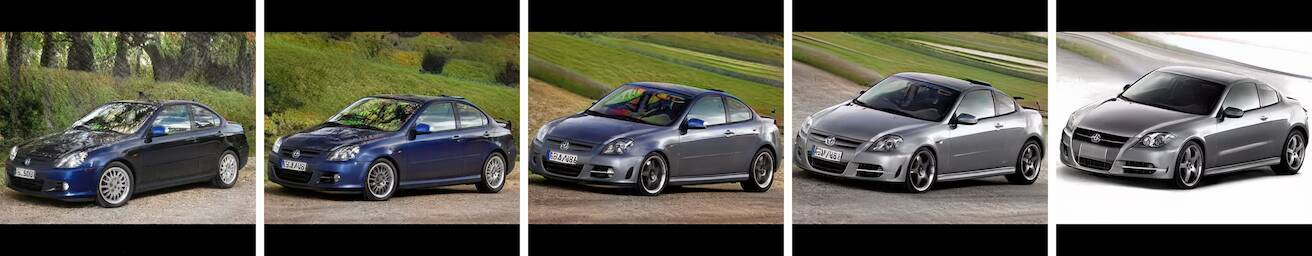}\\
\includegraphics[width=\h]{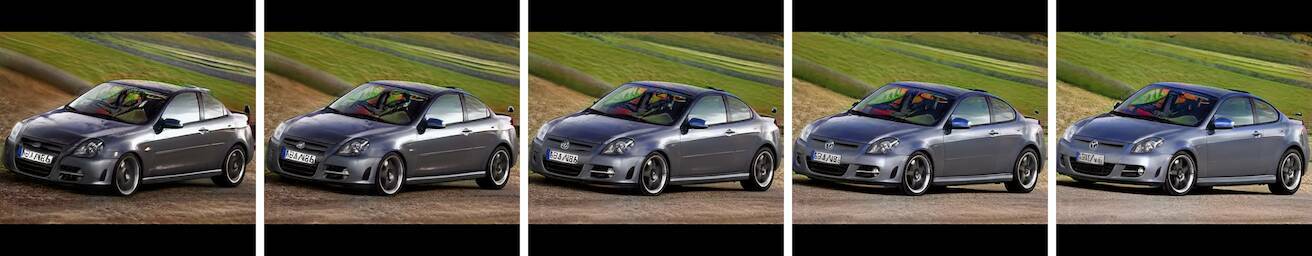}\hspace*{1cm}
\includegraphics[width=\h]{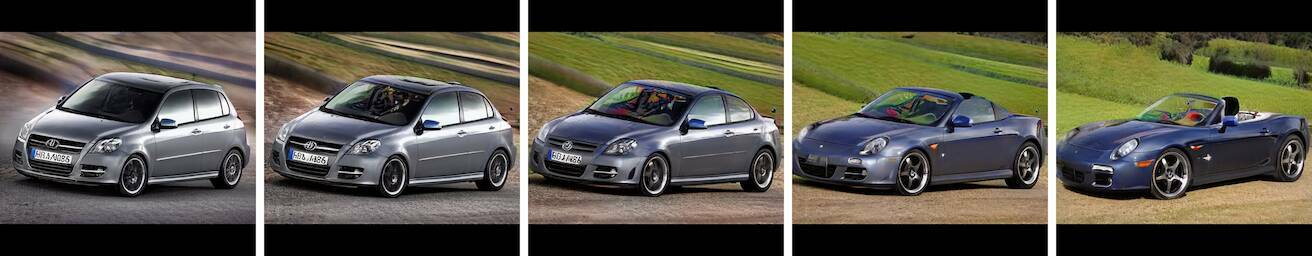}\\
\includegraphics[width=\h]{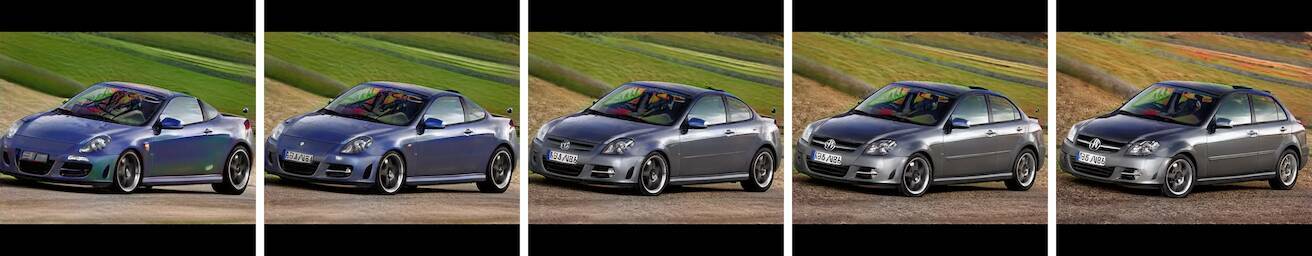}\hspace*{1cm}
\includegraphics[width=\h]{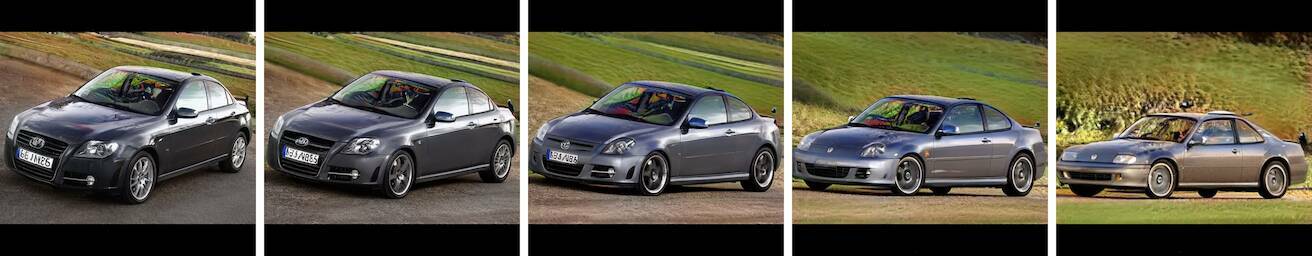}\\
\includegraphics[width=\h]{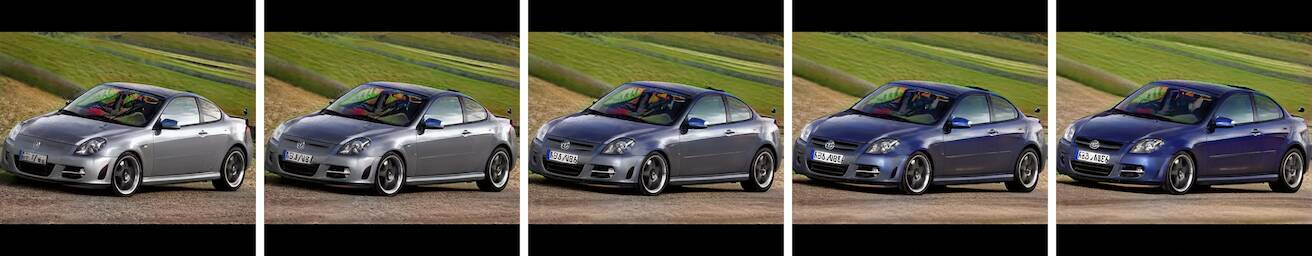}\hspace*{1cm}
\includegraphics[width=\h]{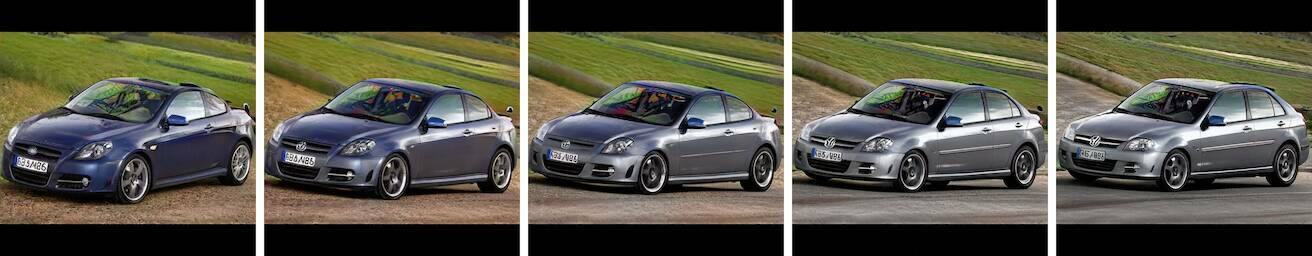}\\
\includegraphics[width=\h]{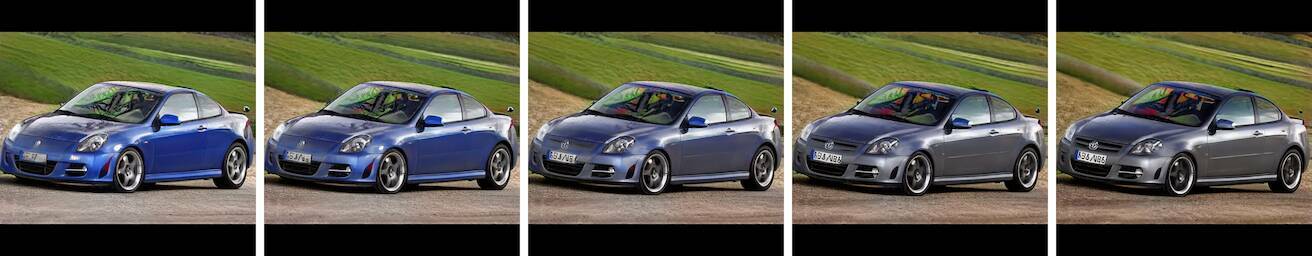}\hspace*{1cm}
\includegraphics[width=\h]{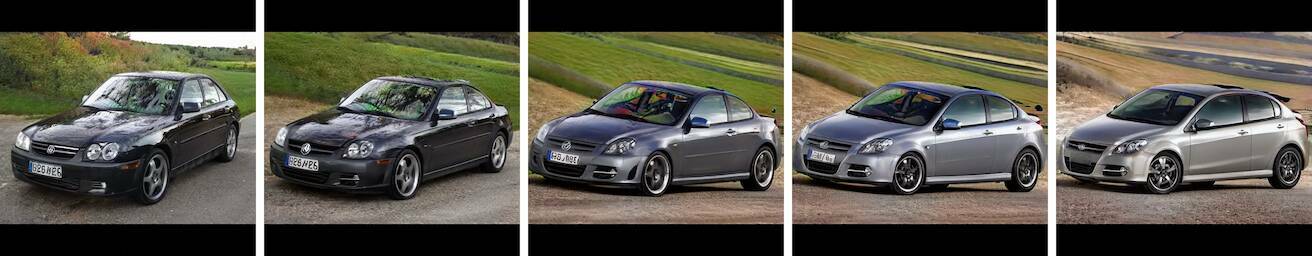}\\
\includegraphics[width=\h]{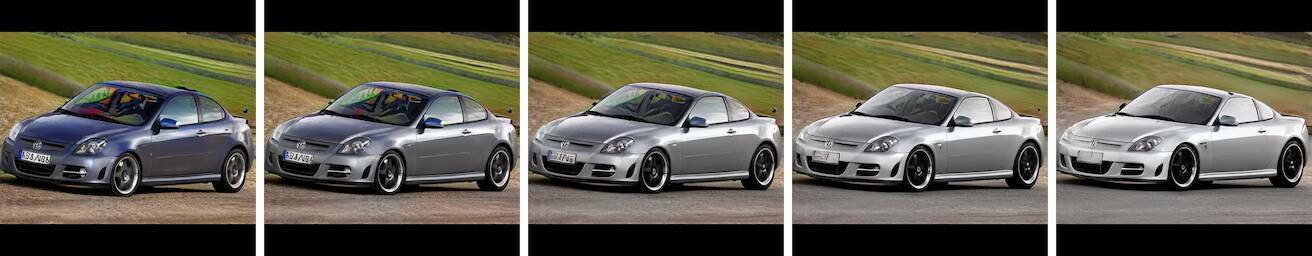}\hspace*{1cm}
\includegraphics[width=\h]{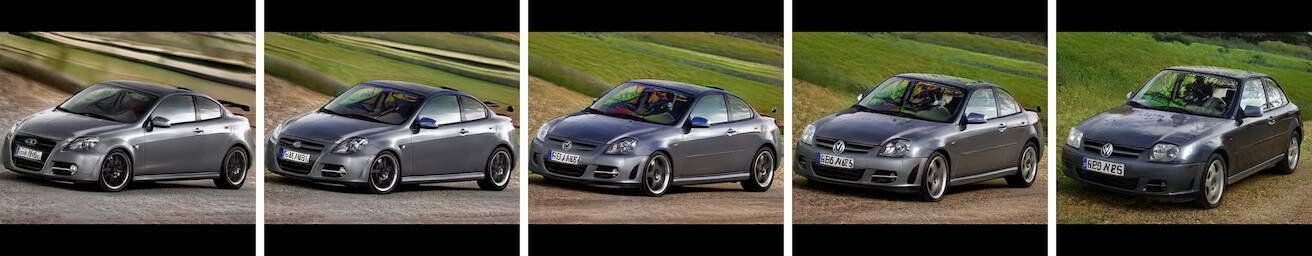}\\
\makebox[0.4\linewidth]{(a) Principal components $\mathbf{v_{0}}-\mathbf{v_{19}}$, $\pm 2\sigma$}\hspace*{1cm}
\makebox[0.4\linewidth]{(b) Normally distributed directions in $\mathcal{Z}$, $\pm 10\hat{r_i}$}%

\caption{\label{fig:topPCsCars} A visualization of the first 20 principal components of StyleGAN2 Cars (a), and of 20 isotropic Gaussian directions in $\mathcal{Z}$ (b). The random directions are scaled to emphasize their effect.}
\end{figure*}
}

\newcommand{\figTopPCsBGHusky}{
	\renewcommand{\h}{0.38\linewidth}
\begin{figure*}[t]
\centering
\includegraphics[width=\h]{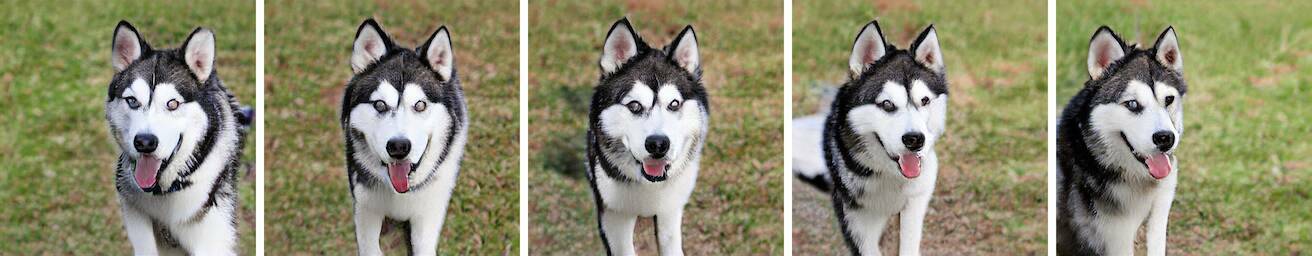}\hspace*{1cm}
\includegraphics[width=\h]{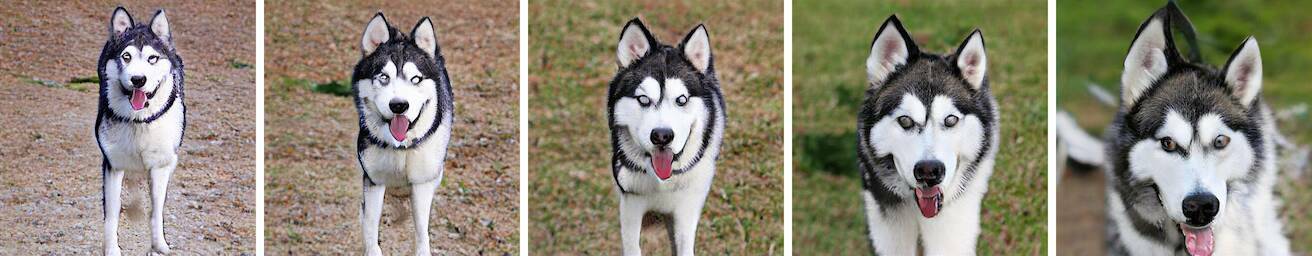}\\
\includegraphics[width=\h]{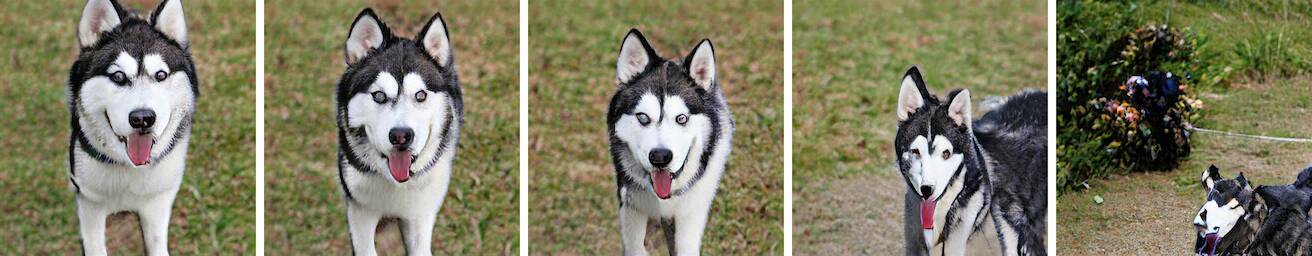}\hspace*{1cm}
\includegraphics[width=\h]{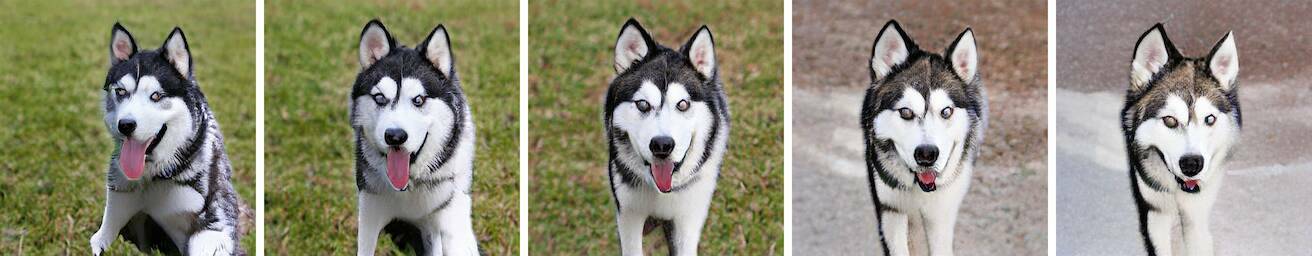}\\
\includegraphics[width=\h]{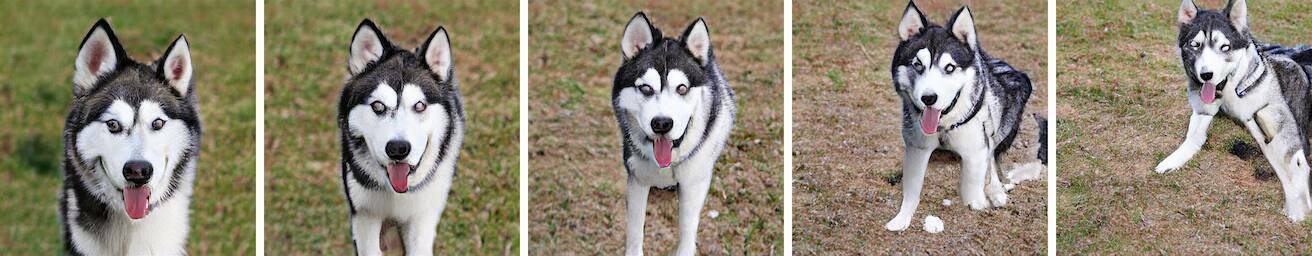}\hspace*{1cm}
\includegraphics[width=\h]{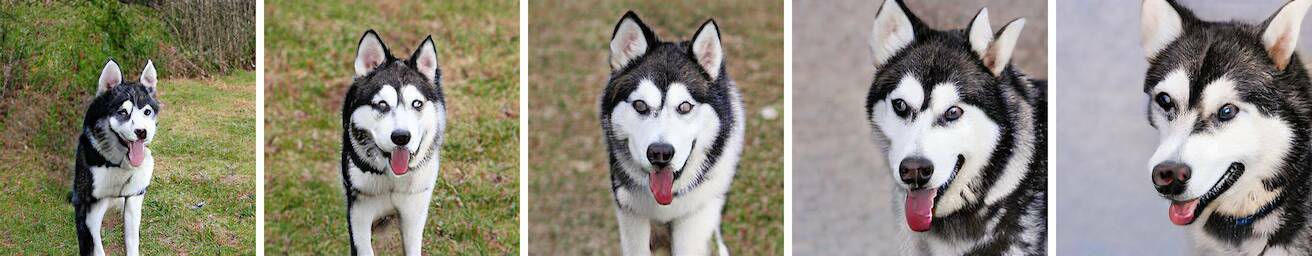}\\
\includegraphics[width=\h]{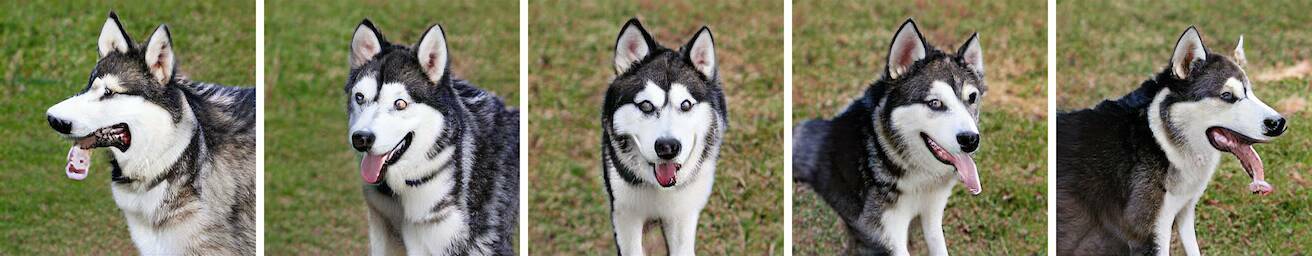}\hspace*{1cm}
\includegraphics[width=\h]{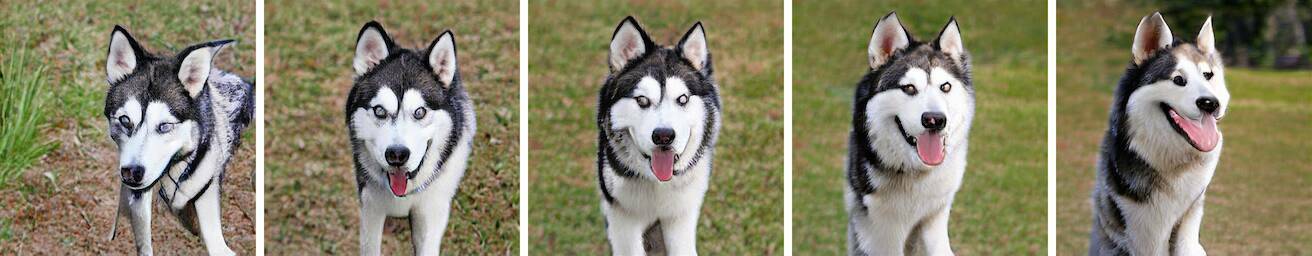}\\
\includegraphics[width=\h]{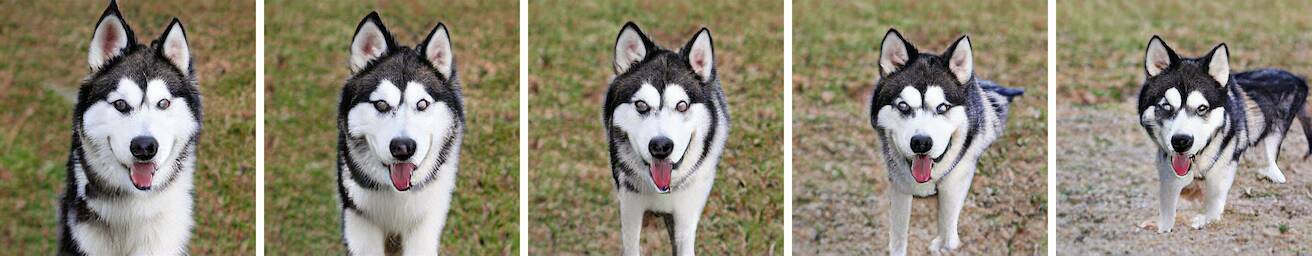}\hspace*{1cm}
\includegraphics[width=\h]{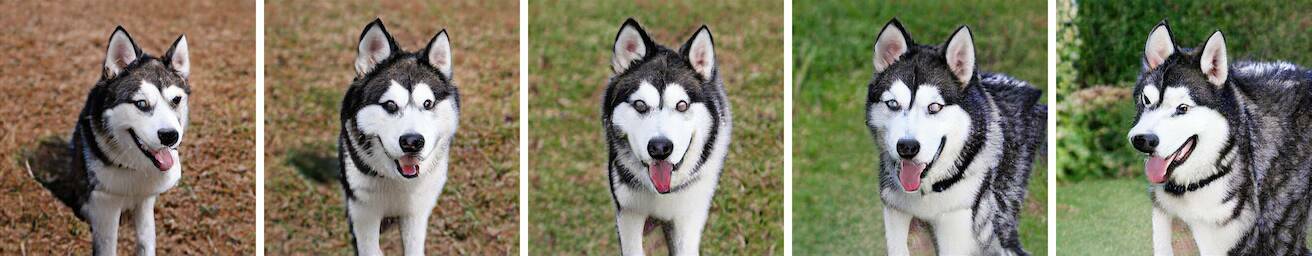}\\
\includegraphics[width=\h]{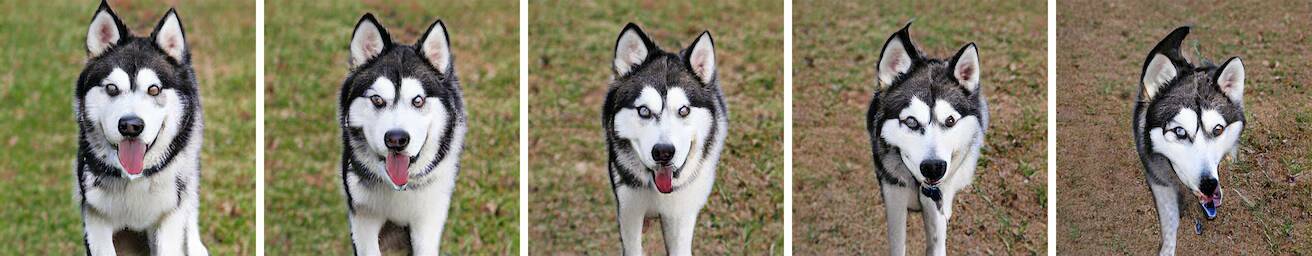}\hspace*{1cm}
\includegraphics[width=\h]{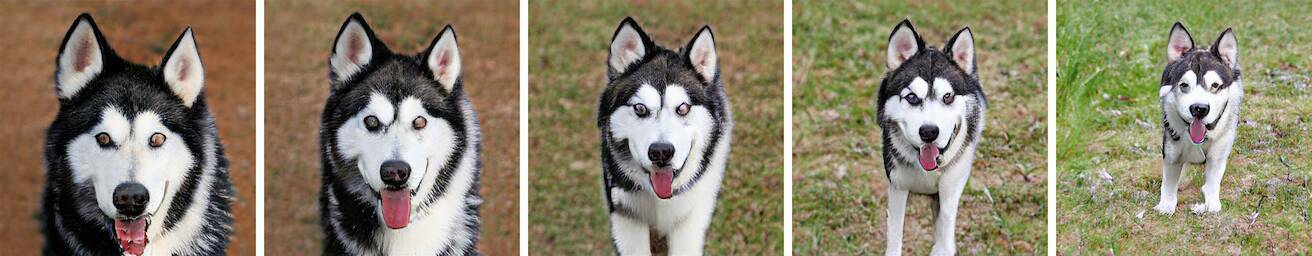}\\
\includegraphics[width=\h]{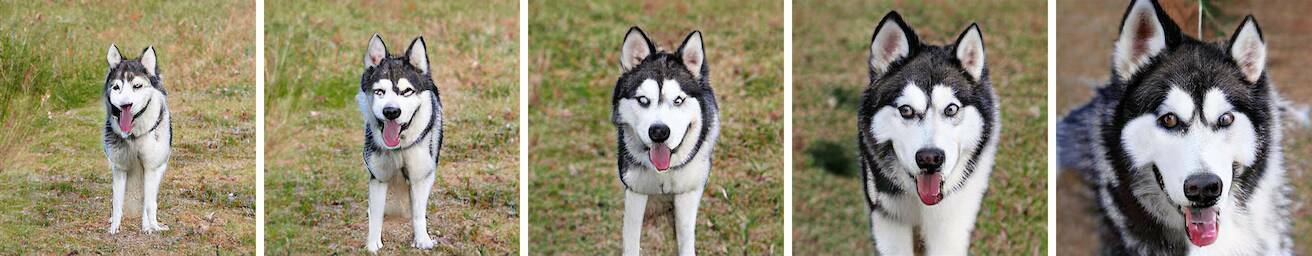}\hspace*{1cm}
\includegraphics[width=\h]{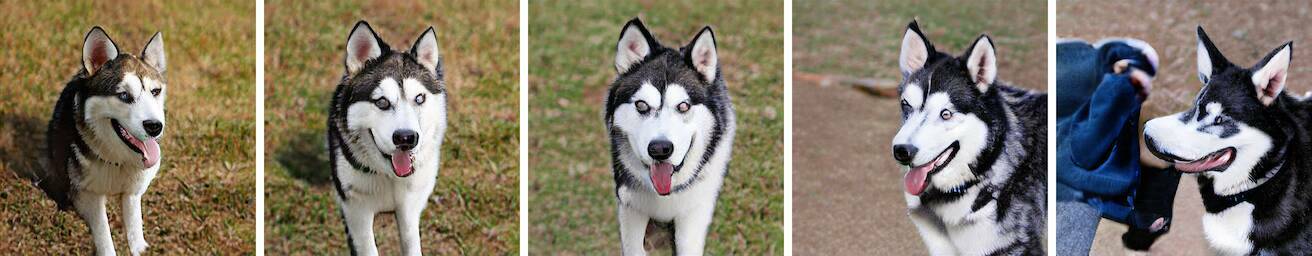}\\
\includegraphics[width=\h]{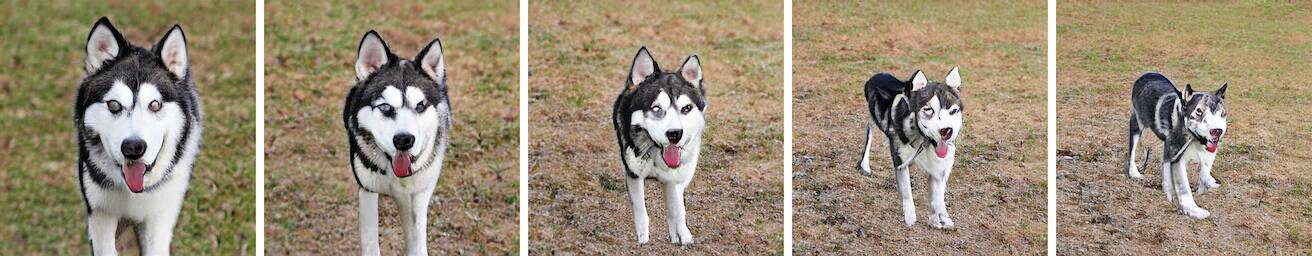}\hspace*{1cm}
\includegraphics[width=\h]{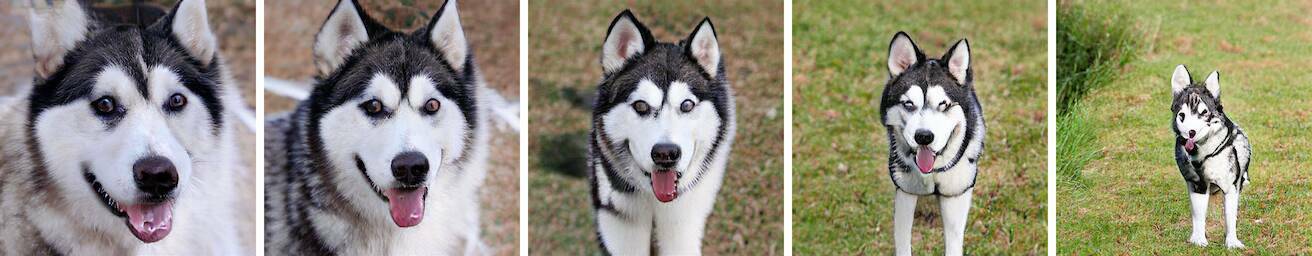}\\
\includegraphics[width=\h]{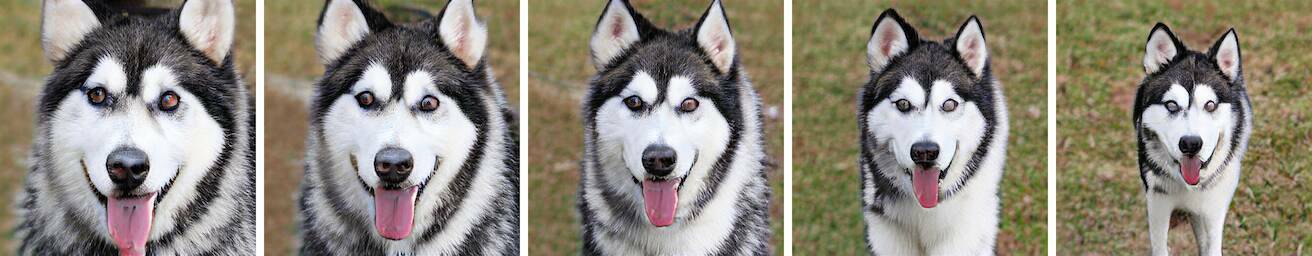}\hspace*{1cm}
\includegraphics[width=\h]{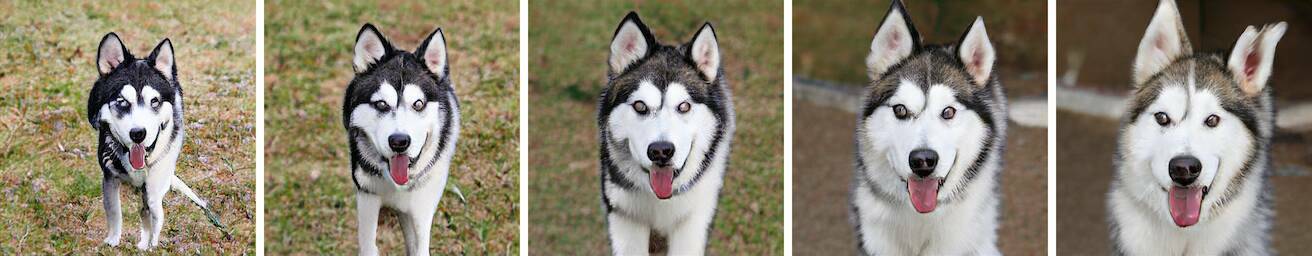}\\
\includegraphics[width=\h]{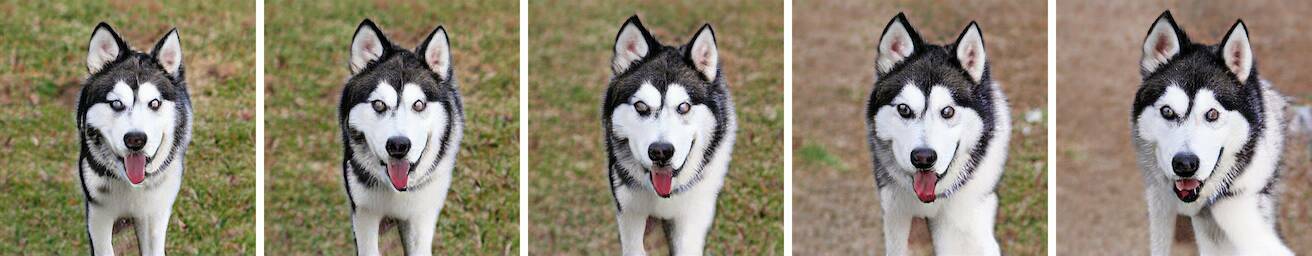}\hspace*{1cm}
\includegraphics[width=\h]{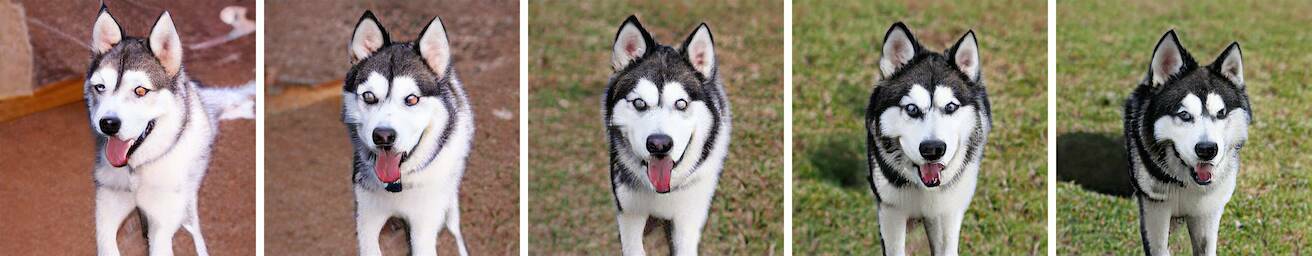}\\
\includegraphics[width=\h]{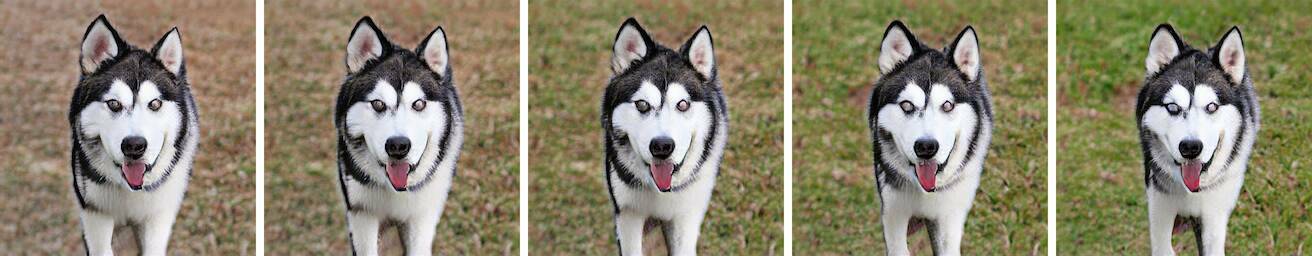}\hspace*{1cm}
\includegraphics[width=\h]{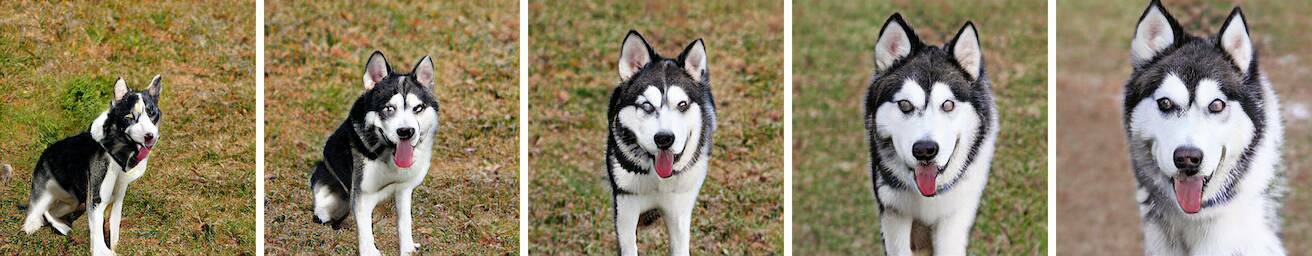}\\
\includegraphics[width=\h]{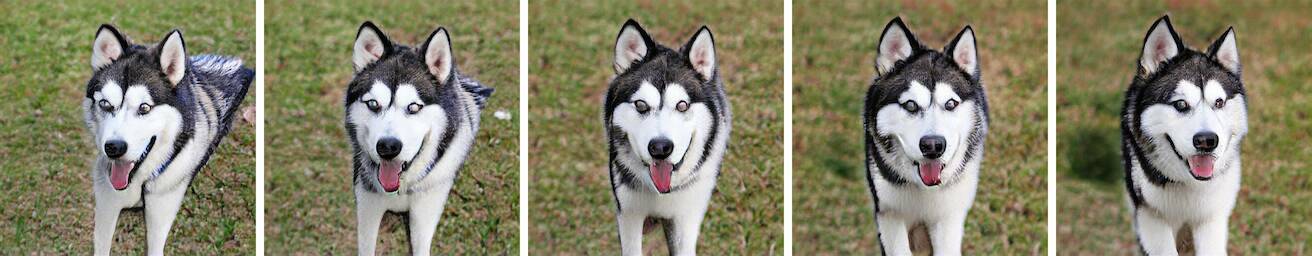}\hspace*{1cm}
\includegraphics[width=\h]{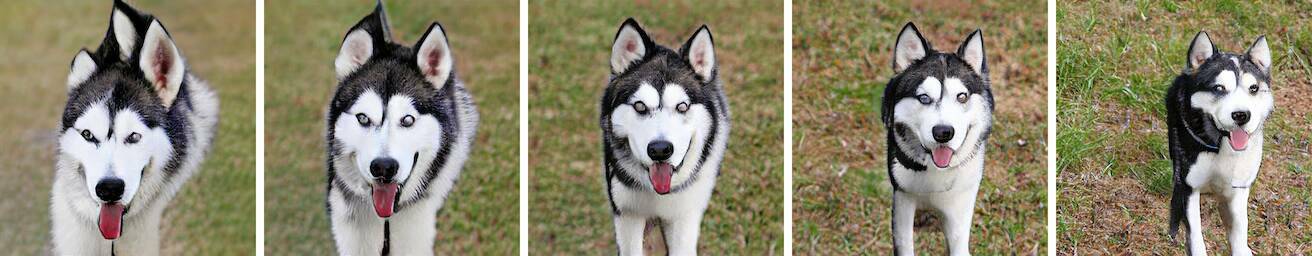}\\
\includegraphics[width=\h]{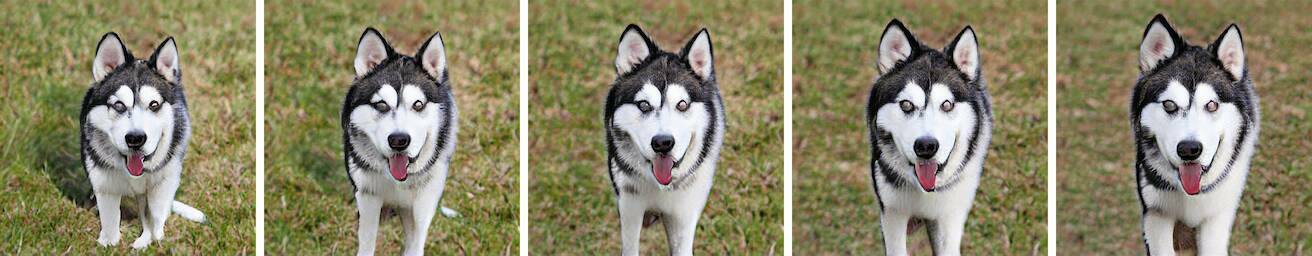}\hspace*{1cm}
\includegraphics[width=\h]{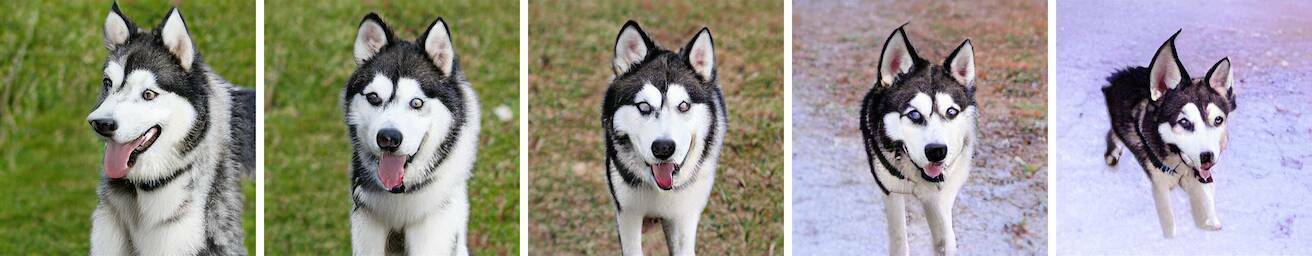}\\
\includegraphics[width=\h]{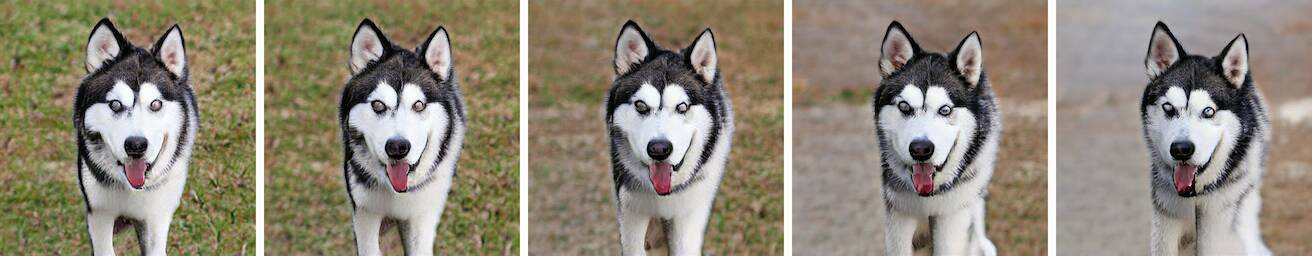}\hspace*{1cm}
\includegraphics[width=\h]{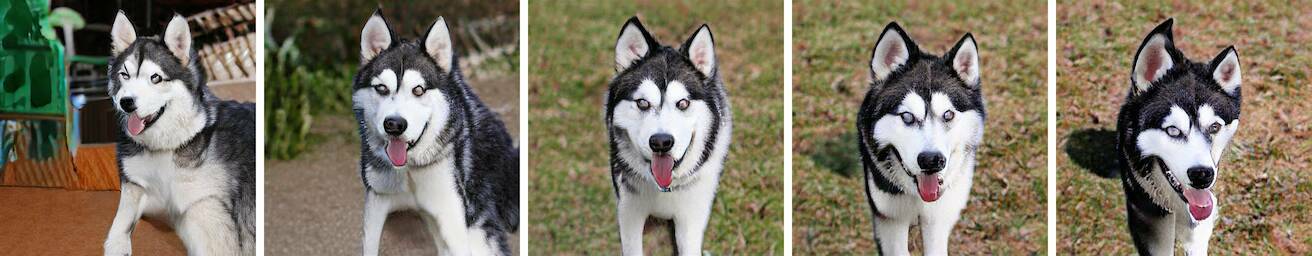}\\
\includegraphics[width=\h]{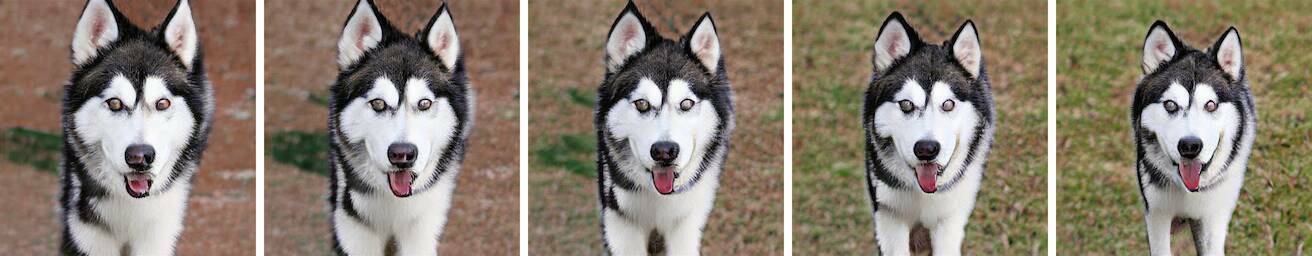}\hspace*{1cm}
\includegraphics[width=\h]{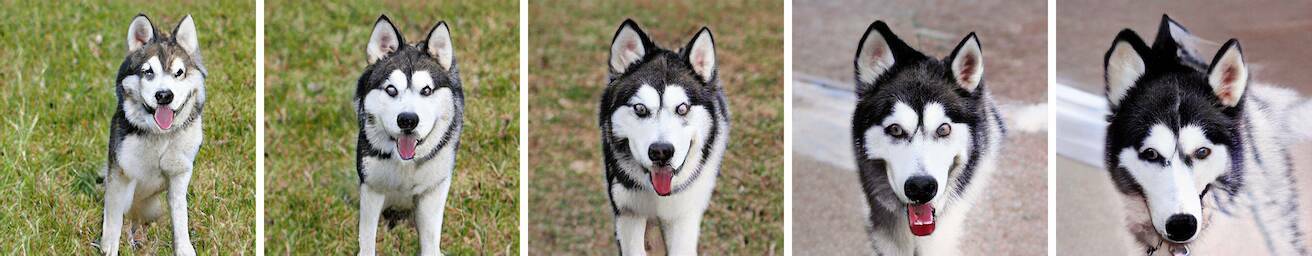}\\
\includegraphics[width=\h]{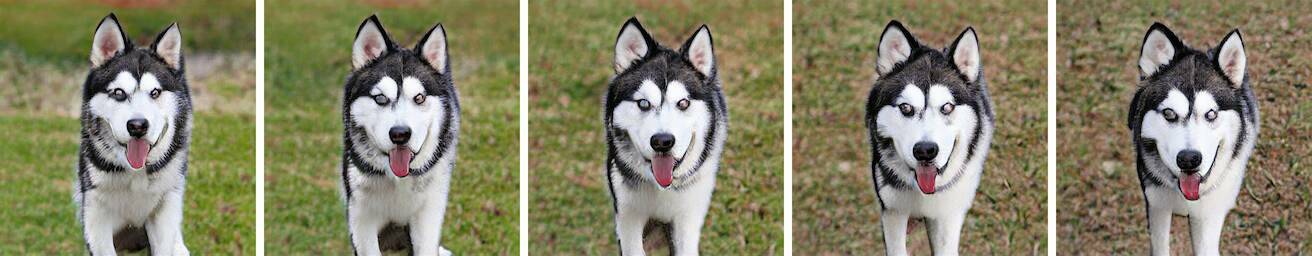}\hspace*{1cm}
\includegraphics[width=\h]{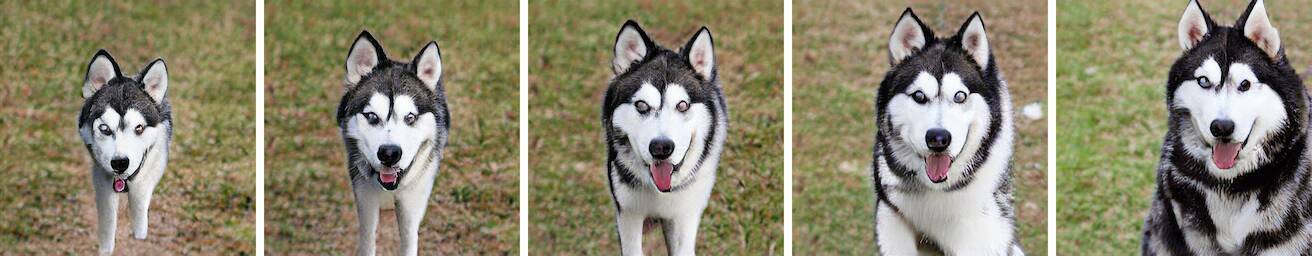}\\
\includegraphics[width=\h]{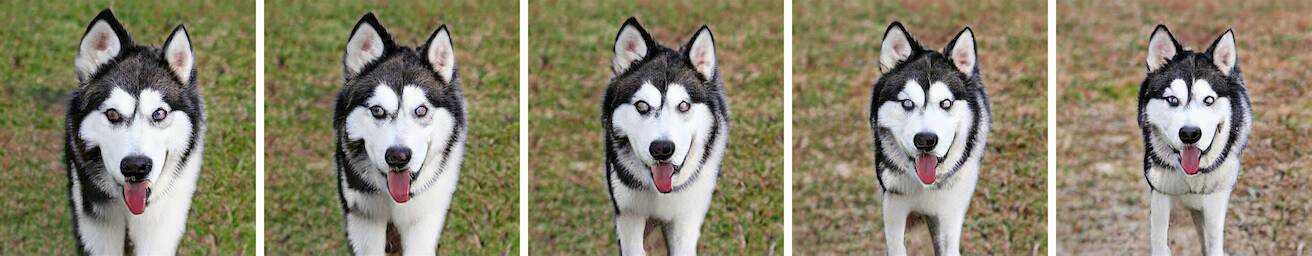}\hspace*{1cm}
\includegraphics[width=\h]{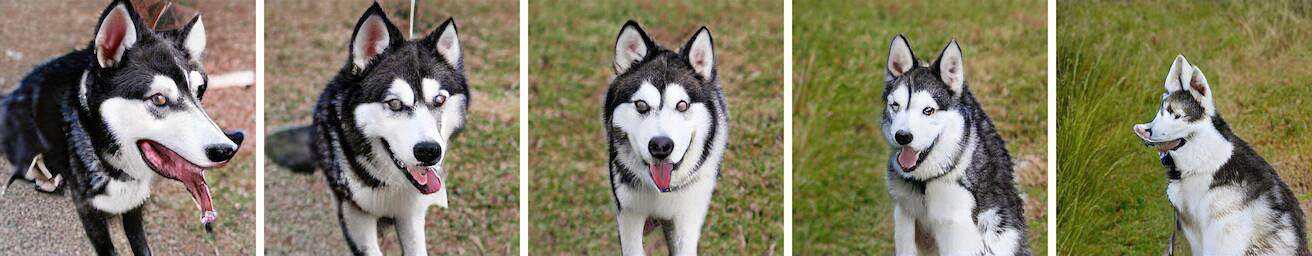}\\
\includegraphics[width=\h]{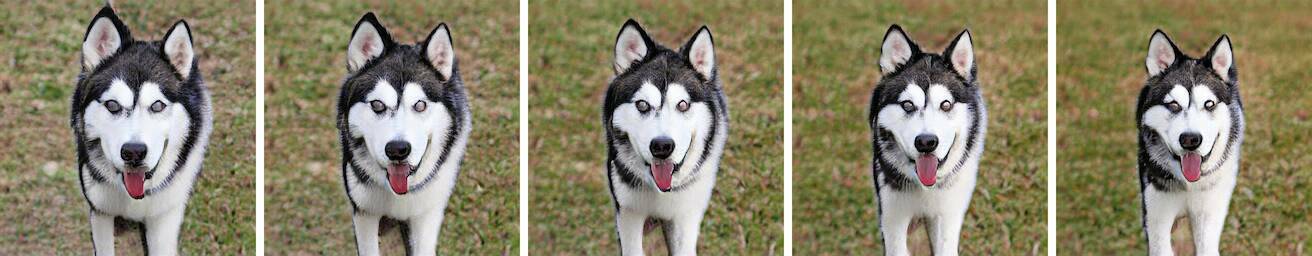}\hspace*{1cm}
\includegraphics[width=\h]{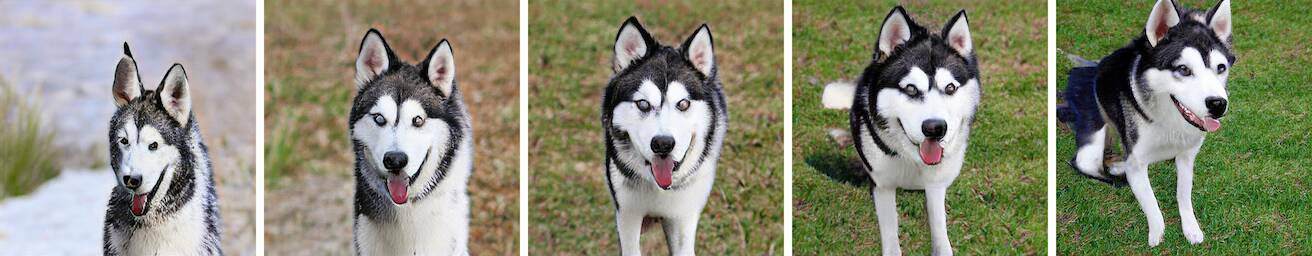}\\
\includegraphics[width=\h]{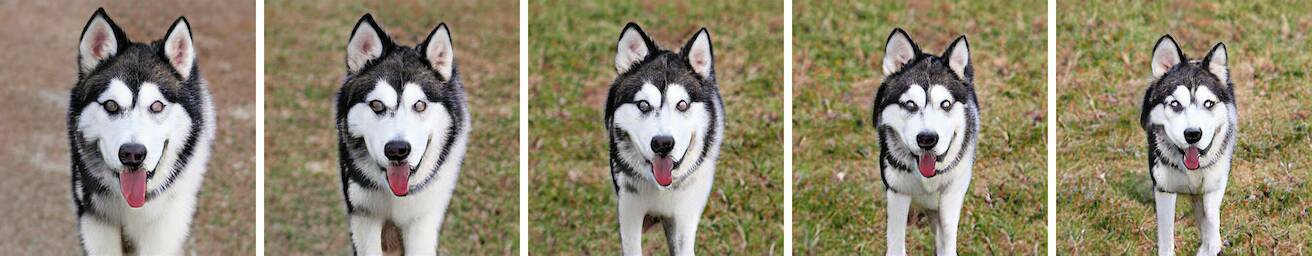}\hspace*{1cm}
\includegraphics[width=\h]{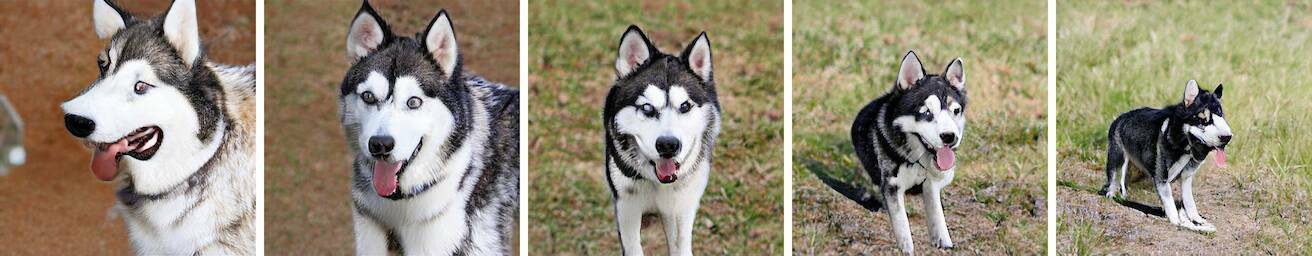}\\
\includegraphics[width=\h]{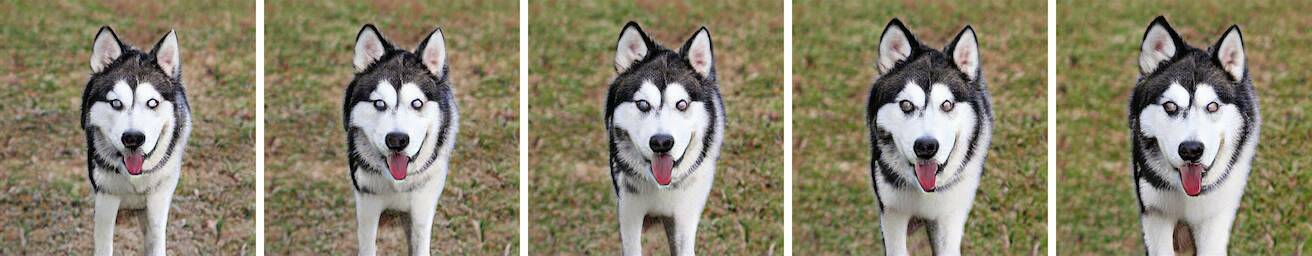}\hspace*{1cm}
\includegraphics[width=\h]{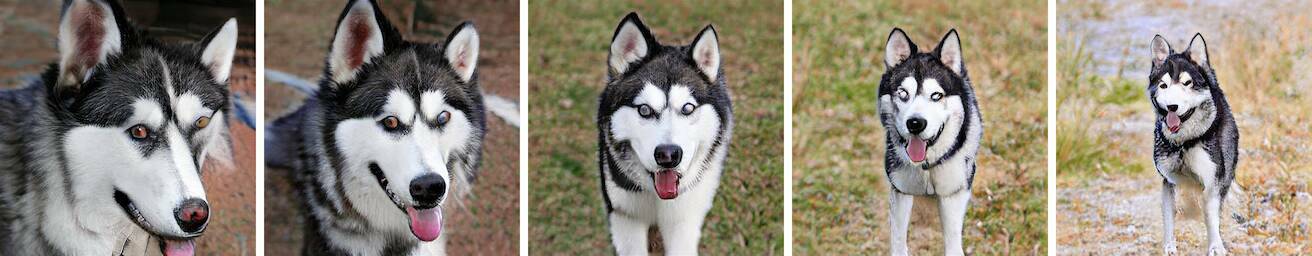}\\
\makebox[0.4\linewidth]{(a) Principal components $\mathbf{u_{0}}-\mathbf{u_{19}}$, $\pm 2\sigma$}\hspace*{1cm}
\makebox[0.4\linewidth]{(b) Normally distributed directions in $\mathcal{Z}$, $\pm 6\hat{r_i}$}%
\caption{\label{fig:topPCsHusky} A visualization of the first 20 principal components of BigGAN512-deep husky (a), and of 20 isotropic Gaussian directions in $\mathcal{Z}$ (b). The random directions are scaled to emphasize their effect.}
\end{figure*}
}

\title{GANSpace: Discovering Interpretable GAN Controls}

\begin{document}

\author{Erik H{\"a}rk{\"o}nen$^{1,2}$ \and Aaron Hertzmann$^2$ \and Jaakko Lehtinen$^{1,3}$ \and Sylvain Paris$^2$}


\hyphenation{StyleGAN}
\hyphenation{BigGAN}

\maketitle

\vspace{-1cm}
\begin{center}
\hspace{-7ex} $^1$Aalto University \hspace{2ex} $^2$Adobe Research \hspace{3ex} $^3$NVIDIA
\end{center}

\begin{abstract}
    This paper describes a simple technique to analyze Generative Adversarial Networks (GANs) and create interpretable controls for image synthesis, such as change of viewpoint, aging, lighting, and time of day.  We identify important latent directions based on Principal Component Analysis (PCA) applied either in latent space or feature space. Then, we show that a large number of interpretable controls can be defined by layer-wise perturbation along the principal directions.  Moreover, we show that BigGAN can be controlled with layer-wise inputs in a StyleGAN-like manner. We show results on different GANs trained on various datasets, and demonstrate good qualitative matches to edit directions found through earlier supervised approaches.
    


\end{abstract}


\section{Introduction}

Generative Adversarial Networks (GANs) \cite{GANs}, like BigGAN \cite{biggan} and StyleGAN \cite{stylegan,stylegan2}, are powerful  image synthesis models that  can generate a wide variety of high-quality images, and have already been adopted by digital artists \cite{aia-ganart}. 
Unfortunately, such models provide little direct control over image content, other than selecting image classes or adjusting StyleGAN's style vectors.  Current attempts to add user control over the output focus on supervised learning of latent directions \cite{gansteerability,ganalyze,hierarchyemerges,shen2019interpreting,Plumerault}, GAN training with labeled images \cite{kulkarni,drgan,finegan}. However, this requires expensive manual supervision for each new control to be learned. 
A few methods provide useful control over spatial layout of the generated image \cite{gaugan,igan,bau2019gandissect,Bau:Ganpaint:2019}, provided a user is willing to paint label or edge maps.

This paper shows how to identify new interpretable control directions for existing GANs, without requiring post hoc supervision or expensive optimization: rather than setting out to find a representation for particular concepts (``show me your representation for smile''), our exploratory approach makes it easy to browse through the concepts that the GAN has learned.  We build on two main discoveries. First, we show that important directions in GAN latent spaces can be found by applying Principal Component Analysis (PCA) in latent space for StyleGAN, and feature space for BigGAN.  Second, we show how BigGAN can be modified to allow StyleGAN-like layer-wise style mixing and control, without retraining.  Using these ideas, we show that layer-wise decomposition of PCA edit directions leads to many interpretable controls.  Identifying useful control directions then involves an optional one-time user labeling effort.

These mechanisms are algorithmically extremely simple, but lead to surprisingly powerful controls. They allow control over image attributes that vary from straightforward high-level properties such as object pose and shape, to many more-nuanced properties like lighting, facial attributes, and landscape attributes (Figure \ref{fig:Teaser}).   These directions, moreover, provide understanding about how the GAN operates, by visualizing its ``EiGANspace.'' We show results with BigGAN512-deep and many different StyleGAN and StyleGAN2 models, and demonstrate many novel types of image controls.  

\figTeaser

One approach is to attempt to train new GANs to be disentangled, e.g., \cite{ramesh}. However, training general models like BigGAN requires enormous computational resources beyond the reach of nearly all potential researchers and users.  Hence, we expect 
that research to interpret and extend the capabilities of existing GANs will become increasingly important.

\section{Discovering GAN Controls}
\label{sec:discovering}

This section describes our new techniques for augmenting existing GANs with new control variables. Our techniques are, algorithmically, very simple. This simplicity is an advantage: for very little effort, these methods enable a range of powerful tools for analysis and control of GANs, that have previously not been demonstrated, or else required expensive supervision. In this paper, we work exclusively with pretrained GANs.  

\subsection{Background}
We begin with a brief review of GAN representations \cite{GANs}. The most basic GAN comprises a probability distribution $p(\bz)$, from which a latent vector $\bz$ is sampled, and a neural network $G(\bz)$ that produces an output image $I$: $\bz \sim p(\bz)$, $I = G(\bz)$.
%
%
The network can be further decomposed into a series of $L$ intermediate layers $G_1 ... G_L$. The first layer takes the latent vector as input and produces a feature tensor $\by_1 = G_1(\bz)$ consisting of set of feature maps. The remaining layers each produce features as a function of the previous layer's output: $\by_i = \hat{G}_i(\bz) \equiv G_i\left(\by_{i-1}\right)$.
The output of the last layer $I=G_L(\by_{L-1})$ is an RGB image. In the BigGAN model \cite{biggan}, the intermediate layers also take the latent vector as input:
\begin{equation}
    \by_i = G_i(\by_{i-1}, \bz)
\end{equation}
which are called Skip-$z$ inputs.
BigGAN also uses a class vector as input.  In each of our experiments, the class vector is held fixed, so we omit it from this discussion for clarity.
In a StyleGAN model \cite{stylegan,stylegan2}, the first layer takes a constant input $\by_0$. Instead, the output is controlled by 
a non-linear function of $\bz$ as input to intermediate layers:
\begin{align}
    \by_{i} = G_i(\by_{i-1}, \bw) \qquad \text{with} \; \bw = M(\bz)
    \label{eq:normlayer}
\end{align}
where
%
$M$ is
an 8-layer multilayer perceptron. In basic usage, the vectors $\bw$ controlling the synthesis at each layer are all equal; the authors demonstrate that allowing each layer to have its own $\bw_i$ enables powerful ``style mixing,'' the combination of features of various abstraction levels across generated images.

\subsection{Principal Components and Principal Feature Directions}

How can we find useful directions in $\bz$ space?  The isotropic prior distribution $p(\bz)$ does not indicate which directions are useful. On the other hand, the distribution of outputs in the high-dimensional pixel space is extremely complex, and difficult to reason about. Our main observation is, simply, that the principal components of feature tensors on the early layers of GANs represent important factors of variation. We first describe how the principal components are computed, and then study the properties of the basis they form.

\textbf{StyleGAN.}
Our procedure is simplest for StyleGAN \cite{stylegan,stylegan2}. Our goal is to  identify the principal axes of $p(\bw)$. To do so, we sample $N$ random vectors $\bz_{1:N}$, and compute the corresponding $\bw_i=M(\bz_i)$ values.  We then compute PCA of these $\bw_{1:N}$ values. This gives a basis $\bV$ for $\mathcal{W}$.
Given a new image defined by $\bw$, we can edit it by varying PCA coordinates $\bx$ before feeding to the synthesis network:
\begin{align}
\bw' &= \bw + \bV \bx
\end{align}
where each entry $x_k$ of $\bx$ is a separate control parameter. The entries $x_k$ are initially zero until modified by a user.

\textbf{BigGAN.}
For BigGAN \cite{biggan}, the procedure is more complex, because the $\bz$ distribution is not learned, and there is no $\bw$ latent that parameterizes the output image. We instead perform PCA at an intermediate network layer $i$, and then transfer these directions back to the $\bz$ latent space, as follows. We first sample $N$ random latent vectors $\bz_{1:N}$; these are processed through the model to produce $N$ feature tensors $\by_{1:N}$ at the $i$th layer, where $\by_j = \hat{G}_i(\bz_j)$.  We then compute PCA from the $N$ feature tensors, which produces a low-rank basis matrix $\bV$, and the data mean $\bm{\mu}$.  The PCA coordinates $\bx_j$ of each feature tensor are then computed by projection: $\bx_j = \bV^T(\by_j - \bm{\mu})$.   

We then transfer this basis to latent space by linear regression, as follows.
We start with an individual basis vector $\bv_k$ (i.e., a column of $\bV$), and the corresponding PCA coordinates $x^k_{1:N}$, where $x^k_j$ is the scalar $k$-th coordinate of $\bx_j$. We solve for the corresponding latent basis vector $\bu_k$ as:
\begin{equation}
    \bu_k = \argmin \sum_j \left \| \bu_k x^k_j - \bz_j \right \|^2
\end{equation}
to identify a latent direction corresponding to this principal component (Figure \ref{fig:PCAdiagram}). Equivalently, the whole basis is computed simultaneously with
$    \bU = \argmin \sum_j \left \| \bU \bx_j - \bz_j \right \|^2 $,
using a standard least-squares solver, without any additional orthogonality constraints. Each column of $\bU$ then aligns to the variation along the corresponding column of $\bV$. We call the columns $\bu_k$ \emph{principal directions}. We use a new set of $N$ random latent vectors for the regression. Editing images proceeds similarly to the StyleGAN case, with the $x_k$ coordinates specifying offsets along the columns $\bu_k$ of the principal direction matrix: $\bz' = \bz + \bU \bx$.

\figPCADiagram



We compute PCA at the first linear layer of BigGAN512-deep, which is the first layer  with a non-isotropic distribution.  We found that this gave more useful controls than later layers. Likewise, for StyleGAN, we found that PCA in $\mathcal{W}$ gave better results than applying PCA on feature tensors and then transferring to latent space $\bw$. 

Examples of the first few principal components are shown in Figure \ref{fig:PCs}(top) for StyleGAN2 trained on FFHQ; see also the beginning of the accompanying video. While they capture important concepts, some of them entangle several separate concepts. Similar visualizations are shown for other models (in Section~1 of the Supplemental Material, abbreviated \SMref{1} later).

\figPCACleanup

\subsection{Layer-wise Edits}
\label{sec:layerwise}

Given the directions found with PCA, we now show that these can be decomposed into interpretable edits by applying them only to certain layers.

\paragraph{StyleGAN.}
StyleGAN provides layerwise control via the $\bw_i$ intermediate latent vectors. Given an image with latent vector $\bw$, layerwise edits entail modifying only the $\bw$ inputs to a range of layers, leaving the other layers' inputs unchanged.  We use notation \editv{i}{j-k} to denote edit directions; for example, \editv{1}{0-3} means moving along component $\bv_1$ at the first four layers only.
\editv{2}{all} means moving along component $\bv_2$ globally: in the latent space and to all layer inputs. Edits in the $\mathcal{Z}$ latent space are denoted \editu{i}{j-k}.

This is illustrated in the last rows of Figure~\ref{fig:PCs}. For example, component $\bv_1$, which controls head rotation and gender in an entangled manner, controls a purer rotation when only applied to the first three layers in \editv{1}{0-2}; similarly, the age and hairstyle changes associated with component $\bv_4$ can be removed to yield a cleaner change of lighting by restricting the effect to later layers in \editv{4}{5-17}. It is generally easy to discover surprisingly targeted changes from the later principal components. Examples include \editv{10}{7-8} that controls hair color, as well as \editv{11}{0-4} that controls the height of the hair above the forehead. More examples across several models are shown in Figure~\ref{fig:EditZoo}; see also the accompanying video.
As shown in Figure~\ref{fig:Teaser}, multiple edits applied simultaneously across multiple principal directions and internal layer ranges compose well.

\paragraph{BigGAN.}
BigGAN does not have a built-in layerwise control mechanism.
However, we find that \textbf{BigGAN can be modified to produce behavior similar to StyleGAN}, by varying the intermediate Skip-$z$ inputs $\bz_i$ separately from the latent $\bz$: $\by_{i} = G(\by_{i-1}, \bz_i)$.
Here the latent inputs $\bz_i$ are allowed to vary individually between layers in a direct analogy to the style mixing of StyleGAN. By default, all inputs are determined by an initial sampled or estimated $\bz$, but then edits may be performed to the inputs to different layers independently. Despite the fact that BigGAN is trained without style mixing regularization, we find that it still models images in a form of style/content hierarchy. Figure~\ref{fig:BigGANStyles} shows the effect of transferring intermediate latent vectors from one image to another. Like StyleGAN, transferring at lower layers (closer to the output) yields lower-level style edits.  See \SMref{2} 
for more examples of BigGAN style mixing.
Since the Skip-$z$ connections were not trained for style resampling, we find them to be subjectively ``more entangled'' than the StyleGAN style vectors.  However, they are still useful for layerwise editing, as shown in Figures \ref{fig:EditZoo} and \SMref{1}: 
we discover components that control, for instance, lushness of foliage, illumination and time of day, and cloudiness, when applied to a select range of layers. 

\paragraph{Interface.}
We have created a simple user interface that enables interactive exploration of the principal directions via simple slider controls. 
Layer-wise application is enabled by specifying a start and end layer for which the edits are to be applied. The GUI also enables the user to name the discovered directions, as well as load and save sets of directions. 
The exploration process is demonstrated in the video, and the runnable Python code is attached as supplemental material. 


\section{Findings and Results}

We describe a number of discoveries from our PCA analysis, some of which we believe are rather surprising. We also show baseline comparisons.  We show edits discovered on state-of-the-art pretrained GANs, including BigGAN512-deep, StyleGAN (Bedrooms, Landscapes, WikiArt training sets), and StyleGAN2 (FFHQ, Cars, Cats, Church, Horse training sets). Details of the computation and the pretrained model sources are found in \SMref{3}.  This analysis reveals properties underlying the StyleGAN and BigGAN models.

\subsection{GAN and PCA Properties}

Across all trained models we have explored, \textbf{large-scale changes to geometric configuration and viewpoint are limited to the first 20 principal components ($\bv_0$-$\bv_{20}$); successive components leave layout unchanged, and instead control object appearance/background and details}. As an example, Figure \ref{fig:PCs} shows edit directions for the top 3 PCA components in a StyleGAN2 model trained on the FFHQ face dataset \cite{stylegan}. We observe that the first few components control large-scale variations, including apparent gender expression and head rotation. For example, component $\bv_0$ is a relatively disentangled gender control; component $\bv_1$ mixes head rotation and gender, and so on. See \SMref{1} for a visualization of the first 20 principal components.

\figRandomBaseline




PCA also reveals that \textbf{StyleGANv2's latent distribution $p(\bw)$ has a relatively simple structure:} the principal coordinates are nearly-independent variables with non-Gaussian unimodal distributions. 
We also find that \textbf{the first 100 principal components are sufficient to describe overall image appearance}; the remaining 412 dimensions control subtle though perceptible changes in appearance; see \SMref{4} and \SMref{5}
for details and examples.  

We find that \textbf{BigGAN components appear to be class-independent}, e.g., PCA components for one class were identical to PCA components for another class in the cases we tested. \SMref{6}
shows examples of PCA components computed at the first linear layer of BigGAN512-deep for the husky class. We find that the global motion components have the same effect in different classes (e.g., component 6 is zoom for all classes tested), but later components may have differing interpretations across classes. For instance, a direction that makes the image more blue might mean winter for some classes, but just nighttime for others.

\subsection{Model entanglements and disallowed combinations}
\label{sec:entanglements}

We observe a number of properties of GAN principal components that seem to be inherited from GANs' training sets. In some cases, these properties may be desirable, and some may be limitations of our approach. Some of these may also be seen as undesirable biases of the trained GAN. Our analysis provides one way to identify these properties and biases that would otherwise be hard to find.

For StyleGAN2 trained on the FFHQ face dataset, geometric changes are limited to rotations in the first 3 components. No translations are discovered, due to the carefully aligned training set.

Even with our layer-wise edits, we observe some entanglements between distinct concepts. For example, adjusting a car to be more ``sporty'' causes a more ``open road'' background, whereas a more ``family'' car appears in woods or city streets. This plausibly reflects typical backgrounds in marketing photographs of cars. Rotating a dog often causes its mouth to open, perhaps a product of correlations in portraits of dogs. 
For the ``gender'' edit, one extreme ``male'' side  seems to place the subject in front of a microphone; whereas the ``female'' side is a more frontal portrait.
 See \SMref{7} for examples. 

We also observe ``disallowed combinations,'' attributes that the model will not apply to certain faces. The ``Wrinkles'' edit will age and add wrinkles to adult faces, but has no significant effect on a child's face. Makeup and Lipstick edits add/remove makeup to female-presenting faces, but have little or no effect on male faces. When combining the two edits for ``masculine" and ``adult," all combinations work, except for when trying to make a ``masculine child."  
See \SMref{7} 
for Figures.






\subsection{Comparisons}
\figSteerabilityComp

No previously published work addresses the problem we consider, namely, unsupervised identification of interpretable directions in an existing GAN. In order to demonstrate the benefits of our approach, we show qualitative comparisons to random directions and supervised methods.

\paragraph{Random directions.}
We first compare the PCA directions to randomly-selected directions in $\mathcal{W}$. Note that there are no intrinsically-preferred directions in this space, i.e., since $\bz$ is isotropic, the canonical directions in $\bz$ are equivalent to random directions. As discussed in the previous section, PCA provides a useful ordering of directions, separating the pose and the most significant appearance into the first components. As illustrated in \SMref{8}, 
each random direction includes some mixture of pose and appearance, with no separation among them.

We further illustrate this point by randomizing different subsets of principal coordinates versus random coordinates. Figure~\ref{fig:RandomBaseline}  contains four quadrants, each of which shows random perturbations about a latent vector that is shared for the entire figure. In Figure~\ref{fig:RandomBaseline}a, the first eight principal coordinates $x_{0\hdots7}$ are fixed and the remaining 504 coordinates $x_{8\hdots512}$ are randomized. This yields images where the cat pose and camera angle are held roughly constant, but the appearance of the cat and the background vary. Conversely, fixing the last 504 coordinates and randomizing the first eight (Figure~\ref{fig:RandomBaseline}b) yields images where the color and appearance are held roughly constant, but the camera and orientation vary. The bottom row shows the results of the same process applied to random directions; illustrating that any given 8 directions have no distinctive effect on the output. \SMref{8} contains more examples. 

\paragraph{Supervised methods}
Previous methods for finding interpretable directions in GAN latent spaces require outside supervision, such as labeled training images or pretrained classifiers, whereas our approach aims to automatically identify  variations intrinsic to the model without supervision.


In Figure 
 \ref{fig:SteerabilityComparison}, 
we compare some of our BigGAN zoom and translation edits to comparable  edits found by supervised methods \cite{gansteerability}, and our StyleGAN face attribute edits to a supervised method \cite{shen2019interpreting}.
 In our results, we observe a tendency for slightly more entanglement (for example, loss of microphone and hair in Figure~\ref{fig:SteerabilityComparison}d); moreover, variations of similar effects can often be obtained using multiple components. More examples from different latent vectors are shown in \SMref{8}.
 However, we emphasize that (a) our method obtained these results without any supervision, and (b) we have been able to identify many edits that have not previously been demonstrated; supervising each of these would be very costly, and, moreover, it would be hard to know in advance which edits are even possible with these GANs.



\figStyleResampling

\figEditZoo







 \section{Discussion}

This paper demonstrates simple but powerful ways to create images with existing GANs.  Rather than training a new model for each task, we take existing general-purpose image representations and discover techniques for controlling them.  
This work suggests considerable future opportunity to analyze these image representations and 
discover richer control techniques in these spaces, for example, using other unsupervised methods besides PCA. Our early experiments with performing PCA on other arrangements of the feature maps were promising.
A number of our observations 
suggest improvements to GAN architectures and training, perhaps similar to \cite{defreitas}. It would be interesting to compare PCA directions to those learned by concurrent work in  disentanglement, e.g.,  
\cite{Info-StyleGAN}.
Our approach also suggests  ideas for supervised training of edits, such as using our representation to narrow the search space. 
Several methods developed concurrently to our own explore similar or related ideas \cite{voynov,peebles2020hessian,wolff,shen2020closedform,abdal2020styleflow}, and comparing or combining approaches may prove useful as well.


\section*{Broader Impact}

As our method is an image synthesis tool, it shares with other image synthesis tools the same potential benefits (e.g., \cite{aia-ganart}) and dangers that have been discussed extensively elsewhere, e.g., see \cite{Rothman} for one such discussion.  

Our method does not perform any training on images; it takes an existing GAN as input.
As discussed in Section \ref{sec:entanglements}, our method inherits the biases of the input GAN, e.g., limited ability to place makeup on male-presenting faces. Conversely, this method provides a tool for discovering biases that would otherwise be hard to identify. 

\begin{ack}

We thank Miika Aittala for insightful discussions and  Tuomas Kynk{\"a}{\"a}nniemi for help in preparing the comparison to Jahanian et~al. \cite{gansteerability}. Thanks to Joel Simon for providing the Artbreeder Landscape model. This work was created using computational resources provided by the Aalto Science-IT project. 
\end{ack}

\bibliographystyle{abbrv}
\bibliography{gancontrol}

\newpage
\begin{center}
\textbf{\LARGE Supplementary Material}
\end{center}
\setcounter{equation}{0}
\setcounter{figure}{0}
\setcounter{table}{0}
\setcounter{page}{1}
\setcounter{section}{0}
\makeatletter
\renewcommand{\theequation}{S\arabic{equation}}
\renewcommand{\thefigure}{S\arabic{figure}}

\section{Examples of Principal Components and Layerwise Edits}
\label{sm:sec1}

Figure~\ref{fig:EditZooSupp} shows an assortment of interpretable edits discovered with our method for many different models.

We visualize the first 20 Principal Components for several models:StyleGAN2 FFHQ (Figure~\ref{fig:topPCsFFHQ}a), StyleGAN2 Cars (Figure~\ref{fig:topPCsCars}a), StyleGAN2 Cats (Figure~\ref{fig:topPCsCats}a), and BigGAN512-deep Husky (Figure~\ref{fig:topPCsHusky}a). The images are centered at the mean of each component, which causes slight differences within the center columns.

\figEditZooSupplemental

\figTopPCsSGtwoFFHQ
\figTopPCsSGtwoCats
\figTopPCsSGtwoCars
\figTopPCsBGHusky

\section{BigGAN style mixing}
\label{sm:sec2}

Figure~\ref{fig:BigganStyleMixing} shows a more detailed example of mixing style and content at different layers in BigGAN \cite{biggan}.

\figStyleMixing

\section{Model and Computation Details}
\label{sm:sec3}

We use incremental PCA \cite{RossIPCA} for efficient computation, and use $N=10^6$ samples. On a relatively high-end desktop PC, computation takes around 1.5 hours on BigGAN512-deep and 2 minutes on StyleGAN and StyleGAN2.

Our StyleGAN model weights were obtained from \url{https://github.com/justinpinkney/awesome-pretrained-stylegan}, except for Landscapes, which was provided by \url{artbreeder.com}. Our StyleGAN2 models were those provided by the authors online \cite{stylegan2}.

%

The sliders in our GUI operate in units of standard deviations, and we find that later components work for wider ranges of values than earlier ones. The first ten or so principal components, such as head rotation (\editv{1}{0-2}) and lightness/background (\editv{8}{5}), operate well in the range $[-2...2]$, beyond which the image becomes unrealistic.  In contrast, face roundness (\editv{37}{0-4}) can work well in the range $[-20...20]$, when using $0.7$ as the truncation parameter.

For truncation, we use interpolation to the mean as in StyleGAN \cite{stylegan}. The variation in slider ranges described above suggests that truncation by restricting $\bw$ to lie within 2 standard deviations of the mean would be a very conservative limitation on the expressivity of the interface, since it can produce interesting images outside this range.

A video showcasing our exploration UI is available at \url{https://youtu.be/jdTICDa_eAI}. The code of our method is hosted online: \url{https://github.com/harskish/ganspace}.

\section{How many components are needed?}
\label{sm:sec4}

\begin{figure}[h]
    \centering
    \includegraphics[width=3in]{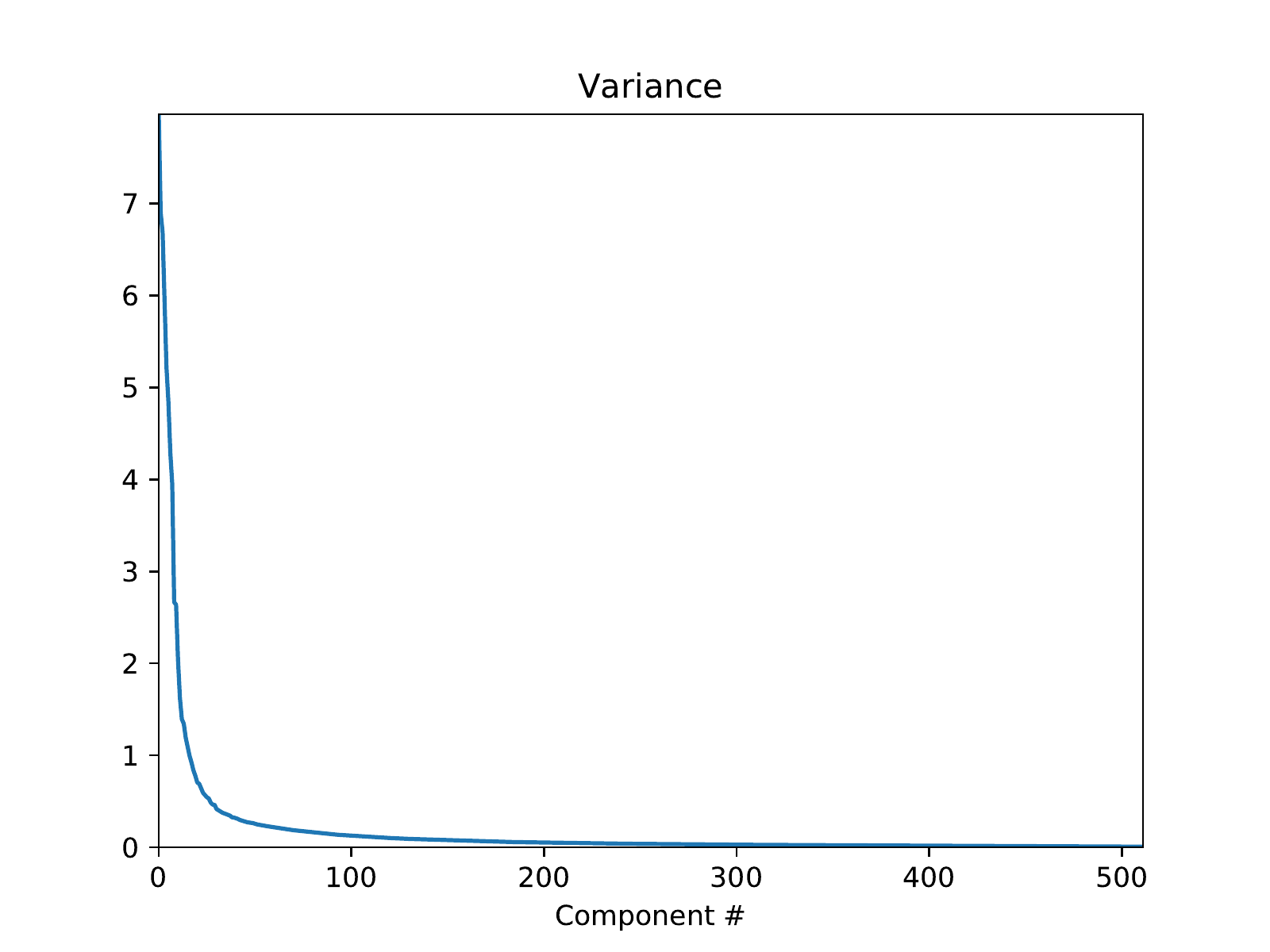}
    \caption{Variance of the principal components for StyleGANv2 FFHQ.}
    \label{fig:stats}
\end{figure}

We first investigate how many dimensions of the latent space are important to image synthesis. Figure \ref{fig:stats} shows the variance captured in each dimension of the PCA for the FFHQ model.  The first 100 dimensions capture 85\% of the varaince; the first 200 dimensions capture 92.5\%, and the first 400 dimensions capture 98.5\%.

What does this correspond to visually?
Figure \ref{fig:reduced_images} shows images randomly sampled, and then projected to a reduced set of PCA components. That is, we sample $\bw \sim p(\bw)$, and then replace it with $\bw \leftarrow \bV_K \bV^T_K (\bw - \mathbf{\mu}) + \mathbf{\mu}$, where $\bV_K$ are the columns for the first $K$ principal components.
Observe that nearly all overall face details are captured by the first 100 components; the remaining 412 components make small adjustments to shape and tone.

\figPCAReduced

\section{What is $p(\bw)$?}
\label{sm:sec5}

Inspecting the marginal distributions of the principal coordinates gives insight as to the shape of $p(\bw)$, the distribution over latents . In principle, the learned distribution could have any shape, within the range of what can be parameterized by an 8-layer fully-connected network $M(\bz)$.  For example, it could be highly multimodal, with different modes for different clusters of training image.  One could imagine, for example, different clusters for discrete properties like eyeglasses/no-eyeglasses, or the non-uniform distribution of other attributes in the training data.

In fact, we find that this is not the case: for all of the StyleGANv2 models, PCA analysis reveals that $p(\bw)$ has a rather simple form. Through this analysis, we can describe the shape of $p(\bw)$ very thoroughly.  The conclusions we describe here could be used in the future to reduce the dimensionality of StyleGAN models, either during training or as a post-process.

\paragraph{Sampling.}
To perform this analysis, we sample $N=10^6$ new samples $\bw_i \sim p(\bw)$, and then project them with our estimated PCA basis:
\begin{equation}
    \bx_i= \bV^T(\bw_i - \mathbf{\mu})
\end{equation}
where $\bV$ is a full-rank PCA matrix $(512\times512)$ for our StyleGAN2 models. We then analyze the empirical distribution of these $\bx$ samples.

The experiments described here are for the FFHQ face model, but we have observed similar phenomena for other models.

\paragraph{Independence.}
PCA projection decorrelates variables, but does not guarantee independence; it may not even be possible to obtain linear independent components for the distribution.

Let $x^{i}$ and $x^{j}$ be two entries of the $\bx$ variable. We can assess their independence by computing Mutual Information (MI) between these variables. We compute MI numerically, using a $1000\times1000$-bin histogram of the joint distribution $p(x^{(j)},x^{(k)})$. The MI of this distribution is denoted $I_{jk}$. Note that the MI of a variable with itself is equal to the entropy of that variable $H_j = I_{jj}$, and both quantities are measured in bits. We find that the entropies lie in the range $H_j \in [6.9,8.7]$ bits. In contrast, the MIs lie in the range $I_{jk} \in [0,0.3]$ bits.  

This indicates that, empirically, the principal components are very nearly independent, and we can understand the distribution by studying the individual components separately.

\paragraph{Individual distributions.}
What do the individual distributions $p(x^{j})$ look like? Figure \ref{fig:histograms} shows example histograms of these variables. As visible in the plots, the histograms are remarkably unimodal, without heavy tails. Visually they all appear Gaussian, though plotting them in the log domain reveals some asymmetries.

\newcommand{\histwidth}{1.3in}
\begin{figure}
    \centering
    \includegraphics[width=\histwidth]{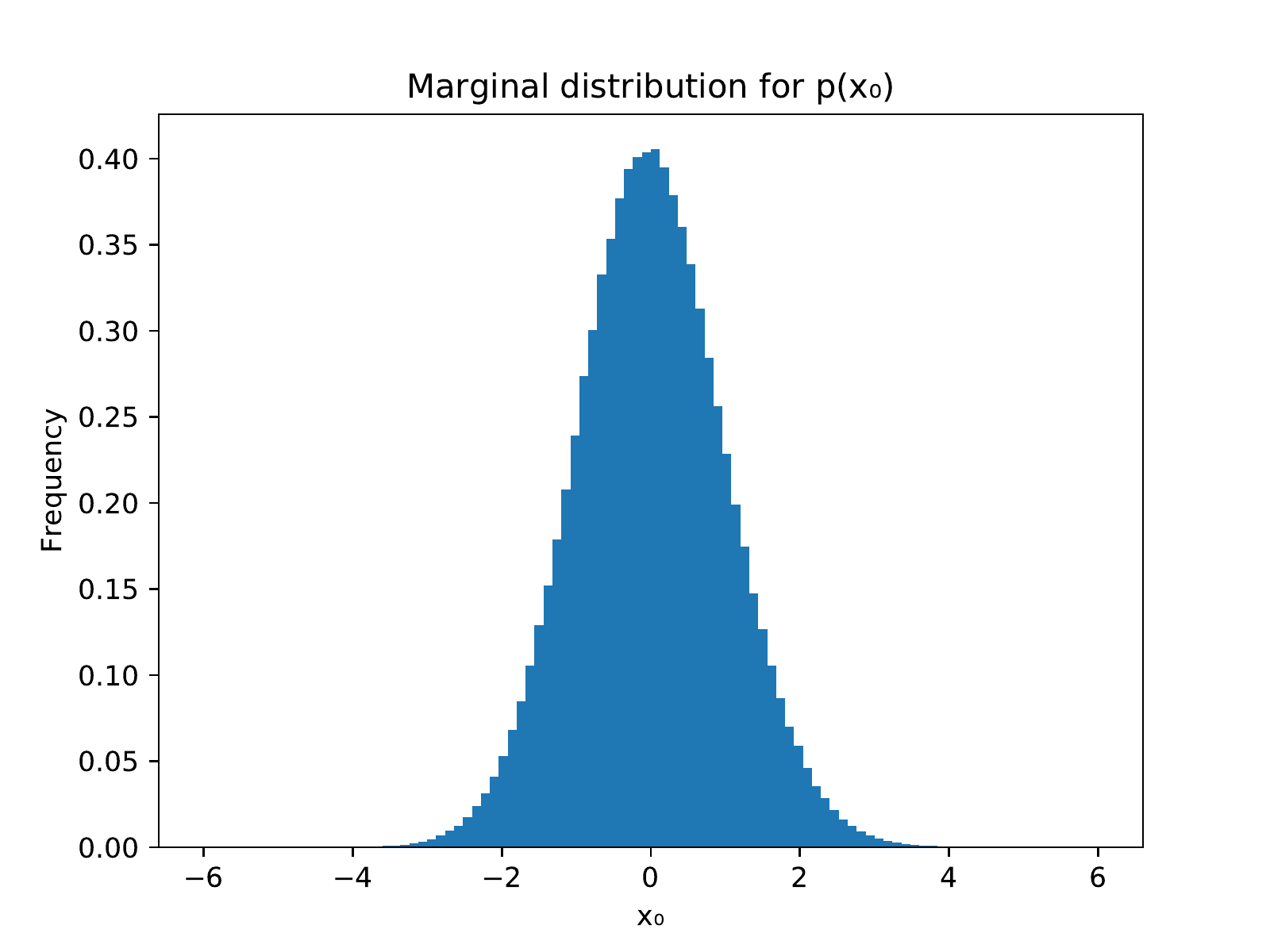}\hfill%
    \includegraphics[width=\histwidth]{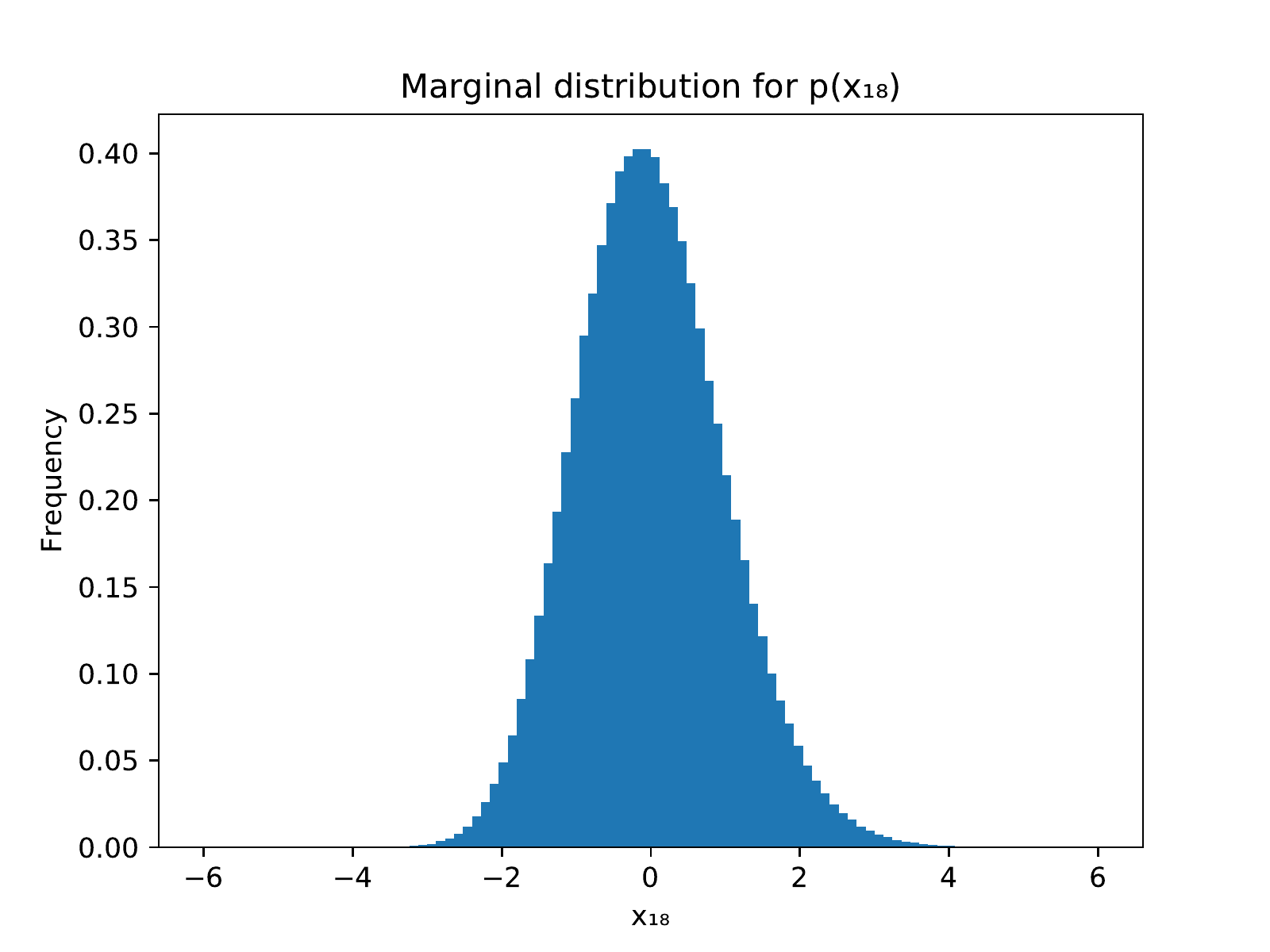}\hfill%
    \includegraphics[width=\histwidth]{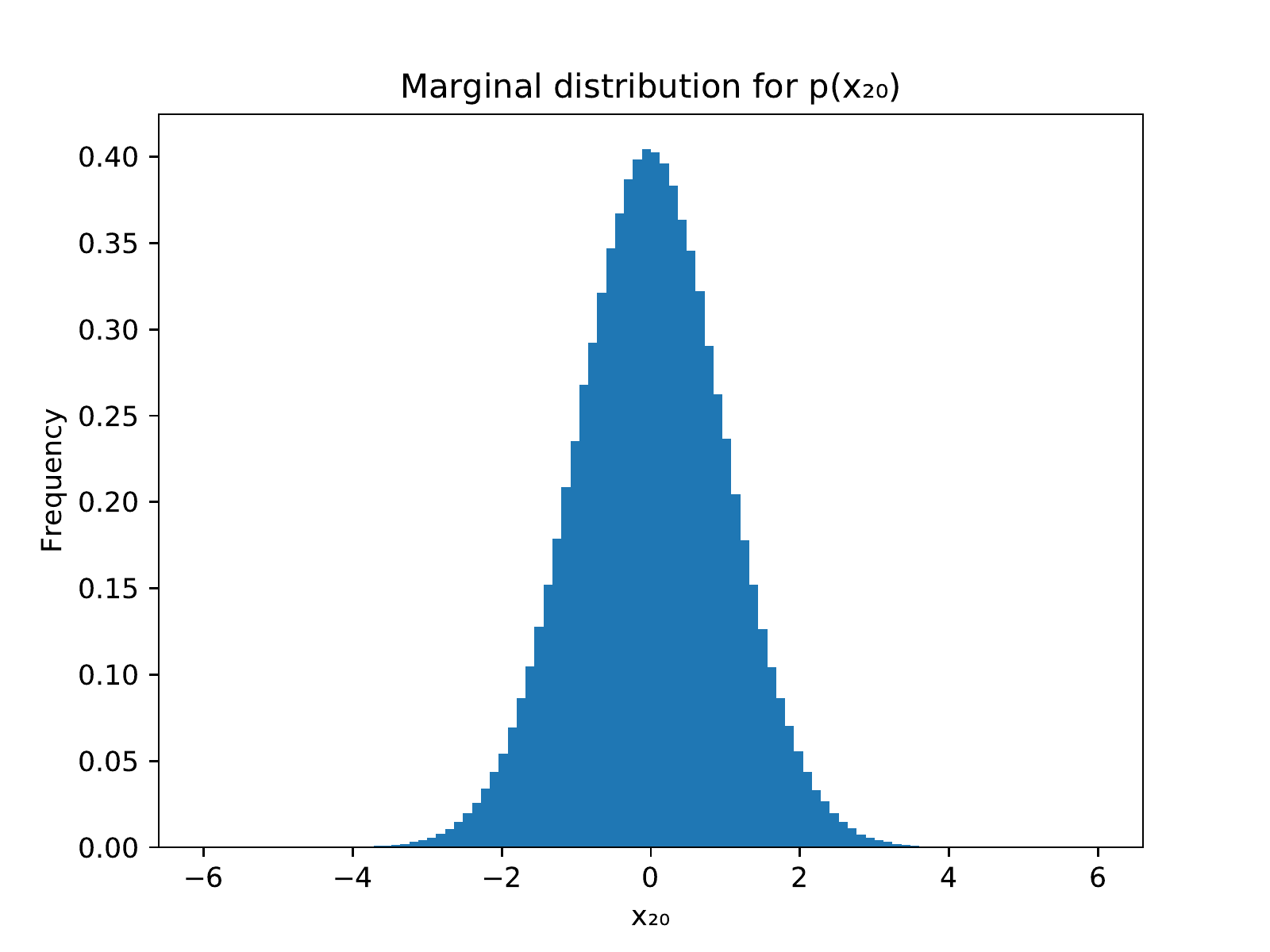}\hfill%
    \includegraphics[width=\histwidth]{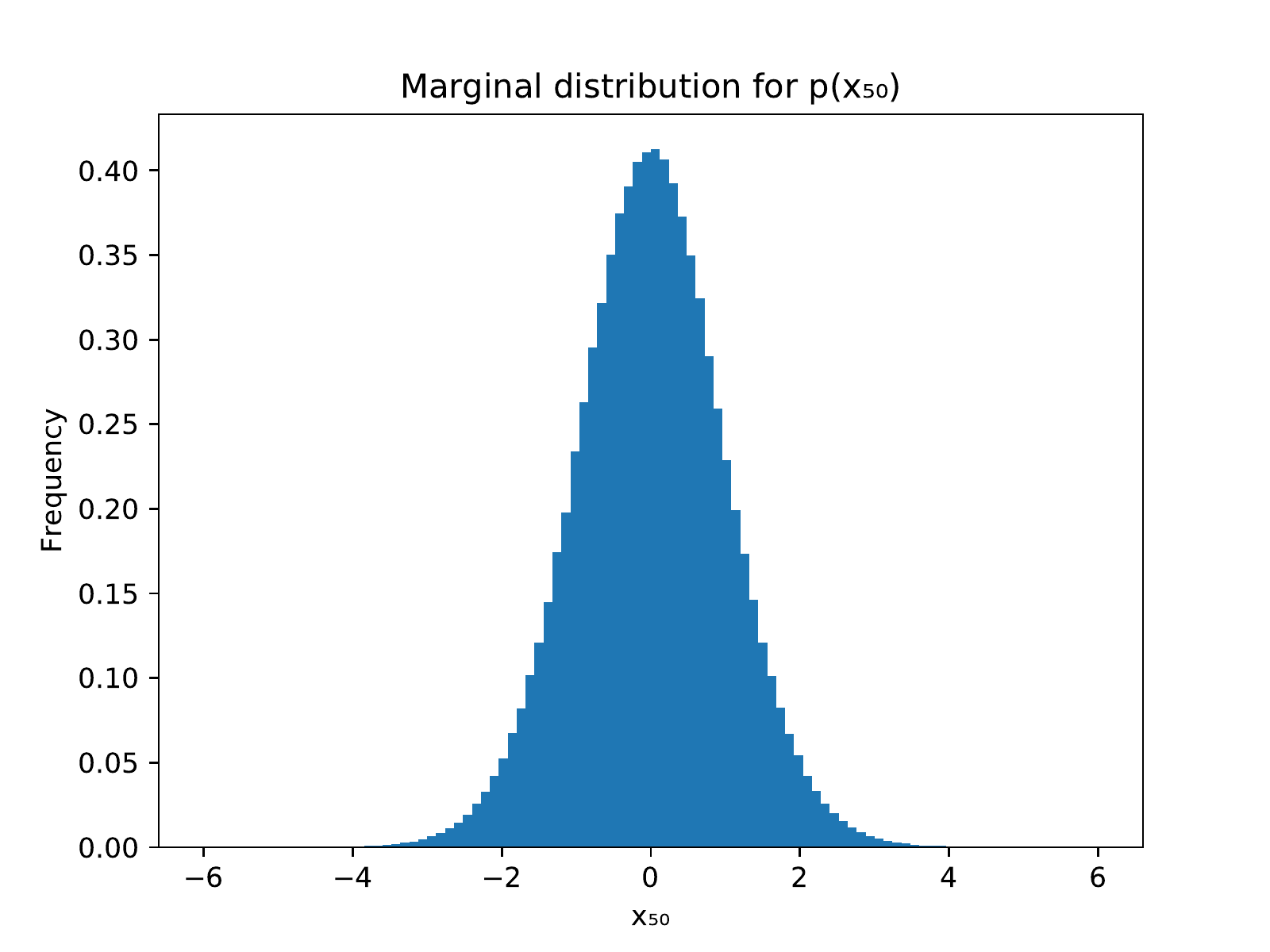} \\
    \includegraphics[width=\histwidth]{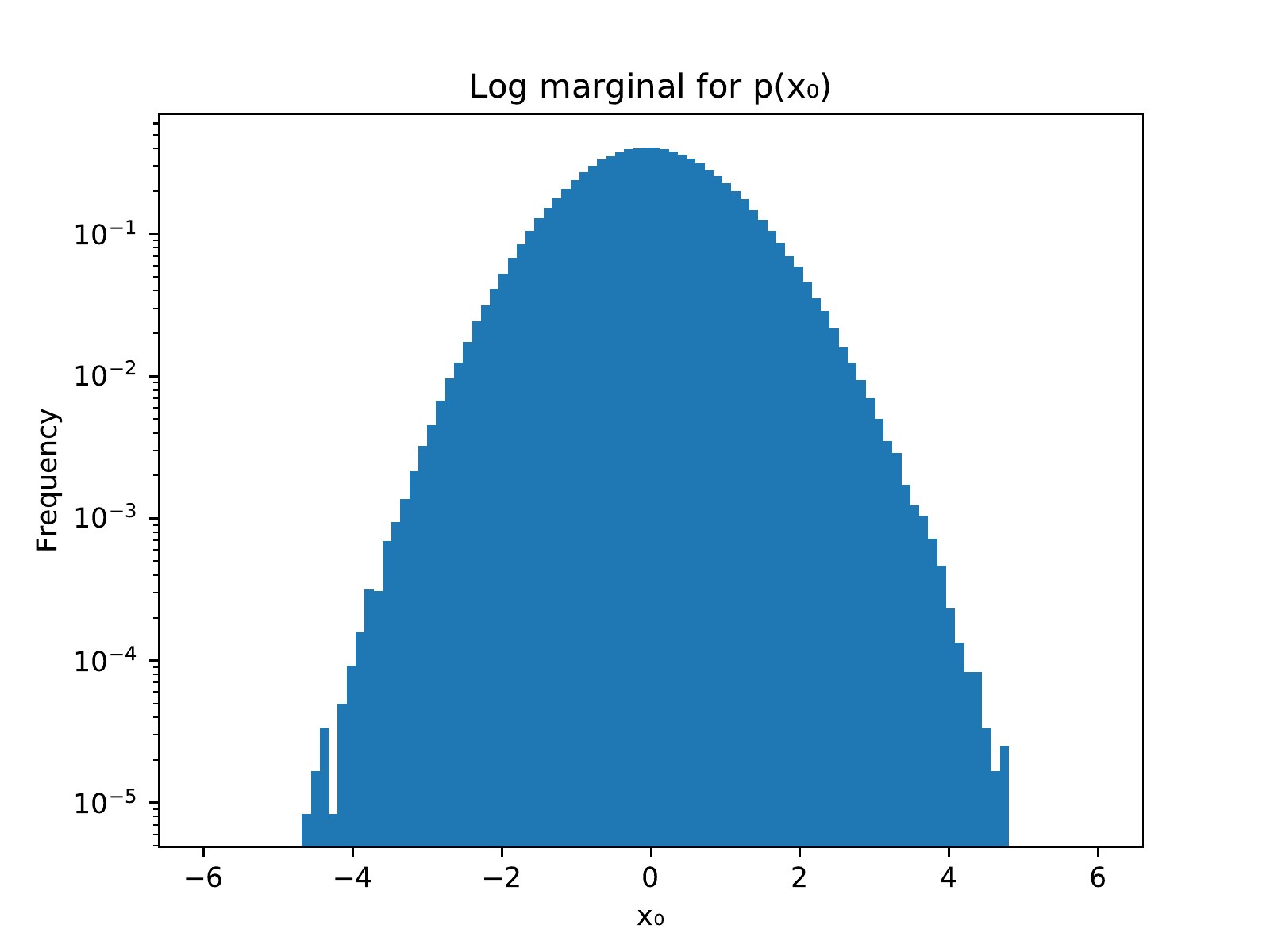}\hfill%
    \includegraphics[width=\histwidth]{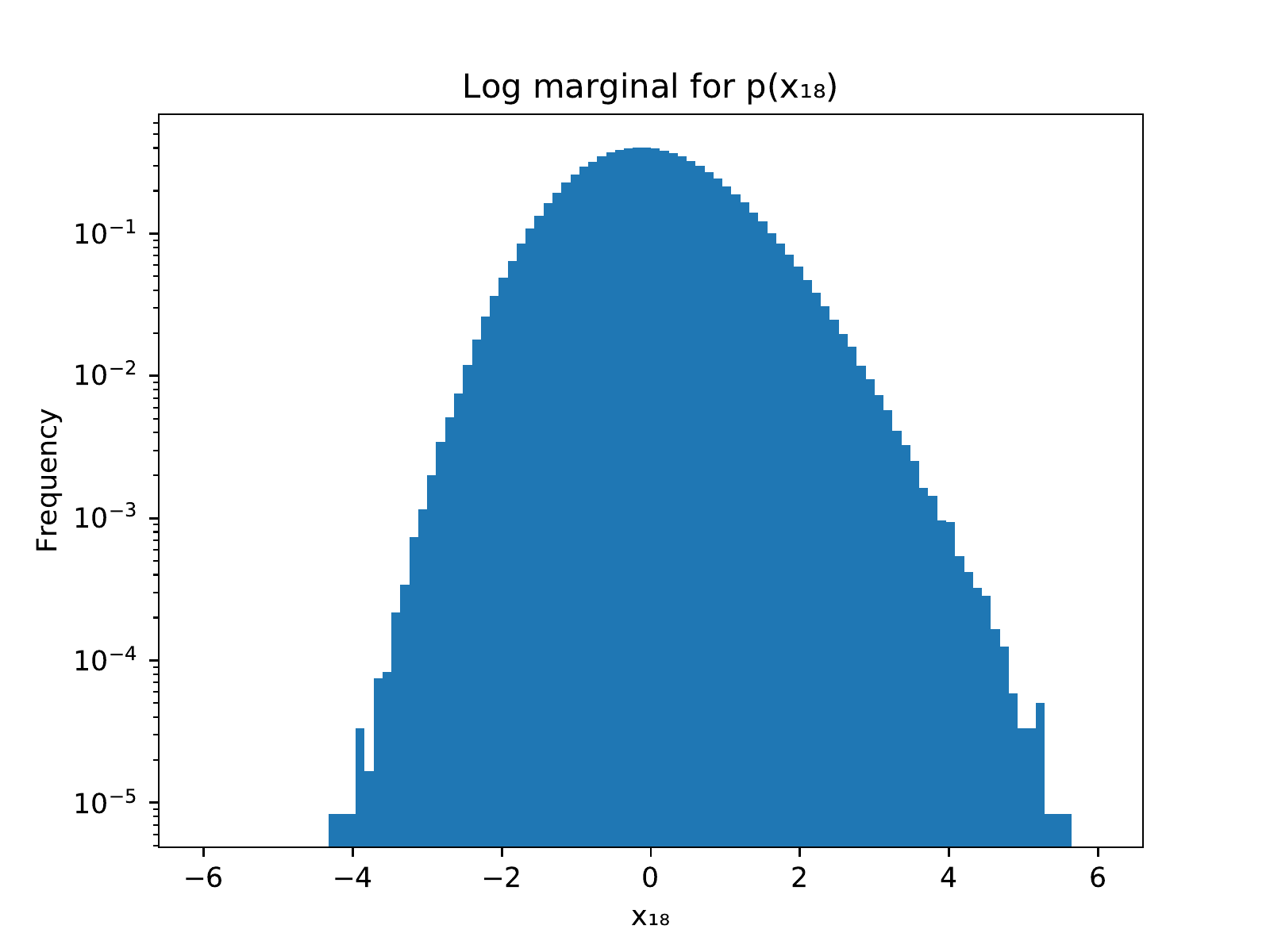}\hfill%
    \includegraphics[width=\histwidth]{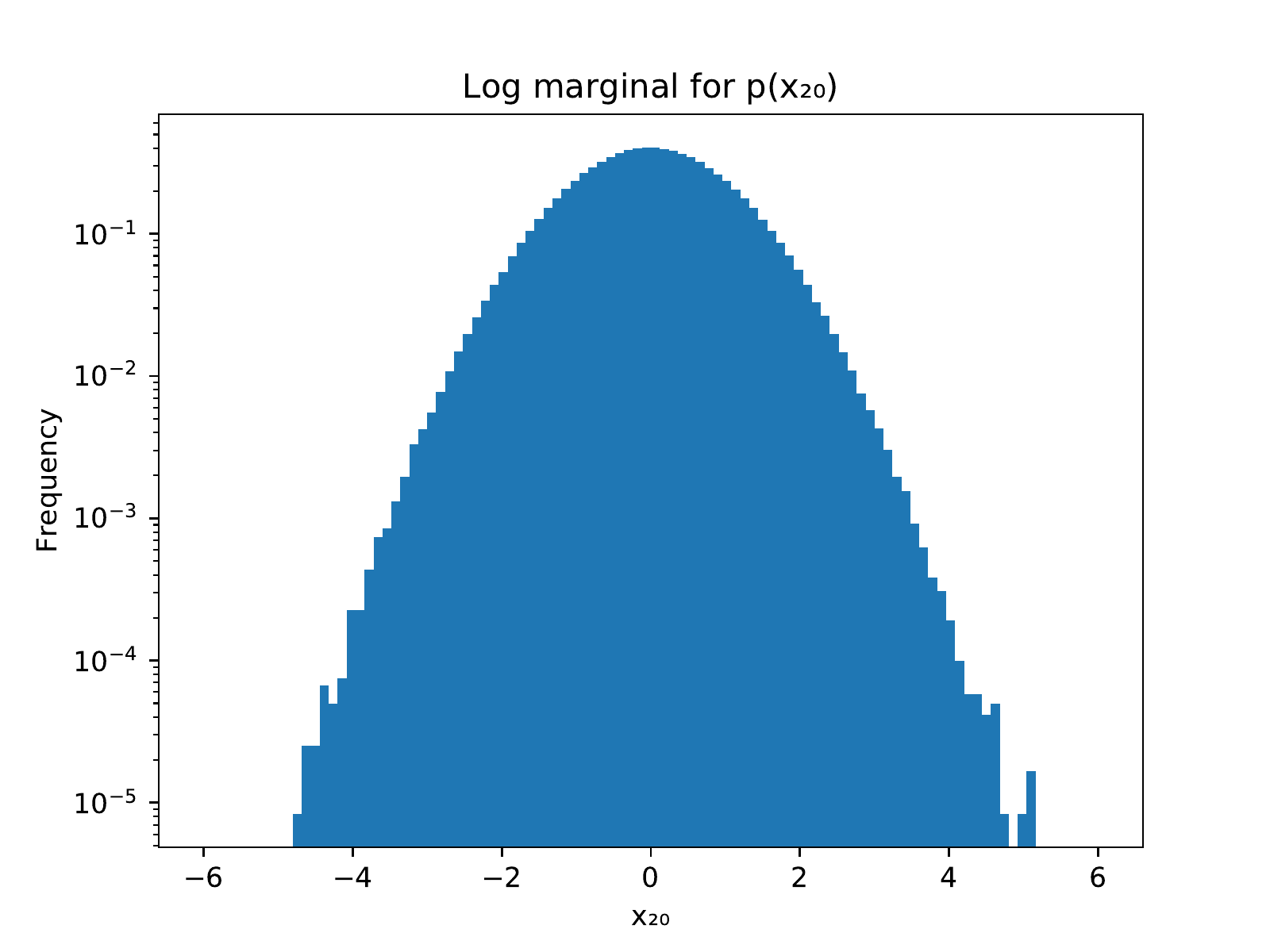}\hfill%
    \includegraphics[width=\histwidth]{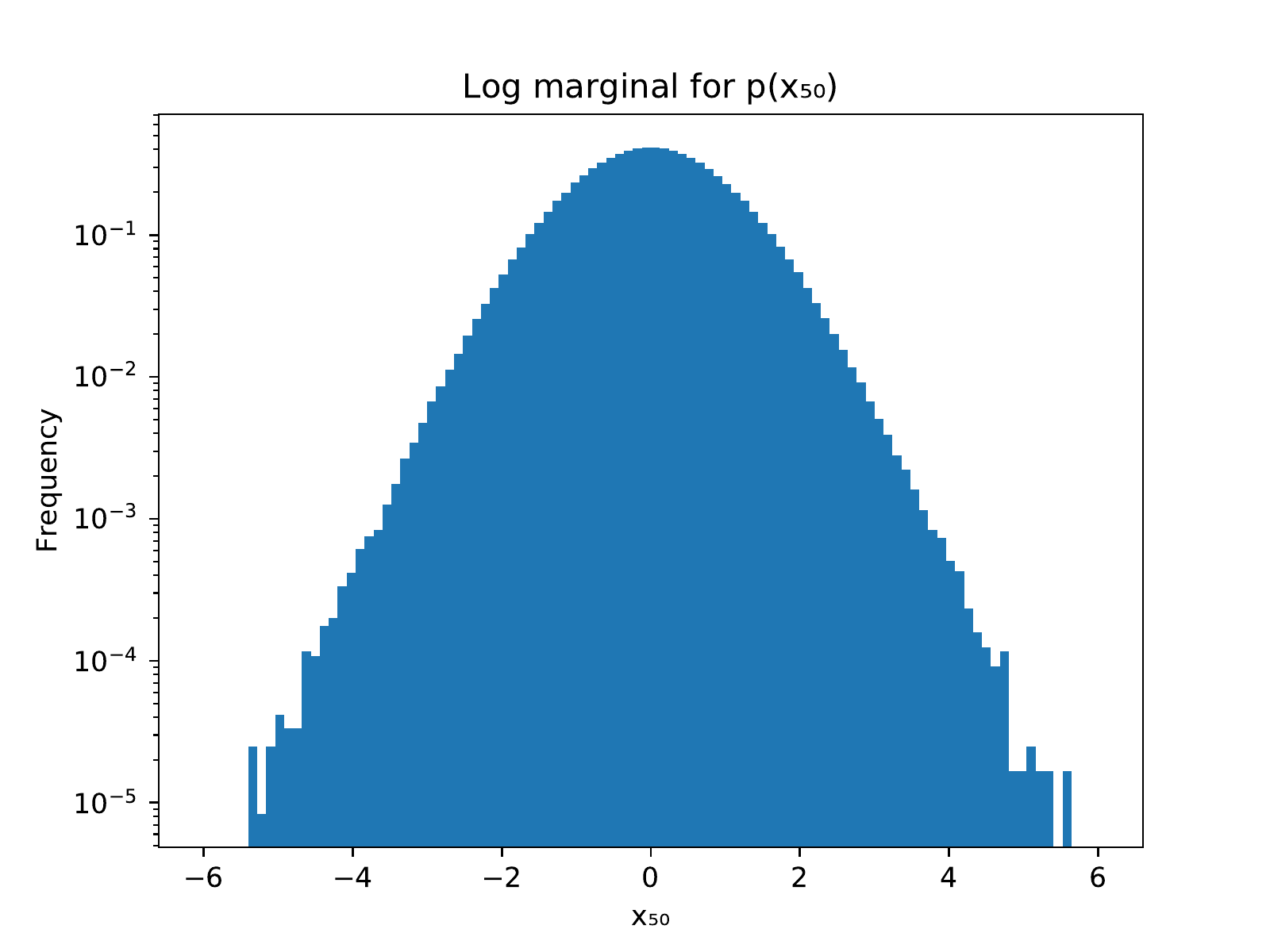} 
    \caption{Top: Marginal distributions for $x^{(0)}$,  $x^{(18)}$, $x^{(20)}$, $x^{(50)}$. Bottom: log domain for these distributions}
    \label{fig:histograms}
\end{figure}

\paragraph{Complete distribution.}
This analysis suggests that the sampler for $\bw$ could be replaced with the following model:
\begin{align}
    x^{j} &\sim p(x^{j}) \\
    \by &= \bV \bx + \mathbf{\mu}
\end{align}
where the one-dimensional distributions $p(x^{j})$ are in some suitable form to capture the unimodal distributions described above.  This is a multivariate distribution slightly distorted from a Gaussian. 

This representation would have substantially fewer parameters than the $M(\bz)$ representation in StyleGAN.

\section{BigGAN Principal Directions are Class-agnostic}
\label{sm:sec6}
\figEditTransferability

Figure~\ref{fig:Transferability} shows examples of transferring edits between BigGAN classes, illustrating our observation that PCA components seem to be the same across different BigGAN classes.

\section{Entanglements and Disallowed Combinations}
\label{sm:sec7}

Most of our edits work across different starting images in a predictable way. For example, the head rotation edit accurately rotates any head in our tests. However, as discussed in Section 3.2 of the paper, some edits show behavior that may reveal built-in priors or biases learned by the GAN.  
These are illustrated in Figures \ref{fig:Baldness} ("baldness"), \ref{fig:Makeup} ("makeup"), \ref{fig:white_hair} ("white hair"), and \ref{fig:Wrinkles} ("wrinkles"): in each case, different results occur when the same edit is applied to difference starting images. Figure \ref{fig:combining} shows an example of combining edits, where one combination is not allowed by the model.

\figEntanglementBaldness
\figEntanglementMakeup
\figEntanglementWhiteHair
\figWrinkles
\figCombiningEdits


\section{Comparisons}
\label{sm:sec8}

Figures \ref{fig:SteerabilityCompSupplementBigGAN}, \ref{fig:SteerabilityCompSupplementFFHQ}, and \ref{fig:SteerabilityCompSupplementCelebaHQ} show comparisons of edits discovered with our method to those discovered by the supervised methods \cite{shen2019interpreting} and \cite{gansteerability}.

\figSteerabilityCompSupplementBigGAN
\figSteerabilityCompSupplementFFHQ
\figSteerabilityCompSupplementCelebaHQ

Sets of 20 normally distributed random directions $\{\mathbf{\hat{r}_{0}}\dots \mathbf{\hat{r}_{19}}\}$ in $\mathcal{Z}$ are shown for StyleGAN2 FFHQ (Figure~\ref{fig:topPCsFFHQ}a), StyleGAN2 Cars (Figure~\ref{fig:topPCsCars}b), StyleGAN2 Cats (Figure~\ref{fig:topPCsCats}b), and BigGAN512-deep Husky (Figure~\ref{fig:topPCsHusky}b). The directions are scaled in order to make the effects more visible.

Figures \ref{fig:RandomBaselineCar}, \ref{fig:RandomBaselineDuck}, \ref{fig:RandomBaselineBedroom}, and \ref{fig:RandomBaselineFFHQ} visualize the significance of the PCA basis as compared to a random basis in latent space.

\figRandomBaselineCar
\figRandomBaselineDuck
\figRandomBaseBedroom
\figRandomBaseFFHQ

\end{document}